\pdfoutput=1
\documentclass[10pt]{article}
\usepackage[OT1]{fontenc}
\usepackage{amsmath,amssymb}
\usepackage{natbib}
\usepackage[dvipsnames,table]{xcolor}
\usepackage{geometry}
\usepackage{smile}
\usepackage{multirow}
\usepackage{enumerate}
\usepackage{tikz}
\usepackage[colorlinks,
            linkcolor=red,
            anchorcolor=blue,
            citecolor=blue
            ]{hyperref} 
\usepackage{algorithm}
\usepackage{algpseudocode}
\usepackage{subcaption}
\usepackage[most,skins]{tcolorbox}
\usepackage{enumitem}
\usepackage{acronym}
\usepackage[T1]{fontenc}

\usepackage{wrapfig}
\newcommand{\tilz}{\tilde{z}}
\newcommand{\htheta}{\hat{\theta}}

\acrodef{ar}[AR]{Auto-Regressive}
\acrodef{t2i}[T2I]{Text-to-Image}
\acrodef{cfg}[CFG]{Classifier-Free Guidance}
\acrodef{ntk}[NTK]{Neural Tangent Kernel}
\acrodef{awr}[AttnReuse]{Attention
Weight Reuse }

\definecolor{sorange}{RGB}{227,94,56}
\definecolor{sblue}{RGB}{23,56,87}

\newcommand{\horange}[1]{{\color{sorange}#1}}
\newcommand{\hblue}[1]{{\color{sblue}#1}}

\definecolor{morandibluelight}{HTML}{E2ECF7} %
\definecolor{morandibluedeep}{HTML}{6E8AAD}  %

\newtcolorbox{highlightbox}{
  enhanced,
  colback=morandibluelight,
  colframe=morandibluelight,
  boxrule=0pt, arc=0pt,
  left=10pt, right=8pt, top=6pt, bottom=6pt,
  borderline west={2pt}{0pt}{morandibluedeep}
}

\newcommand{\att}{\texttt{attn}}
\newcommand{\sm}{\texttt{sm}}
\newcommand{\vis}{\text{vis}}
\newcommand{\txt}{\text{txt}}
\newcommand{\logit}{\mathrm{logit}}
\newcommand{\cond}{\mathrm{cond}}
\newcommand{\uncond}{\mathrm{uncond}}
\newcommand{\mask}{\texttt{MASK}}
\newcommand{\lm}{\text{lm}}
\newcommand{\rep}{\text{rep}}
\newcommand{\diff}{\mathrm{diff}}
\newcommand{\poly}{\mathsf{poly}}

\newcommand{\uc}{\text{uc}}
\newcommand{\bone}{\mathbf{1}}
\newcommand{\reuse}{\mathrm{reuse}}

\theoremstyle{plain}

\usepackage{fullpage}

\usepackage[protrusion=false,expansion=true]{microtype}

\title{\huge Demystifying Classifier-Free Guidance for Auto-Regressive Image Generation}
\author{Zhiling Zhou\textsuperscript{1,$\ast$}\quad Jiachun Pan\textsuperscript{2,$\ast$} \quad Fengzhuo Zhang\textsuperscript{1,$\ast$,$\dagger$}\quad Dirk Bergemann\textsuperscript{1}\quad Zhuoran Yang\textsuperscript{1,3}\\
\textsuperscript{1}Yale University\quad\textsuperscript{2}National University of Singapore\\
\textsuperscript{3}Cisco Foundation AI
}
\date{}
\begin{document}
\maketitle
\renewcommand\thefootnote{}\footnotetext{$\ast$ Equal contribution.}\footnotetext{$\dagger$ Project Lead.}\footnotetext{Correspondence to: \textless{}fengzhuo.zhang@yale.edu\textgreater{},\textless{}zhuoran.yang@yale.edu\textgreater{}}

\begin{abstract}
Classifier-free guidance (CFG) has been widely adopted in autoregressive (AR) models for high-quality image generation. Despite its strong empirical performance, its mechanism in AR models remains unclear. This paper demystifies the mechanism of CFG through both empirical and theoretical studies. By examining the top-ranked semantics of different components in CFG, we show that texture information is primarily encoded in the difference between the conditional and unconditional logits, whereas both logits share previous-token repetition as their leading semantics. This strong repetition semantics obscures the desired texture information, causing greedy decoding from the conditional logits alone to produce nearly pure-color images. Through a training-dynamics analysis of shallow transformers, we prove that this shared repetition semantics does not arise from limitations or failures of pretraining, but instead originates from the texture sparsity of images. We further show that CFG improves generation quality by rectifying this repetition bias during inference. Motivated by the shared semantics between conditional and unconditional logits, we propose Attention Weight Reuse (AttnReuse), which reuses intermediate attention computations from the conditional-logit forward pass to accelerate the unconditional-logit computation. AttnReuse reduces about $25\%$ of attention computation with little performance degradation across different models.
\end{abstract}

\linespread{1.05}\selectfont
\section{Introduction}
\ac{ar} models have recently emerged as a promising paradigm for unifying text and image generation~\citep{wang2024emu3,liu2024lumina,chen2025janus}, offering a competitive alternative to diffusion models~\citep{rombach2022high,esser2024scaling}. Analogous to text generation, they discretize images into visual tokens and cast image generation as next-token prediction. Recent advances have pushed the performance frontier of them through high-quality data curation~\citep{xin2025lumina}, improved discrete latent representations~\citep{tian2024visual}, and flexible inference frameworks~\citep{cui2025emu3}. During inference, most \ac{ar} models adopt \ac{cfg}, which improves visual-token prediction by extrapolating the conditional logits away from the unconditional logits. Empirically, \ac{cfg} substantially improves generation quality and roughly doubles inference computation.

Despite the wide adoption of \ac{cfg}, its mechanism in \ac{ar} models remains poorly understood. Most existing studies analyze \ac{cfg} in diffusion models, where the explanations rely heavily on properties of the reverse diffusion process. However, this process has no direct analogue in \ac{ar} models, making existing understandings of \ac{cfg} not directly applicable. This paper takes a first step toward understanding the mechanism underlying \ac{cfg} for \ac{ar} models. Specifically, we ask:
\begin{highlightbox}
\begin{itemize}[leftmargin=2em]
    \item[1.] \emph{Why is classifier-free guidance effective for \ac{ar} image generation?}
    \item[2.] \emph{Given that \ac{ar} models succeed in text generation without \ac{cfg}, does the necessity of \ac{cfg} in image generation rely on a limitation or failure of \ac{ar} pretraining?}
    \item[3.] \emph{Is doubled inference computation necessary for the effectiveness of \ac{cfg} in \ac{ar} models?}
\end{itemize}
\end{highlightbox}
To answer the first question, we first conduct empirical studies to reveal the roles of different components in \ac{cfg}. Through both qualitative and quantitative analyses, we find that image texture information is primarily captured by the difference between the conditional and unconditional logits, whereas the top-ranked decoded semantics of the conditional logits alone correspond to nearly pure-color images. To understand why this occurs, we compare the conditional and unconditional logits in both logit space and pixel space. The results show that their top-ranked decoded semantics are highly similar. Through causal interventions during image generation, we further show that these shared top-ranked semantics mainly correspond to repeating previous tokens, explaining why decoding from the conditional logits alone tends to produce pure-color images. Overall, our experiments suggest that \ac{cfg} amplifies the texture information encoded in the logit difference, thereby correcting the repetition bias present in the conditional logits.

To answer the second question, we theoretically study a shallow transformer with a single attention layer to explain why conditional and unconditional logits share repetition as their top-ranked semantics. Through training-dynamics analysis, we prove that this shared repetition semantics is not caused by insufficient or failed pretraining. Instead, it can arise from the sparsity of image textures, namely the prevalence of nearly pure-color blocks in training images. This sparsity makes the conditional and unconditional logits of well-pretrained models provably favor previous-token repetition, while \ac{cfg} effectively rectifies this bias to generate the desired images.

To answer the third question, we reduce the computational burden of \ac{cfg} by reusing intermediate states from the conditional-logit computation. Motivated by the shared semantics of conditional and unconditional logits, we propose \ac{awr}. The algorithm decomposes the attention output of conditional logits into weighted text-token values and weighted image-token values, and directly reuses the weighted image-token values as the attention output for unconditional-logit computation. It reduces about $25\%$ of attention computation with little performance degradation across different models. In addition, the efficiency of \ac{awr} generalizes to diffusion models.

Summarizing these results, we identify a clear mechanism underlying the effectiveness of \ac{cfg}: 
\ac{cfg} \textbf{rectifies the repetition bias} in the conditional logits by \textbf{amplifying texture information in the conditional--unconditional logit difference}. This bias arises from the texture sparsity of images, and can be reused to reduce the computation of \ac{cfg}.

\section{Related Work}\label{app:related_work}
\textbf{Auto-Regressive Image Generation} has recently re-emerged as a promising paradigm for unifying text and image generation, alongside diffusion-based models~\citep{rombach2022high} and flow-matching methods~\citep{esser2024scaling}. Early work on autoregressive image generation dates back to PixelCNN~\citep{van2016conditional}, followed by a series of improvements in architecture design and training methodology~\citep{reed2016generating,salimans2017pixelcnn++,reed2017parallel,parmar2018image,chen2018pixelsnail}. However, these methods generate images directly in pixel space. With the development of latent representations and visual tokenizers, such as VQ-VAE~\citep{NIPS2017_7a98af17} and VQGAN~\citep{Esser_2021_CVPR}, subsequent work moved autoregressive modeling to the latent space~\citep{pmlr-v139-ramesh21a,NEURIPS2021_a4d92e2c,yu2022scaling}, substantially easing learning and improving generation quality. Building on these advances, recent works since 2024 have further scaled \ac{ar} models to achieve performance competitive with diffusion models~\citep{sun2024autoregressive,tian2024visual,ma2024star,tang2024hart}. Beyond competitive image-generation performance, \ac{ar} models also provide an effective framework for unified multimodal generation, including text~\citep{chen2025janus}, video~\citep{wang2024emu3}, and depth~\citep{liu2024lumina}. See \citet{xiong2024autoregressive,jiang2024survey} for comprehensive surveys of \ac{ar} models. Recent efforts extend AR modeling to jointly handle text and vision, such as Janus-Pro~\citep{chen2025janus}, which unify understanding and generation within a single transformer. Similarly, Lumina-GPT~\citep{liu2024lumina, xin2025lumina} explores AR modeling for multi-modal signals including depth and geometry, while the Emu3~\citep{wang2024emu3} series and its successor Emu3.5~\citep{cui2025emu3} scale AR frameworks to support high-quality image and video generation with strong cross-modal alignment. These works highlight the flexibility of AR modeling in handling diverse modalities within a unified token space.

\paragraph{Classifier-Free Guidance} \hspace{-0.8em} is a simple test-time algorithm for diffusion models that substantially improves generation quality~\citep{ho2022classifier}. A large body of work has subsequently sought to further enhance \ac{cfg}. For example, prior studies improve \ac{cfg} by designing low-quality score estimates~\citep{karras2024guiding}, imposing manifold constraints~\citep{chung2024cfg++,cai2026improving}, incorporating frequency-domain information~\citep{sadat2025guidance}, and designing adaptive guidance strategies~\citep{papalampidi2025dynamic,malarz2025classifier,galashov2025learn}. Beyond these algorithmic improvements, another line of work aims to understand why the original \ac{cfg} is effective. \citet{chidambaram2024does,fu2024unveil,zhao2026studying,wu2024theoretical,ye2025exactly} study simplified learning settings for diffusion models, deriving theoretical insights into guidance in large pretrained models. In contrast, \citet{bradley2024classifier,xia2025rectified,li2025provable,jiao2025towards} directly assume bounded pretraining error and analyze the sampling process itself. These works interpret \ac{cfg} through the lens of sampling algorithms and establish convergence guarantees from different perspectives. Most of the above works treat the role of \ac{cfg} as fixed throughout sampling. By contrast, \citet{gao2026c,lehman2025understanding,li2025towards,jin2025stage,rojas2026improving,wang2024analysis,kynkaanniemi2024applying} examine how the effect of \ac{cfg} varies across different sampling stages, propose non-stationary views of guidance, and develop adaptive algorithms accordingly. From a different perspective, \citet{wang2026cfg} studies \ac{cfg} through the lens of control theory. However, all of these works focus on understanding \ac{cfg} in diffusion models. In contrast, our work provides an understanding of \ac{cfg} in \ac{ar} models.

\paragraph{Theoretical studies of training dynamics} \hspace{-0.8em} have provided important insights into neural networks for computer vision. Existing works cover a broad range of architectures, including MLPs~\citep{du2019gradient}, CNNs~\citep{cao2022benign}, and Vision Transformers~\citep{jelassi2022vision}, as well as different learning regimes, including lazy learning~\citep{jacot2018neural} and feature learning~\citep{lu2020mean}. Following the seminal work of \citet{jacot2018neural}, a line of research uses the \ac{ntk} to study the optimization and generalization behavior of neural networks. For CNNs, \citet{arora2019exact} develops the convolutional \ac{ntk}, while \citet{arora2019fine} derives explicit kernel expressions and shows how the kernel spectrum determines the convergence rate. \citet{mei2021learning} further relates the kernel regime to invariance properties of neural networks; see \citet{golikov2022neural} for a survey of \ac{ntk} analyses. Beyond the kernel regime, \citet{lu2020mean} and \citet{xiao2018dynamical} study mean-field limits of CNNs and establish convergence results. Another line of work builds problem-specific feature models to characterize training dynamics. For example, \citet{allen2020towards} shows the importance of multi-view data structures for ensemble learning, and \citet{cao2022benign} derives phase-transition results for benign overfitting. More recently, several works have studied the learning dynamics of transformers. \citet{jelassi2022vision} and \citet{huang2024theoretical} analyze how transformers learn spatial relationships among visual features. Our work also falls into this category: we study how the attention module contributes to \ac{cfg} in \ac{ar} image generation.

\section{Preliminaries}
\label{sec:pre}
\textbf{Auto-regressive models} provide a promising paradigm for jointly modeling images and text within a unified architecture~\citep{liu2024lumina,chen2025janus}. These \ac{ar} models represent images as visual tokens in $\calX_{\vis}$ and text as textual tokens in $\calX_{\txt}$~\citep{wang2024emu3}. Let $\calX=\calX_{\vis}\cup\calX_{\txt}$ denote the token vocabulary, $\Delta(\calX)$ the set of distributions over $\calX$, and $\calX^*$ the set of all finite token sequences over $\calX$. A model $P_\theta:\calX^*\to\Delta(\calX)$ with parameters $\theta\in\Theta$ maps each token sequence to a distribution over the next token. It is trained by next-token prediction, where the total cross-entropy loss decomposes into a textual loss and a visual loss~\citep{wang2024emu3,chen2025janus}. The textual loss is the same as in standard LLM training. For the visual loss, to enable \ac{t2i} generation, the model predicts the next visual token $x_{t+1}\in\calX_{\vis}$ conditioned on the text prompt $z_1,\ldots,z_{T_{\txt}}\in\calX_{\txt}$ and the preceding visual tokens $x_1,\ldots,x_t\in\calX_{\vis}$, for $1\le t<T_{\vis}$. Formally,
\begin{align}
    \calL_{\vis}(\theta)
    = \mathbb{E}_{t,x_{1:t+1}}
    \big[-\log P_\theta(x_{t+1}\given z_1, \cdots, z_{T_{\txt}}, x_1,\cdots,x_t)\big],
    \label{eq:pretrain_loss}
\end{align}
where $T_{\txt},T_{\vis}>0$ denote the numbers of textual and visual tokens, $x_{1:t}\in\calX^t$ denotes the prefix, and the expectation is taken over $t$ sampled uniformly from the visual-token positions $1 \leq t < T_{\vis}$, and over the corresponding prefix $x_{1:t+1}$ drawn from the training-data distribution.

\textbf{Classifier-Free Guidance} is a decoding algorithm originally designed for diffusion models~\citep{ho2022classifier}, which is found to be also effective in the \ac{ar} models. During the decoding, \ac{cfg} enhances the logits $\log P_\theta$ learned in the pretraining by adding the difference between the conditional logit and the unconditional logit. Concretely, \ac{cfg} samples the visual token at temperature $\tau\geq 0$ as follows
\begin{align}
    &x_{t+1}\! \sim\! \sm\big(\tau^{-1}\logit^{\lambda}(\cdot \,|\, z_{1:T_{\txt}}, x_1,\cdots,x_t)\big), \, 
    \logit^{\lambda}\!=\!\logit^{\cond}+\lambda (\logit^{\cond} - \logit^{\uncond}).\label{eq:cfg}
\end{align}
The conditional logit is the logit learned in the pretraining, i.e., $\logit^{\cond}(\cdot \,|\, z_{1:T_\txt}, x_1,\cdots,x_t) = \log P_\theta(\cdot\given z_{1:T_\txt},\allowbreak  x_1,\cdots,x_t)\in\bbR^{|\calX|}$. The unconditional logit is the logit with empty or negative prompts, i.e., $\logit^{\uncond}(\cdot \,|\,\tilde{z}_{1:T_\txt},\allowbreak x_1,\cdots,x_t) = \log P_\theta(\cdot\given \tilde{z}_{1:T_\txt},x_{1}, \cdots,x_t)$, where $\tilde{z}_{1:T_\txt}$ is an empty prompt, i.e., $\tilz_i= \varnothing$, or a negative prompt, e.g., ``Generate an image with low quality''. For diffusion models, existing works explain the effectiveness of \ac{cfg} from a sampling perspective by exploiting the reverse-time dynamics of the diffusion process~\citep{wu2024theoretical,bradley2024classifier}. However, these explanations do not directly apply to \ac{ar} models, where no analogous reverse-time diffusion equation is available. We refer readers to Appendix~\ref{app:related_work} for further related work.

\textbf{Transformer} is the backbone of modern \ac{ar} sequence modeling~\citep{vaswani2017attention}. Its attention modules play a central role in capturing token dependencies by aggregating hidden states through data-dependent weights. Specifically, given the hidden states $H=[h_1,\ldots,h_T]^\top\in\bbR^{T\times D}$ for $T$ tokens, the attention module is defined as
$
    \att(H; W_Q, W_K, W_{V}) = \sm(\mask(H W_{Q}^{\top}W_{K} H^\top))HW_V^\top\in \RR^{T\times D},
$
where $W_Q,W_K,W_V\in\bbR^{D\times D}$ are the query, key, and value weight matrices, $\sm(\cdot)$ denotes the row-wise softmax, and $\mask(\cdot)$ denotes the causal mask operator, i.e., $\mask(M)_{t,l}=M_{t,l}$ if $l\le t$ and $\mask(M)_{t,l}=-\infty$ otherwise. Thus, each token can attend only to itself and previous tokens.

\section{Empirical Understandings of CFG in AR Models}\label{sec:exp}

In this section, we empirically investigate why and how \ac{cfg} contributes to the high-quality image generation in \ac{ar} models. We conduct experiments on three representative models of different scales: LlamaGen (0.7B)~\citep{sun2024autoregressive}, Lumina-mGPT 2.0 (7B)~\citep{xin2025lumina}, and Emu3 (8B)~\citep{wang2024emu3}. Experimental details are provided in Appendix~\ref{app:exp_detail}.

\subsection{Top-ranked Semantics of Components in Classifier-Free Guidance}\label{sec:semantic}

\begin{figure}[t]
    \centering
    \includegraphics[width=0.8\linewidth]{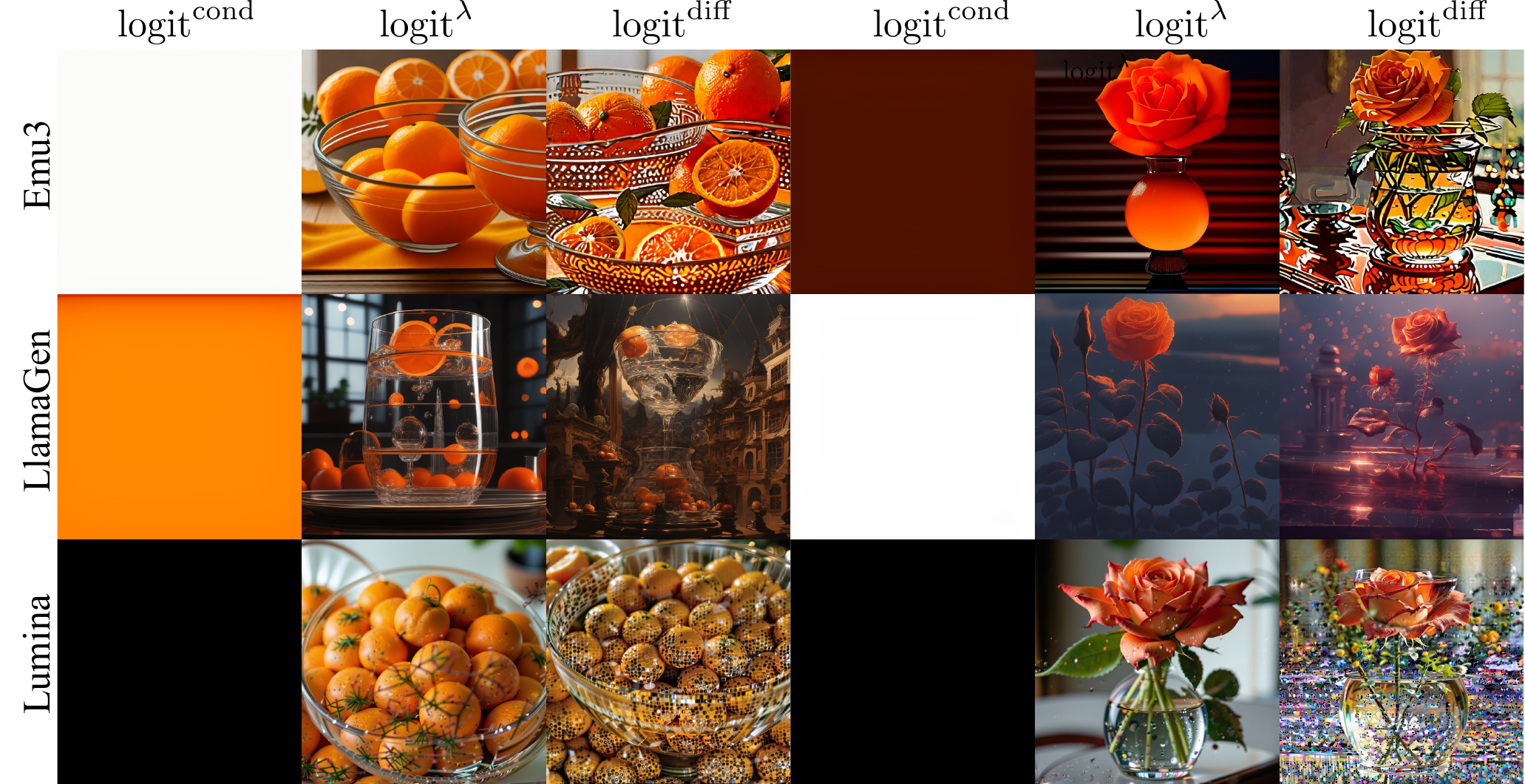}
    \caption{Generated images from conditional logits $\logit^\cond$, CFG-guided logits $\logit^\lambda$, and $\logit^\diff=\logit^\cond-\logit^\uncond$, shown from left to right in each grid. All images are generated by greedy decoding ($\tau=0$). Prompts: \textbf{Left:} A glass bowl filled with oranges on a table. \textbf{Right:} An orange-reddish rose in a vase filled with water on top of a table. Images generated from $\logit^\cond$ collapse to nearly pure-color outputs, whereas images generated from $\logit^\diff$ exhibit excessively complex textures.}
\label{fig:vis-figure}
\vspace{-1em}
\end{figure}

We first conduct experiments to understand the top-ranked semantics of 
$\logit^{\cond}$ and $\logit^{\cond}-\logit^{\uncond}$ in 
Eqn.~\eqref{eq:cfg}, i.e., the semantics induced by their highest-logit 
visual tokens. Given a text prompt $z_{1:T_\txt}$, we generate images by 
greedy decoding from $\logit^{\cond}$ and from 
$\logit^{\cond}-\logit^{\uncond}$, respectively, using temperature 
$\tau=0$. The former corresponds to \ac{cfg} with $\lambda=0$, while the 
latter captures the limiting guidance direction as $\lambda\to\infty$. 
Since zero-temperature decoding always selects the highest-logit token, 
these experiments reveal the dominant semantics encoded by each logit. As 
a baseline, we also generate images using a standard finite guidance scale 
$\lambda$. Experimental details are provided in 
Appendix~\ref{app:exp_detail}, and representative examples are shown in 
Figure~\ref{fig:vis-figure}. The results show that the dominant behavior 
of $\logit^{\cond}$ is to generate images with nearly pure colors. In 
contrast, compared with images generated from $\logit^\lambda$, images 
generated from $\logit^{\cond}-\logit^{\uncond}$ exhibit richer color 
variation, more complex foreground and background structures, and more 
detailed main objects. In summary, 
$\logit^{\cond}-\logit^{\uncond}$ emphasizes visual details and textures, 
but can also induce excessively complex appearances compared with standard 
CFG-generated images. To quantify this observation, we measure the texture complexity of an image token sequence $x_{1:T_\vis}$ as follows. 

\begin{figure}[t]
    \centering
    \includegraphics[width=0.9\linewidth]{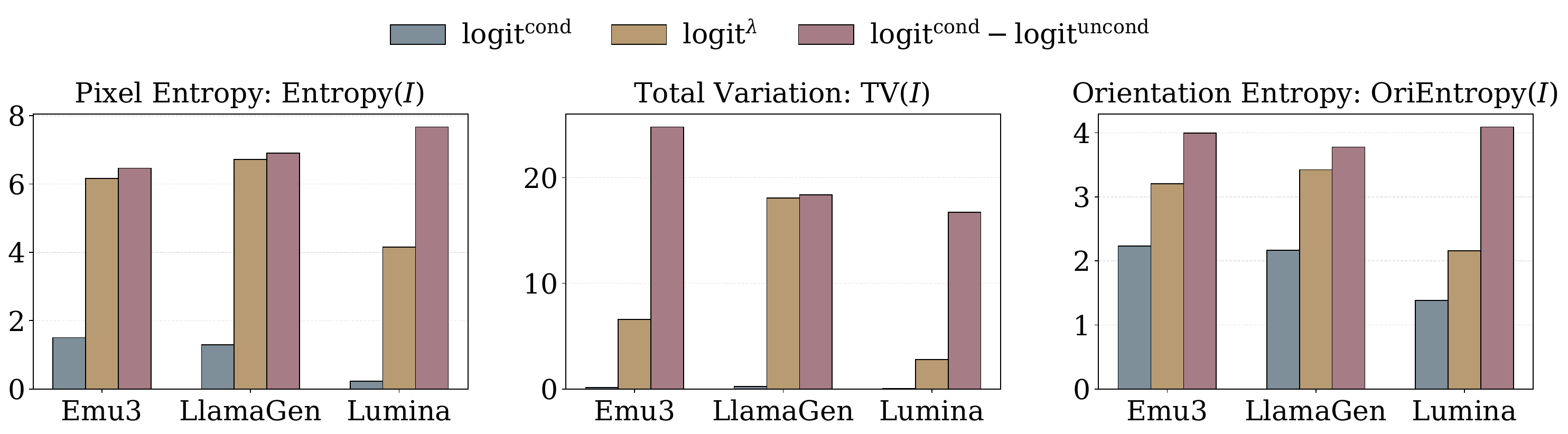}
    \caption{Texture complexity of images generated from $\logit^{\cond}$, $\logit^{\lambda}$, and $\logit^{\diff}=\logit^{\cond}-\logit^{\uncond}$ across multiple metrics. Pixel Entropy, Total Variation, and Orientation Entropy measure pixel-intensity diversity, local intensity variation, and edge-orientation diversity, respectively. Images generated from $\logit^{\diff}$ consistently exhibit the highest texture complexity, whereas those generated from $\logit^{\cond}$ exhibit the lowest.}
    \label{fig:llc}
    \vspace{-1em}
\end{figure}

We first decode the visual token sequence $x_{1:T_\vis}$ into a pixel-space
image, denoted by $I\in\{0,\ldots,255\}^{C\times H\times W}$, where $C=3$
in our experiments. We then quantify the texture complexity using three
metrics: Pixel Entropy $\mathrm{Entropy}(I)$~\citep{shannon1948mathematical},
Total Variation $\mathrm{TV}(I)$~\citep{rudin1992nonlinear}, and Orientation
Entropy $\mathrm{OriEntropy}(I)$~\citep{redies2017high}. They are defined as follows
\begin{align*}
\mathrm{Entropy}(I) = \frac{1}{C} \sum_{c=1}^C H(p_c),\,\mathrm{TV}(I) = \frac{1}{C} \sum_{c=1}^C \mathbb{E}_{i,j} \|\nabla_{i,j} I_{c,i,j}\|_1, \,
\mathrm{OriEntropy}(I) = \frac{1}{C} \sum_{c=1}^C  H(q_c),
\end{align*}
where $\nabla_{i,j} I_{c,i,j} =
[I_{c,i+1,j}-I_{c,i,j},\ I_{c,i,j+1}-I_{c,i,j}]$ denotes the spatial
gradient of channel $c$ at pixel $(i,j)$;
$p_c(v)= (HW)^{-1}\sum_{h=1}^{H}\sum_{w=1}^{W}
\bbI\{I_{c,h,w}=v\}$, for $0\leq v\leq 255$, is the empirical intensity
distribution of channel $c$; and
$q_c(\theta)\propto \sum_{i,j}\|\nabla_{i,j} I_{c,i,j}\|_2
\cdot \bbI\{\mathrm{angle}(\nabla_{i,j} I_{c,i,j})\in \theta\}$
is the magnitude-weighted orientation distribution over orientation bins,
which captures structural regularity. Pixel entropy has been used to
quantify the diversity of intensity distributions~\citep{
haralick2007textural,yuanmlep}; Total Variation has been used to capture
the overall magnitude of local spatial variations~\citep{
rudin1992nonlinear,chambolle2010introduction}; and edge-orientation entropy
has been used to describe the diversity of edge directions~\citep{
redies2017high,grebenkina2018edge,mcadams2025edge}. Figure~\ref{fig:llc} reports the average values of these metrics over
$500$ images generated from different logits. The results show that images
generated from $\logit^{\cond}-\logit^{\uncond}$ exhibit the highest texture
complexity for most models. In contrast, images generated from $\logit^{\cond}$ exhibit the lowest texture complexity, consistent with the
qualitative results in Figure~\ref{fig:vis-figure}. Thus, we reach the
following conclusion.

\begin{highlightbox}
    {\bf Finding 1:} The top-ranked semantics of $\logit^{\cond}$ correspond to pure-color images, whereas those of $\logit^{\cond}-\logit^{\uncond}$ emphasize overly complex textures.
\end{highlightbox}

\subsection{The Dominant Role of Unconditional Logits in Conditional Logits}\label{sec:overlap}

Using the decomposition
\[
\logit^{\cond}=\logit^{\uncond}+(\logit^{\cond}-\logit^{\uncond}),
\]
Finding~1 suggests that the semantics encoded in $\logit^{\cond}$ can be
approximately interpreted as
\begin{align}
    \logit^{\cond}
    \approx
    \logit^{\uncond}
    +
    \text{Complex Textures Info}.
    \label{eq:approx_decomp}
\end{align}
However, zero-temperature decoding from $\logit^{\cond}$ produces images
with nearly pure colors, as shown in Figure~\ref{fig:vis-figure}. This
indicates that the top-ranked semantics of $\logit^{\cond}$ are dominated
by $\logit^{\uncond}$ rather than by the complex texture information. In this section, we verify this interpretation by showing that, when the model is prefilled with images generated under a proper guidance scale $\lambda^*$, $\logit^{\cond}$ and $\logit^{\uncond}$ are aligned among high-logit entries. In other words, the top-ranked semantics of $\logit^{\cond}$ are largely shared with those of $\logit^{\uncond}$, so the unconditional logits explain the dominant semantics in conditional generation.

We measure the similarity between $\logit_t^{\cond}$ and
$\logit_t^{\uncond}$ for $t \in [T_\vis]$ in logit and pixel spaces.
For the logit-space comparison, let
\begin{align*}
\calK_t^{\cond}
= \operatorname{TopK}(\logit_t^{\cond}, k),\quad
\calK_t^{\uncond}
= \operatorname{TopK}(\logit_t^{\uncond}, k)
\end{align*}
denote the top-$k$ token indices of the conditional and unconditional
logits at position $t$, respectively, and define their union as
$\calU_t = \calK_t^{\cond} \cup \calK_t^{\uncond}$.
We measure their normalized $\ell_1$ discrepancy on these high-logit
coordinates by
\begin{align*}
\alpha^{\diff}\!\big(
    (\logit_t^{\cond})_{t=1}^{T_\vis},
    (\logit_t^{\uncond})_{t=1}^{T_\vis}
\big) =
\frac{1}{T_\vis}
\sum_{t=1}^{T_\vis}
\frac{1}{|\calU_t|}
\|
    \logit_{t,\calU_t}^{\cond}
    - \logit_{t,\calU_t}^{\uncond}
\|_1,
\end{align*}
where $\logit_{t,\calU_t}$ denotes the restriction of $\logit_t$ to
coordinates in $\calU_t$.
As a scale baseline, for any logit sequence
$(\logit_t)_{t=1}^{T_\vis}$ with top-$k$ set $\calK_t$, we define
\[
\alpha\!\big((\logit_t)_{t=1}^{T_\vis}\big)
=\frac{1}{kT_\vis}\sum_{t=1}^{T_\vis}
\|\logit_{t,\calK_t}\|_1.
\]
We denote the corresponding values for $\logit^{\cond}$ and
$\logit^{\uncond}$ by $\alpha^{\cond}$ and $\alpha^{\uncond}$,
respectively.

Second, to measure the similarity between 
$(\logit_t^\cond)_{t=1}^{T_\vis}$ and 
$(\logit_t^\uncond)_{t=1}^{T_\vis}$ in pixel space, we first obtain token
sequences $(x_t^\cond)_{t=1}^{T_\vis}$ and 
$(x_t^\uncond)_{t=1}^{T_\vis}$ by greedy decoding at each position. We then
decode these sequences with the VQ decoder into normalized images
$I^\cond,I^\uncond\in[0,1]^{C\times H\times W}$, and define their normalized
$\ell_2$ discrepancy as
\[
\beta^{\diff}(I^\cond,I^\uncond)
=\frac{1}{\sqrt{CHW}}\|I^\cond-I^\uncond\|_2,
\]
where $\|\cdot\|_2$ treats each image as a vector. As scale baselines, we
define
\[
\beta^\cond=\frac{1}{\sqrt{CHW}}\|I^\cond\|_2
\quad\text{and}\quad
\beta^\uncond=\frac{1}{\sqrt{CHW}}\|I^\uncond\|_2.
\]

\begin{figure}[t]
    \centering
    \begin{subfigure}[t]{0.4\linewidth}
        \centering
        \includegraphics[width=\linewidth]{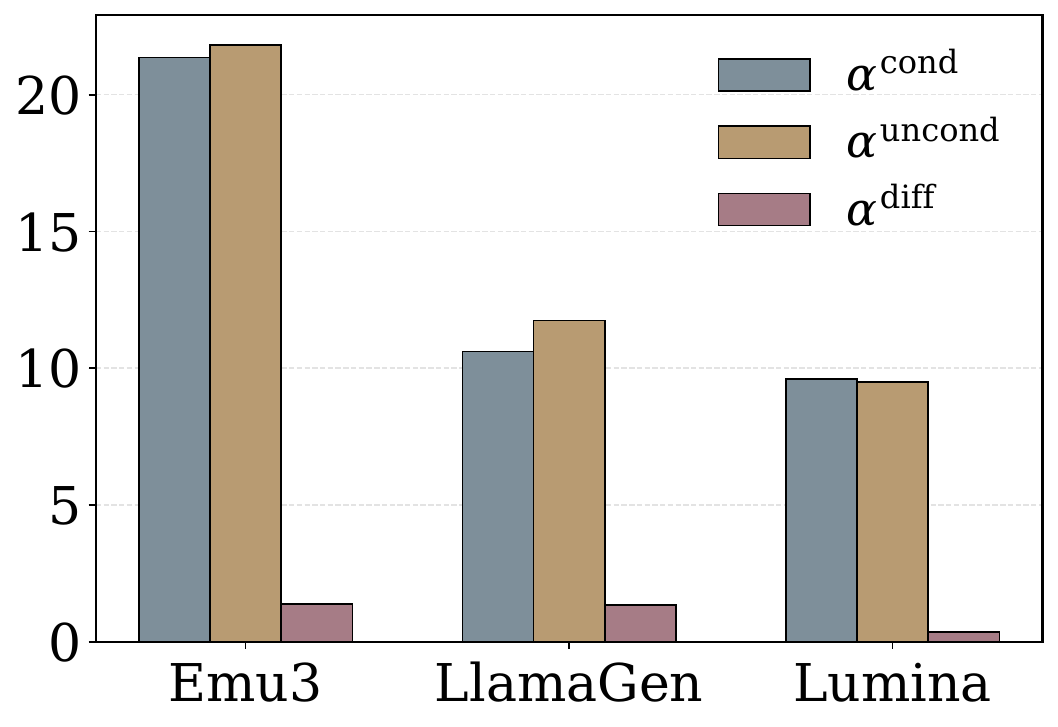}
    \end{subfigure}
    \hspace{2em}
    \begin{subfigure}[t]{0.4\linewidth}
        \centering
        \includegraphics[width=\linewidth]{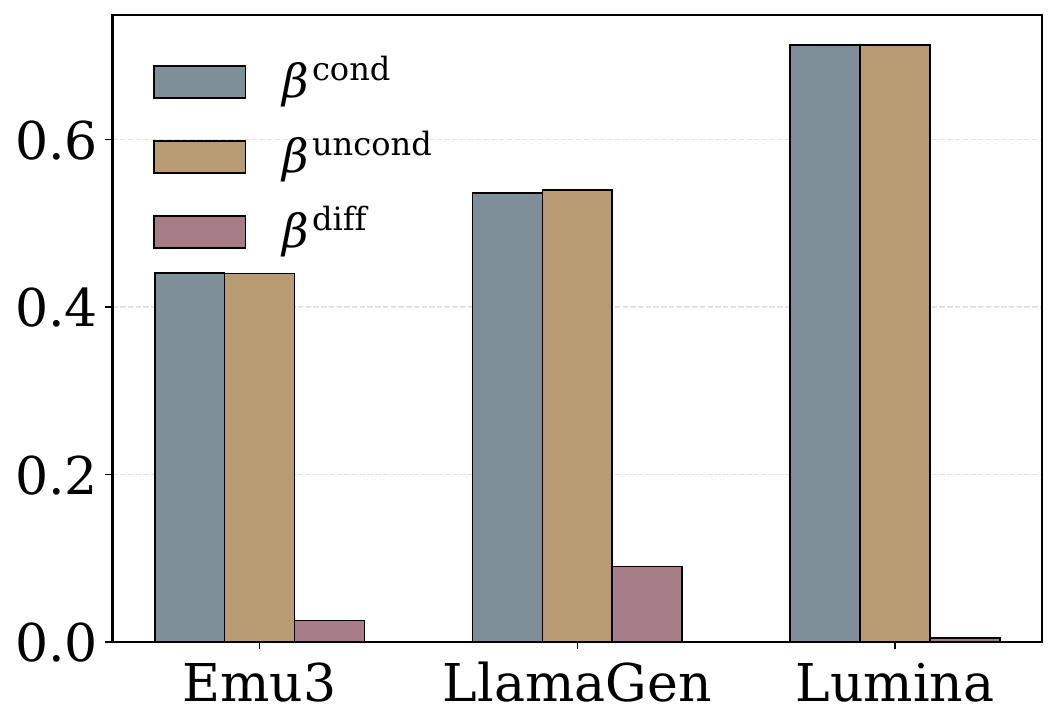}
    \end{subfigure}
    \caption{Similarity measurements between $(\logit_t^{\cond})_{t=1}^{T_\vis}$ and $(\logit_t^{\uncond})_{t=1}^{T_\vis}$ in logit space (\textbf{Left}) and pixel space (\textbf{Right}). Here, $\alpha^\diff$ measures the mean absolute discrepancy between high-logit entries, and $\beta^\diff$ measures the root-mean-square difference between images decoded from the two branches' predictions; superscripts $\cond$ and $\uncond$ denote the corresponding magnitude baselines. Both measurements show small discrepancies relative to these baselines, indicating that the top-ranked entries of $\logit^{\cond}$ are largely dominated by $\logit^{\uncond}$.}
    \label{fig:prefill-exp}
    \vspace{-1em}
\end{figure}

Figure~\ref{fig:prefill-exp} plots the average values of these metrics
across all samples for different models. In both logit and pixel spaces,
$\alpha^\mathrm{diff}$ and $\beta^\mathrm{diff}$ are substantially smaller
than the corresponding scale baselines, namely $\alpha^\cond$ and
$\alpha^\uncond$ in logit space, and $\beta^\cond$ and $\beta^\uncond$ in
pixel space. This indicates that the difference between
$\logit^\cond$ and $\logit^\uncond$ is small relative to the magnitude of
$\logit^\cond$ itself. In other words, most high-logit coordinates of
$\logit^\cond$ are largely dominated by their unconditional counterparts in
$\logit^\uncond$. This is consistent with our previous observation that
zero-temperature decoding from $\logit^\cond$ produces images with nearly
pure colors and few texture details.

\begin{highlightbox}
    {\bf Finding 2:} The high-logit coordinates of the conditional logits are largely driven by
the unconditional logits, indicating that texture-related information is weakly represented in the conditional logits.
\end{highlightbox}

\subsection{Conditional and Unconditional Logits Both Tend to Repeat Previous Tokens}\label{sec:repeat}

Findings~1 and~2 indicate that \ac{cfg} is crucial for high-quality generation: texture-related information has limited influence on the top-ranked conditional predictions, while the unconditional logits dominate
their high-logit coordinates. However, the training objective in Eqn.~\eqref{eq:pretrain_loss} suggests that the conditional logits should capture text-conditioned image content. This raises a natural question: why
does the component shared by the conditional and unconditional logits give rise to pure-color images?

There are two possible explanations for why $\logit^{\cond}$ and
$\logit^{\uncond}$ produce pure-color images under zero-temperature
decoding. One is that the model assigns consistently high logits to a small
set of special tokens, causing the same token to be selected at nearly all
image-token positions. The other is that the model tends to repeat
previously generated tokens, which would also yield nearly uniform images.
We show that the second explanation is correct via causal intervention: conditioned on a prefix
sequence, both $\logit^{\cond}$ and $\logit^{\uncond}$ favor tokens that
repeat the prefix.

To verify the second explanation, we generate images from prescribed visual
prefixes. Specifically, we fix a text condition $z_{1:T_\txt}$ and a visual
prefix $\hat{x}_{1:L}$, where $\hat{x}_{1:L}$ is a repeated sequence of a
chosen visual token and $L$ is the prefix length. We then continue generation
from the conditional logits using greedy decoding:
$x_{t+1}^\cond=\arg\max_{x\in\calX_\vis}
P_\theta(x\mid z_{1:T_\txt},\hat{x}_{1:L},x^\cond_{L+1:t})$,
for $t=L,\ldots,T_\vis-1$. The unconditional generation is performed
analogously, with the text condition replaced by an empty prompt
$\tilde z_{1:T_\txt}$. This yields two image-token sequences:
$(\hat{x}_{1:L},x_{L+1:T_\vis}^\cond)$ and
$(\hat{x}_{1:L},x_{L+1:T_\vis}^\uncond)$.

To quantify repetition behavior, we measure how often the generated tokens
$x_{L+1:T_\vis}$ repeat the pre-assigned prefix token. Specifically, let
$\hat{x}_{1:L}=(s,\ldots,s)$ be a visual prefix consisting of a single token
$s\in\calX_\vis$ repeated for the first $L$ positions. For each generated position
$L<t\le T_\vis$, we define
\[
r_t^\rep((\hat{x}_{1:L},x_{L+1:T_\vis}))=\bbI\{x_t=s\},
\]
where $\bbI\{\cdot\}$ is the indicator function. The repetition rate of the whole generated sequence is then
\[
r^\rep((\hat{x}_{1:L},x_{L+1:T_\vis}))
=\frac{1}{T_\vis-L}\sum_{t=L+1}^{T_\vis}
r_t^\rep((\hat{x}_{1:L},x_{L+1:T_\vis})).
\]

In our experiments, we set the prefix length $L$ to the first $20\%$ of
visual-token positions and sample $s$ uniformly from the set of common
visual tokens. Experimental details are provided in Appendix~\ref{sec:repeat-appendix}.
Figure~\ref{fig:add_line_lumina} reports the average $r_t^{\rep}$ for
Lumina-mGPT across images. The results show that tokens generated from both
$\logit^{\cond}$ and $\logit^{\uncond}$ largely repeat the prescribed token
$s$. Specifically, among the first generated 
\begin{wraptable}{r}{0.45\textwidth}
\centering
\caption{Average repetition ratio $r^\rep$: the fraction of generated tokens that match the prescribed prefix token.}
\label{tab:tail_mean}
\begin{tabular}{lcc}
\toprule
Model & $\logit^\uncond$ & $\logit^\cond$ \\
\midrule
Emu3      & 0.5168 & 0.5479 \\
LlamaGen  & 0.7457 & 0.7517 \\
Lumina-mGPT & 0.6804 & 0.7333 \\
\bottomrule
\end{tabular}
\vspace{-1em}
\end{wraptable}
$10\%$ of visual tokens, the
repetition rate $r_t^{\rep}$ is around $0.8$, indicating strong repetition
behavior. Results for LlamaGen and Emu3 are deferred to
Appendix~\ref{sec:repeat-appendix} and show the same trend. We also report the
sequence-average repetition rate $r^{\rep}$ in Table~\ref{tab:tail_mean},
which shows that at least $50\%$ of generated tokens repeat the assigned
token on average. This empirical evidence supports that both
$\logit^{\cond}$ and $\logit^{\uncond}$ strongly favor tokens that repeat
the existing prefix, validating the hypothesis that local repetition is the
primary driver of uniform, pure-color images.

\begin{figure}[t]
    \centering
    \begin{subfigure}[t]{0.4\linewidth}
        \centering
        \includegraphics[width=\linewidth]{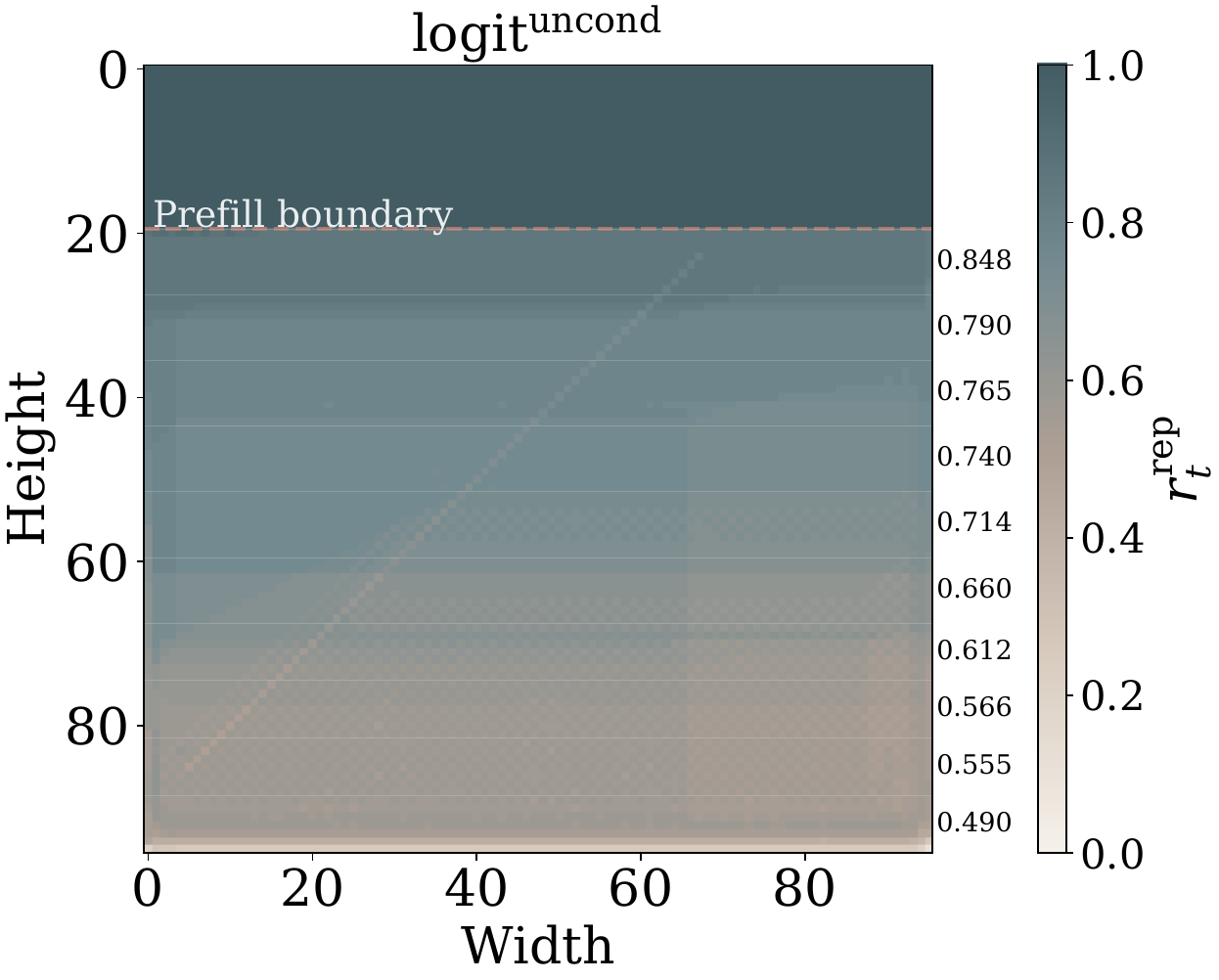}
    \end{subfigure}
    \hspace{2em}
    \begin{subfigure}[t]{0.4\linewidth}
        \centering
        \includegraphics[width=\linewidth]{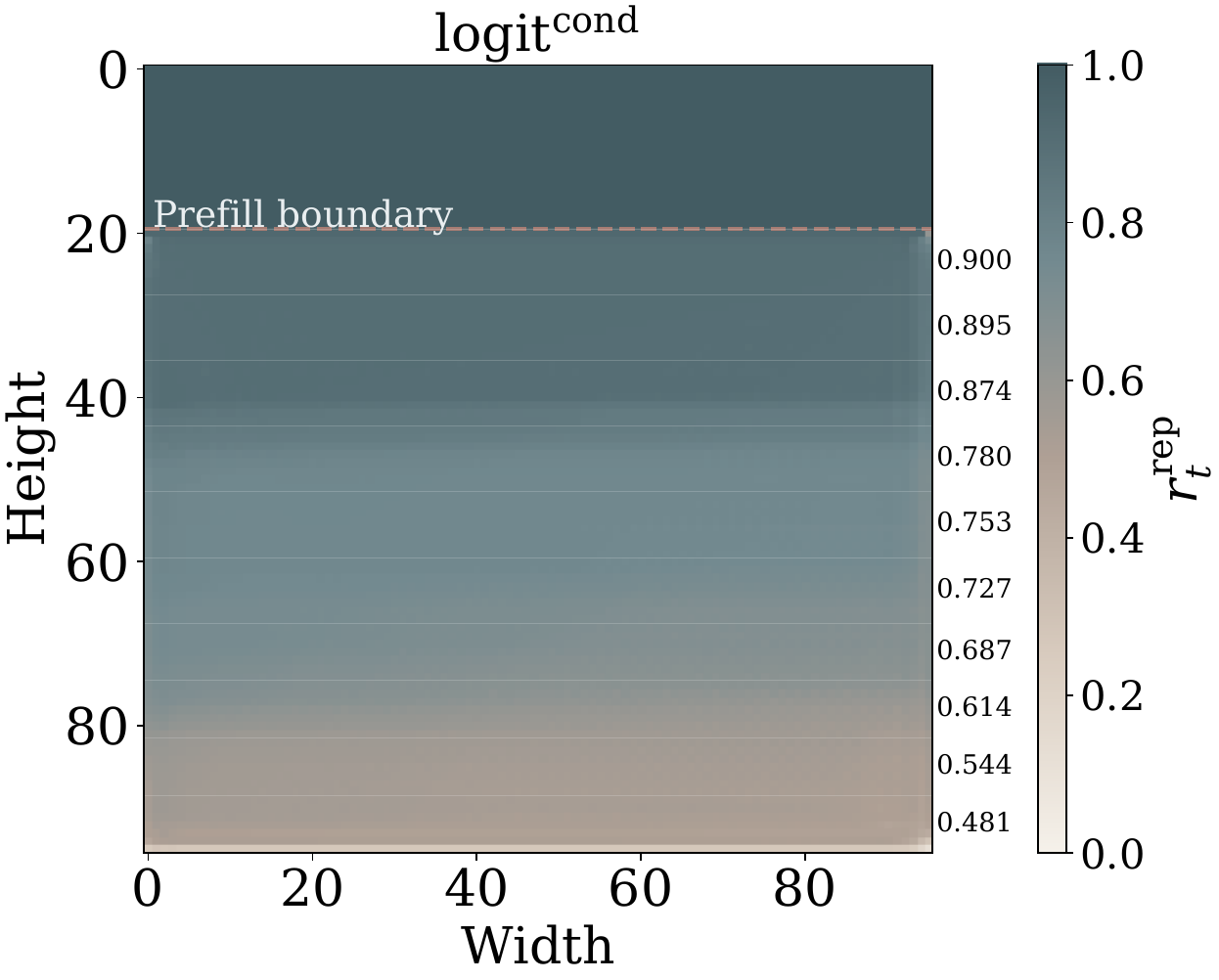}
    \end{subfigure}
    \caption{Position-wise repetition ratio $r_t^\rep$ of Lumina-mGPT, averaged over samples. The ratio measures how often a generated token matches the prescribed prefix token; numbers beside the heatmaps are averages over horizontal bands. Both $\logit^{\cond}$ and $\logit^{\uncond}$ exhibit strong repetition of the prescribed prefix token across generated visual-token positions.}
    \label{fig:add_line_lumina}
\end{figure}

\begin{highlightbox}
    {\bf Finding 3:} The top-ranked semantics of $\logit^{\cond}$ and $\logit^{\uncond}$ are dominated by the repetition of previous tokens, causing zero-temperature decoding from these logits to generate pure-color images.
\end{highlightbox}

\section{Theoretical Understanding of \ac{cfg} in Shallow Transformers}\label{sec:theory}

Section~\ref{sec:exp} empirically shows that \ac{cfg} is important for
high-quality generation: texture-related information has limited influence
on the top-ranked predictions of $\logit^{\cond}$, while the top-ranked
semantics of $\logit^{\cond}$ are dominated by token repetition. However,
these observations do not explain why the training objective in Eqn.~\eqref{eq:pretrain_loss} leads to conditional logits with these properties. In this section, we demystify this phenomenon by studying the training dynamics of shallow transformers on simplified data models. 

We simplify the data model by partitioning visual tokens into visual feature
\begin{wrapfigure}{r}{0.45\textwidth}
\centering
\includegraphics[width=0.8\linewidth]{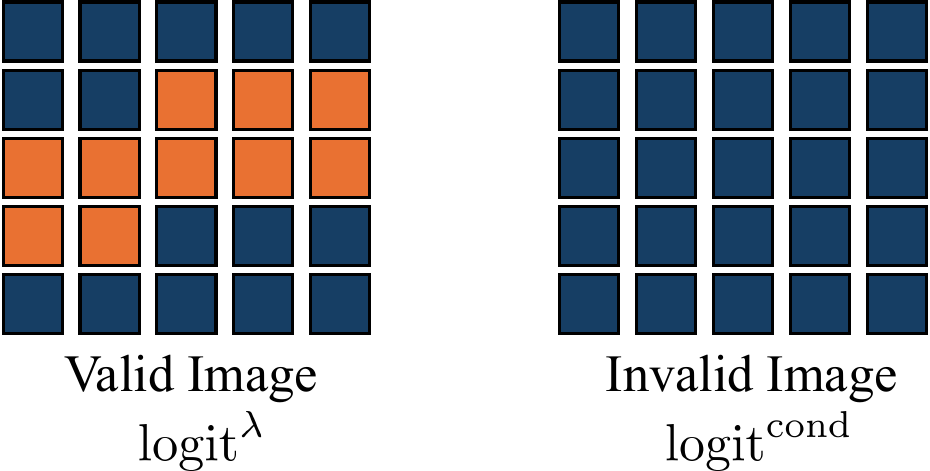}
    \caption{\textbf{Left:} Image $I_{k,j}$ for $k=$\horange{Orange}, $j=$\hblue{Blue} with $\tau_k=8$, $L=10$, and $T_\vis=25$. \textbf{Right:} The invalid image generated by only a conditional model, which will simply repeat the background.}
    \label{fig:theory-fig}
\end{wrapfigure}
tokens and visual background tokens:
$\calX_\vis=\calX_\vis^f\cup\calX_\vis^b$. The feature-token set is
$\calX_\vis^f=\{1,\dots,K\}$, where each $k\in[K]$ corresponds to a unique
texture and semantic concept. Each image contains a single feature-token
type, with token $k$ selected with probability $p_k$ and
$\sum_{k=1}^K p_k=1$. The background-token set is
$\calX_\vis^b=\{K+1,\dots,K+J\}$, where each $K+j$ represents one background
type. Each image contains a single background-token type, with token $K+j$
selected with probability $q_j$ and $\sum_{j=1}^J q_j=1$. Finally, the
text-token set $\calX_\txt=\{K+J+1,\dots,2K+J\}$ represents the semantics
of feature tokens, and we assume without loss of generality that feature
token $k$ is paired with text token $K+J+k$.

An image corresponding to text token $K+J+k$ and background token $K+j$,
for $k\in[K]$ and $j\in[J]$, is represented as a visual-token sequence
$I_{k,j}=(x_1,\dots,x_{T_\vis})\in\calX_\vis^{T_{\vis}}$ of length $T_\vis$. In $I_{k,j}$, the
tokens in the feature block
$\cT_k=\{\tau_k,\dots,\tau_k+L-1\}$ are all equal to the feature token $k$,
while all other tokens are equal to the background token $K+j$. Here,
$\tau_k\ge 3$ denotes the starting position of the feature block for the feature $k$, and all features share the same feature-block length $L$. See Figure~\ref{fig:theory-fig} for an example.

We assume that the length of text token sequence is $1$, i.e., $T_\txt=1$. Then we define the pretraining data distribution $P_{\text{data}}$ over $(z_1, x_{1:T_\vis})$ as follows.
A feature class $k \in [K]$ and a background class $j \in [J]$ are drawn independently according to distributions $(p_k)_{k=1}^K$ and $(q_j)_{j=1}^J$, respectively.
Given $(k,j)$, we set $z_1 = K+J+k \in \calX_\txt$, and $ x_{1:T_\vis}$ are the tokens of $I_{k,j}$.
Further, we assume the feature tokens are sparse in the whole image.
\begin{assumption}[Texture Sparsity]
\label{asm:length-f}
The feature block satisfies $L=\Theta(\log T_\vis)$ as $T_\vis$ increases .
\end{assumption}

\begin{figure}[t]
    \centering
    \begin{subfigure}[t]{0.48\linewidth}
        \centering
        \includegraphics[width=\linewidth]{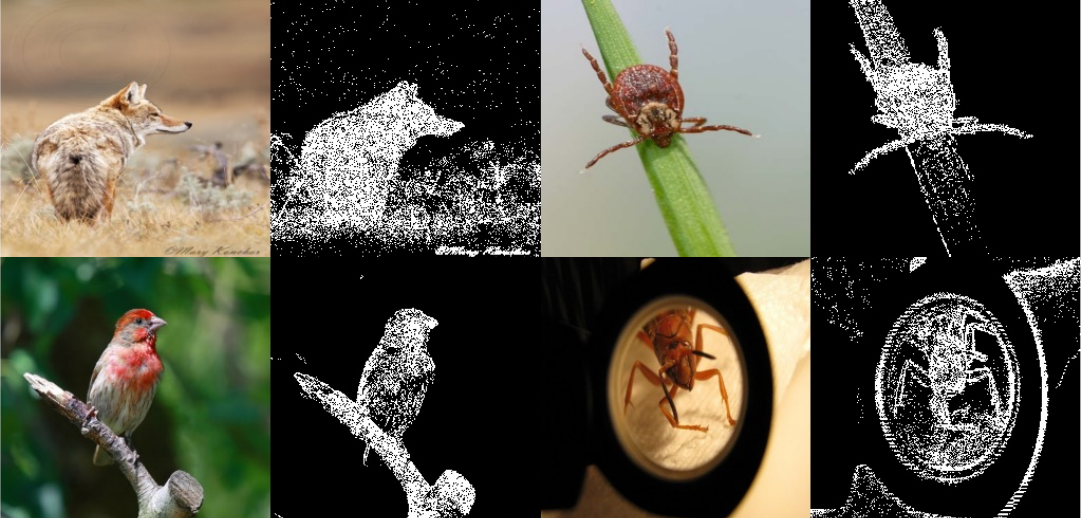}
    \end{subfigure}
    \hfill
    \begin{subfigure}[t]{0.48\linewidth}
        \centering
        \includegraphics[width=0.7\linewidth]{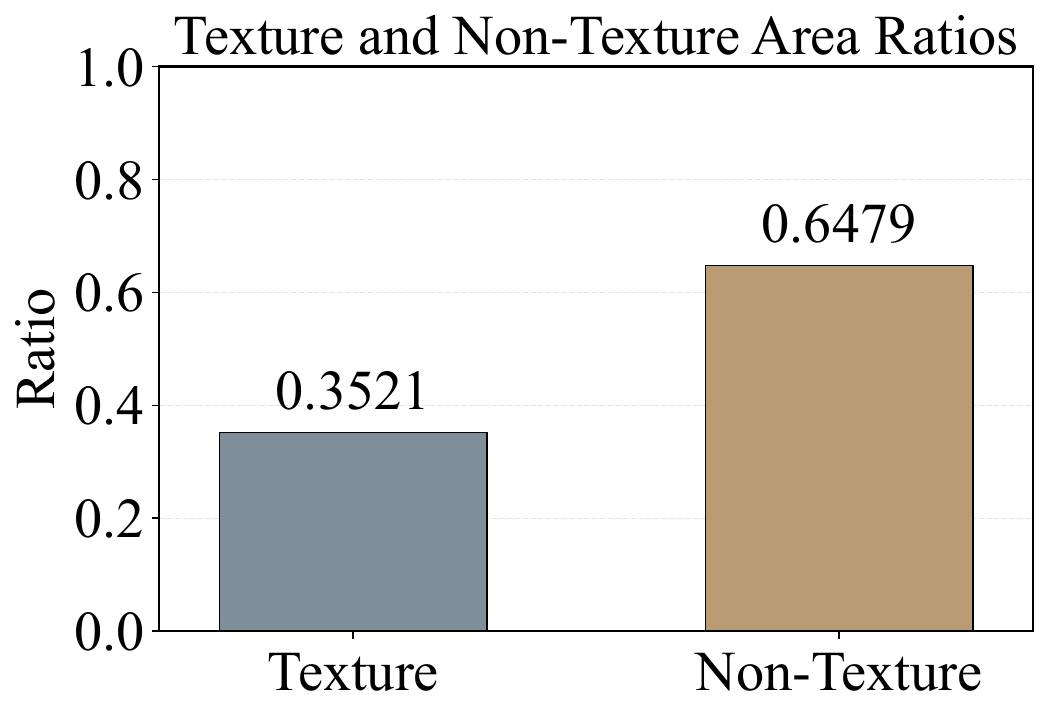}
    \end{subfigure}
    
    \caption{Visualization of texture regions in natural images. \textbf{Left:} Texture regions are highlighted in white. \textbf{Right:} Texture and non-texture area ratios averaged over the ImageNet validation set. The results show that texture regions occupy only a small fraction of the whole image.
}\label{fig:vis-wavelet}
    \vspace{-1em}
\end{figure}

This assumption states that texture occupies only a small portion of the image. This is intuitive, since texture often appears as localized high-frequency patches. To verify this empirically, we extract texture and
flat regions from the ImageNet validation set~\citep{deng2009imagenet} using a single-level 2D wavelet decomposition~\citep{mallat1999wavelet}.
The approximation coefficients capture smooth structures, while the horizontal, vertical, and diagonal detail coefficients encode high-frequency texture and edges~\citep{arivazhagan2003texture,lu1997unsupervised}. We
aggregate the detail components into a high-frequency response map, compute
its magnitude, and apply Otsu thresholding~\citep{otsu1979threshold} to
identify texture pixels. The texture-area ratio is the fraction of pixels
above the threshold, and the flat-area ratio is its complement. As shown in
Figure~\ref{fig:vis-wavelet}, non-texture regions occupy most of the image
area on average, supporting our assumption.

We then introduce the shallow transformer used to learn the image
distribution. The token $x_t$ at position $t$ is encoded as
$h_t=[f_{x_t}^\top,e_t^\top]^\top\in\RR^D$, where
$f_{x_t}\in\bbR^d$ is the token embedding, $e_t\in\RR^{d_e}$ is the
positional embedding, and $D=d+d_e$. The input to the transformer is the
concatenation of these embeddings,
$H=[h_0,h_1,\dots,h_{t}]^\top\in\RR^{(t+1)\times D}$,
where $h_0$ denotes the embedding of the text token. For ease of analysis, we
adopt the standard orthonormality assumption on 
embeddings~\citep{allen2020towards,huang2024theoretical,nichani2024transformers}.

\begin{assumption}[Orthonormality of Embeddings]
\label{asm:ortho}
    The token embeddings and positional embeddings are orthonormal, i.e., $f_i^\top f_{i'} = \bbI\{i = i'\}$, and $e_t^\top e_{t'} = \bbI\{t = t'\}$.
\end{assumption}

We consider a single-layer transformer. Given input $H \in \RR^{(t+1) \times D}$, the parameter-merged attention operator is defined as 
$\att(H; \widetilde{W}_{QK}, \widetilde{W}_V) = \sm(\mask(H \widetilde{W}_{QK} H^\top)) H \widetilde{W}_V^\top \in \RR^{(t+1) \times D}$, 
where $\widetilde{W}_{QK},\widetilde{W}_V \in\RR^{D \times D}$ are the merged query-key parameter and value parameter, respectively. They are specified as
\begin{align*}
\widetilde{W}_{QK} = \begin{bmatrix} 0 & 0 \\ W_{QK} & 0 \end{bmatrix}, \quad
\widetilde{W}_V = \begin{bmatrix} W_V & 0 \\ 0 & 0 \end{bmatrix},
\end{align*}
where $W_{QK} \in \RR^{d_e \times d}$ captures the correlation between token embedding and positions, while $W_V \in \RR^{d \times d}$ captures the correlation among token embeddings. The attention output is then mapped to logits by the Language-Model (LM) head: $\logit = \att(H; \widetilde{W}_{QK}, \widetilde{W}_V)\widetilde{W}_{\lm} \in \RR^{(t+1) \times (K+J)}$. Here the parameter of the LM head $\widetilde{W}_{\lm}\in\RR^{D\times (K+J)}$ is fixed to $\widetilde{W}_{\lm}=[W_{\lm}^\top, 0]^\top$, where $W_{\lm} = [f_1,\dots,f_{K+J}] \in \RR^{d \times (K+J)}$ ties the visual token embeddings to the LM head. Let $\logit_t$ denote the logit of the $t$-th visual token, i.e., the $(t+1)$-th row of $\logit$. The next-token distribution is $P_\theta(x_{t+1} \mid z_1, x_1, \cdots, x_{t}) = \sm(\logit_{t})$. We train $\theta=(W_{QK},W_V)$ using the cross-entropy loss $\mathcal{L}_{\vis}(\cdot)$ in
Eqn.~\eqref{eq:pretrain_loss}, under the population distribution
$P_{\mathrm{data}}$ over $(z_1,x_{1:T_\vis})$. Following prior theoretical
analyses of transformer training dynamics~\citep{nichani2024transformers,wen2024sparse}, we initialize $W_{QK}$ and $W_V$ at zero
and use a two-stage protocol: first updating $W_V$ with $W_{QK}$ fixed, and
then updating $W_{QK}$ with $W_V$ fixed. Details are deferred to
Appendix~\ref{app:algo}. We denote the resulting pretrained parameter by
$\hat{\theta}$.

Given the pretrained model $\hat{\theta}$, we generate the image-token
sequence $\hat{I}_{k,j}^{\lambda}(\hat{\theta})
=(\hat{x}_1,\dots,\hat{x}_T)\in\calX_\vis^{T_\vis}$ corresponding to text token
$K+J+k$ and background token $K+j$, for $k\in[K]$ and $j\in[J]$, as
follows. We first fix the text token and the first visual token as
$z_1=K+J+k$ and $\hat{x}_1=K+j$, respectively. We then generate the
remaining visual tokens by greedy decoding:
$\hat{x}_{t+1}=\arg\max_{x\in\calX_\vis}
\logit_t^\lambda(x\mid z_1,\hat{x}_{1:t})$,
for $t=1,\dots,T_\vis-1$, where $\logit_t^\lambda$ is induced by $\hat{\theta}$
through Eqn.~\eqref{eq:cfg}. We next show that \ac{cfg} is essential for shallow transformers to recover
the desired image-token sequence, i.e.,
$\hat{I}_{k,j}^{\lambda}(\hat{\theta})=I_{k,j}$. To highlight the main
intuition, we first present an informal version of the result and defer the
formal statement to Appendix~\ref{sec:proof}. Throughout this section, we
write $a= \poly(b)$ to indicate that $a$ is asymptotically upper bounded
by a polynomial in $b$.

\begin{theorem}[Informal Statement]
\label{thm:main}
Consider the asymptotic regime $T_\vis\to\infty$ where
$\tau_k=\tau=\poly(T_\vis)$ for all $k\in[K]$,
$J=\poly(T_\vis)$, $K=\Theta(1)$, and
$p_k=\Theta(1)$, $q_j=1/J$. With Assumptions~\ref{asm:length-f} and~\ref{asm:ortho} and suitable learning rates, the
pretrained model $\htheta$ satisfies the following properties.
\begin{itemize}[leftmargin=2em]
  \item[1.] (Minimal Training loss.)
        The population loss is small, i.e., $\mathcal{L}_{\vis}(\htheta)
          = O(T_\vis^{-1}(\log T_\vis)^2).$
    \item[2.] (Alignment of $\logit^{\cond}$ and $\logit^{\uncond}$.)
When prefilled with any target image $I_{k,j}$, the conditional and unconditional models make identical predictions at every visual position, i.e., $\arg\max_{x\in\calX_\vis} \logit_{t,x}^\cond
    =
    \arg\max_{x\in\calX_\vis} \logit_{t,x}^\uncond$, for all $t\in[T_\vis],\ k\in[K],\ j\in[J].$
  \item[3.] (Conditional generation collapses to repeated background tokens.)
With guidance scale $\lambda=0$ (pure conditional generation), the model does not generate a valid image: for every $k\in[K]$ and $j\in[J]$, $
          \hat{I}^{0}_{k,j}(\htheta)=(K+j, \cdots, K+j)\neq I_{k,j}$.
  \item[4.] (CFG recovers valid images.)
       There exists a guidance scale $\lambda^*=\Theta(1)$ such that generation with logits $\logit^{\lambda^*}$ is exact: $\hat{I}^{\lambda^*}_{k,j}(\htheta)=I_{k,j}$, for all $k\in[K],j\in[J]$.
\end{itemize}
\end{theorem}
The small population loss shows that the transformer is well trained, so the phenomena below are not artifacts of failed optimization. The alignment between $\logit^{\cond}$ and $\logit^{\uncond}$ theoretically justifies Finding~2: the top-ranked semantics of the conditional and unconditional logits coincide. Consequently, generation with $\logit^{\cond}$ collapses to repeated background tokens, producing a pure-color image, which is consistent with Findings~1 and~3. Finally, \ac{cfg} with a constant guidance scale $\lambda^*$ extracts the correct signal from the logit difference $\logit^{\cond}-\logit^{\uncond}$ and recovers the target image. Therefore, all the empirical findings in Section~\ref{sec:exp} provably arise in shallow transformers.

\subsection{Proof Sketch}
\begin{figure}[t]
    \centering
    \includegraphics[width=0.8\linewidth]{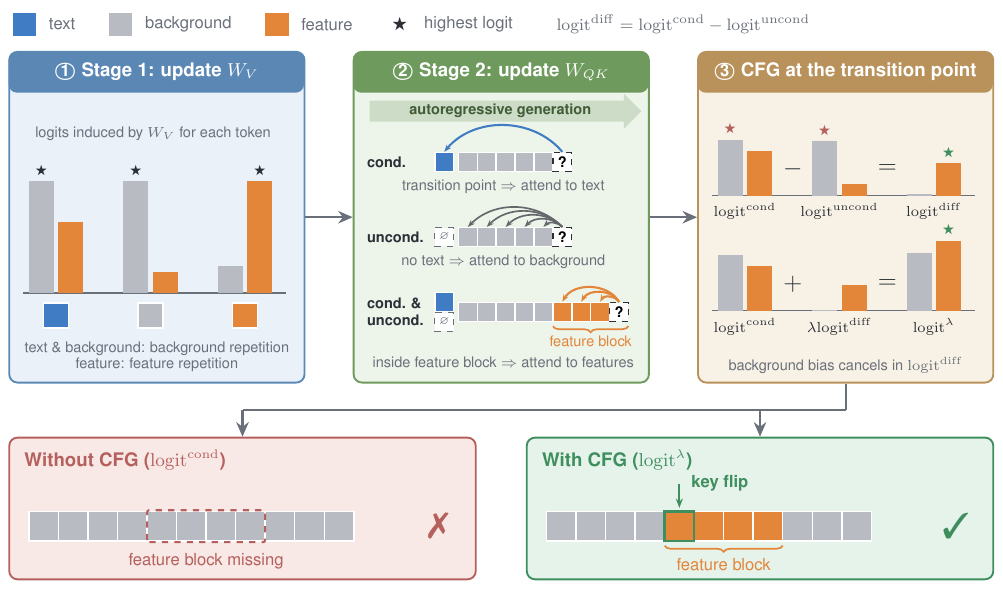}
    \caption{Overview of the proof. \textbf{(1)} Training $W_V$ makes text and background tokens favor background repetition, while feature tokens favor feature repetition. \textbf{(2)} Training $W_{QK}$ makes the conditional branch attend to text when predicting the first feature token. Inside the block, both branches attend to earlier feature tokens. \textbf{(3)} CFG uses the logit difference to generate the first feature token. Repetition then completes the feature block; without CFG, generation repeats the background.}
    \label{fig:proof-sketch}
    \vspace{-1em}
\end{figure}

The proof is divided into two stages, following the two-step training protocol (see Figure~\ref{fig:proof-sketch} for an overview). We first analyze the learning of $W_V$. Since both $W_V$ and $W_{QK}$ are initialized at zero, the attention weights are initially uniform, and the resulting next-token distribution is also uniform. Under this symmetric initialization, the gradient of $W_V$ is determined by correlations between input embeddings and target embeddings. Consequently, the learned $W_V^{(1)}$ takes the form of a sum of outer products of embeddings, enabling it to map an input embedding to a linear combination of target embeddings. Concretely, this transformation can be written as follows:
\begin{align*}
W_{V}^{(1)}f_{K+j} &= \underbrace{-q_{j}A_{K+j}\bar{f}}_{\text{background}\to\text{average}} + \underbrace{q_{j} B_{K+j}f_{K+j}}_{\text{background}\to\text{background}} + \underbrace{q_{j} C_{K+j}\sum_k p_k f_{k}}_{\text{background}\to\text{feature}},\displaybreak[1]\\
W_{V}^{(1)}f_{k} &= \underbrace{-p_{k} A_{k}\bar{f}}_{\text{feature}\to\text{average}} + \underbrace{p_{k}\sum_{j} q_j B_{k}f_{K+j}}_{\text{feature}\to\text{background}} + \underbrace{p_{k} C_{k}f_{k}}_{\text{feature}\to\text{feature}},\displaybreak[1]\\
W_{V}^{(1)}f_{K+J+k} &= \underbrace{-p_{k} A_{K+J+k} \bar{f}}_{\text{text}\to\text{average}} + \underbrace{p_{k}\sum_{j} q_j B_{K+J+k}f_{K+j}}_{\text{text}\to\text{background}} + \underbrace{p_{k}C_{K+J+k}f_{k}}_{\text{text}\to\text{feature}},
\end{align*}

\suppressfloats[t]
where $\bar f=(K+J)^{-1}\sum_{i=1}^{K+J}f_i$ denotes the average visual-token embedding. For each source token $a\in\{k,K+j,K+J+k\}$, the coefficients $A_a,B_a,C_a$ quantify the strengths of the average-embedding, background-token, and feature-token components in $W_V^{(1)}f_a$, respectively. For example, $W_V^{(1)}f_{K+j}$ decomposes into three types of components: a $-q_jA_{K+j}$ contribution along $\bar f$, a $q_jB_{K+j}$ contribution along the background embedding $f_{K+j}$, and a $q_jC_{K+j}$ contribution along each feature embedding $f_k$ for $k\in[K]$. The decompositions of $W_V^{(1)}f_k$ and $W_V^{(1)}f_{K+J+k}$ take analogous forms. In each case, the component strengths are scaled by the probability of the corresponding source token, such as $q_j$ for background tokens $f_{K+j}$ and $p_k$ for feature tokens $f_k$ and corresponding text token $f_{K+J+k}$. Their exact forms are given in Lemma~\ref{lem:grad-wv}. Proposition~\ref{prop:coefficients} further characterizes their asymptotic orders, and Proposition~\ref{prop:param-valid} combines these orders with texture sparsity to yield
\begin{align*}
q_jB_{K+J+k}>C_{K+J+k},
\qquad
q_jB_{K+j} > C_{K+j},
\qquad
q_jB_k<C_k,
\end{align*}
for sufficiently large $T_\vis$. 
These inequalities determine the dominant logit components induced by $W_V^{(1)}$. When comparing logits across visual tokens, the average-embedding component is shared by all visual tokens and therefore cancels out. After removing this common component, $W_V^{(1)}f_{K+J+k}$ and $W_V^{(1)}f_{K+j}$ assign larger logits to the background token $f_{K+j}$ than to the feature token $f_k$, yielding a bias toward background repetition. In contrast, $W_V^{(1)}f_k$ assigns a larger logit to the feature token $k$, yielding a bias toward feature repetition once the feature has already appeared.

We next analyze the second update, where $W_V=W_V^{(1)}$ is fixed and only $W_{QK}$ is updated. Since $W_{QK}$ remains zero before this update, the attention weights are still uniform. Under this uniform attention pattern, the model continues to favor background repetition before the feature appears because of $W_V^{(1)}$, and therefore incurs its largest prediction errors at positions whose next token is the feature token $k$. These errors make the gradient of $W_{QK}$ concentrate on the corresponding feature-prediction positions. As a result, the update learns to correlate these positions with tokens that are informative of the feature embedding $f_k$. More precisely, the updated $W_{QK}^{(1)}$ induces the following position--feature correlation:
\begin{align*}
e_t^\top W_{QK}^{(1)}f_a
\approx \mathbf{1}\{t\in\mathcal F_k\}\cdot \mathrm{Corr}_t(a),
\qquad
\mathcal F_k=\{\tau_k-1,\ldots,\tau_k+L-2\},
\end{align*}
where $\mathcal F_k$ is the set of positions whose next token is feature token $k$, and $\mathrm{Corr}_t(a)$ denotes the corresponding position-token correlation for token $a\in\{K+J+k,K+j,k\}$. Its closed form is given in Proposition~\ref{prop:wqk}.
Consequently, the updated $W_{QK}^{(1)}$ induces a structured attention pattern. In the conditional branch, positions before the feature block attend almost uniformly to the background positions. At the transition point $t=\tau_k-1$, the attention shifts to the text token, which provides information about the upcoming feature. Once the first feature token has been generated, the model attends to the observed feature block; after the feature block ends, the attention returns to an almost uniform pattern. The unconditional branch follows the same structure except that it has no text token to attend to: it remains almost uniform before and after the feature block, and concentrates on the observed feature block once the first feature token appears.

This attention pattern makes the role of \ac{cfg} explicit at the transition point. Fix a feature class $k$ and write $\tau=\tau_k$. In the conditional branch, the position $t=\tau-1$ attends to the text token. However, without guidance, the value carried by this text token still favors background repetition, assigning a larger logit to the background token $K+j$ than to the feature token $k$:
\begin{align*}
\logit_t^{\cond}[K+j]-\logit_t^{\cond}[k]
\approx p_kq_jB_{K+J+k}-p_kC_{K+J+k}>0.
\end{align*}
With guidance, this background-repetition bias is canceled by subtracting the component shared with the unconditional branch. As a result, the guided logit reverses the preference and favors the feature token:
\begin{align*}
\logit_t^\lambda[K+j]-\logit_t^\lambda[k]
\approx -\lambda p_k(q_jB_{K+j}-q_jC_{K+j})<0,
\end{align*}
for any fixed $\lambda\in(0,1)$ and sufficiently large $T_\vis$. 
Thus, \ac{cfg} flips only the transition token from $K+j$ to $k$. After that, feature
repetition keeps the whole feature block correct, and the background dominates
again after the block.
Hence, \ac{cfg} only needs to flip the transition token from $K+j$ to $k$. Once this first feature token is generated, the feature-repetition bias takes over and keeps the entire feature block correct. After the feature block ends, the background-repetition bias again dominates, so the model returns to generating the background token.

Tracing the generation process through the four regimes---before the feature block, at the transition point, inside the feature block, and after the feature block---yields
\begin{align*}
\hat I_{k,j}^{\lambda}(\htheta)=I_{k,j},
\qquad
\hat I_{k,j}^{0}(\htheta)=(K+j,\ldots,K+j).
\end{align*}
In other words, \ac{cfg} recovers the target image, whereas the purely conditional model collapses to repeated background tokens. Finally, the same margin estimates also imply minimal training loss $\mathcal L_{\vis}(\htheta)=O(T_\vis^{-1}(\log T_\vis)^2)$ and alignment of logits
$\arg\max\logit_t^{\cond}=\arg\max\logit_t^{\uncond}$ along the successful path. Therefore, \ac{cfg} succeeds by extracting the sparse feature signal from the logit difference, rather than by changing the top token of either branch alone.

\section{Efficiency Improvement Via Attention Reuse}\label{sec:efficiency}

Building on the insights from Sections~\ref{sec:exp} and~\ref{sec:theory}, we optimize \ac{ar} inference efficiency by analyzing attention decomposition. Since Section~\ref{sec:theory} shows that  a single-layer Transformer can recover \ac{cfg} dynamics, we decompose the attention output for a query $q_t$ into visual and textual contributions. Let $k_{1:t}$ and $v_{1:t}$ be the keys and values for text tokens ($1 \dots T_{\txt}$) and visual tokens ($T_{\txt}+1 \dots t$). The attention output is:
\begin{align*}
    \frac{\sum_{l=1}^{t}\exp(q_t^\top k_{l})v_l}{\sum_{l=1}^{t}\exp(q_t^\top k_{l})}
    = \gamma_t \cdot \frac{\sum_{l=T_{\txt}+1}^{t}\exp(q_t^\top k_{l})v_l}{\sum_{l=T_{\txt}+1}^{t}\exp(q_t^\top k_{l})}
    + (1-\gamma_t) \cdot \frac{\sum_{l=1}^{T_{\txt}}\exp(q_t^\top k_{l})v_l}{\sum_{l=1}^{T_{\txt}}\exp(q_t^\top k_{l})},
\end{align*}
where $\gamma_t = \sum_{l=T_{\txt}+1}^{t}\exp(q_t^\top k_{l})/\sum_{l=1}^{t}\exp(q_t^\top k_{l})$ denotes the fraction of attention mass assigned to visual tokens. This separation clarifies the \ac{cfg} mechanism: the first visual term captures the trajectory shared by both conditional and unconditional branches—consistent with our Section~\ref{sec:exp} observation that both branches tend to repeat previous tokens. In contrast, the second textual term drives the deviation from the unconditional prediction, effectively forming the guidance direction $\logit^{\cond}-\logit^{\uncond}$.

\begin{table*}[t]
\centering
\small
\setlength{\tabcolsep}{3pt}
\begin{tabular}{c l c c c c c c c}
\toprule
Type & Model 
& Global & Entity & Attribute & Relation & Other 
& \textbf{Overall$\uparrow$} & Attn. FLOPs$\downarrow$ \\
\midrule

\multirow{2}{*}{Diffusion}
 & SD3-Medium 
 & 87.9 & 91.0 & 88.8 & 80.7 & 88.7 & 84.1 & 100\% \\

 & \cellcolor{gray!15}\quad + AttnReuse
 & \cellcolor{gray!15}90.1
 & \cellcolor{gray!15}86.5 
 & \cellcolor{gray!15}88.8
 & \cellcolor{gray!15}86.4
 & \cellcolor{gray!15}85.7 
 & \cellcolor{gray!15}82.5 
 & \cellcolor{gray!15} \textbf{87\%} \\

\midrule

\multirow{6}{*}{AR ($\tau=0$)}

 & Emu3 
 & 87.5 & 83.5 & 84.9 & 85.0 & 85.3 & 76.7 & 100\% \\

 & \cellcolor{gray!15}\quad + AttnReuse
 & \cellcolor{gray!15}84.3
 & \cellcolor{gray!15}83.5
 & \cellcolor{gray!15}83.4
 & \cellcolor{gray!15}83.5
 & \cellcolor{gray!15}82.9
 & \cellcolor{gray!15}74.7 
 & \cellcolor{gray!15}\textbf{75\%} \\

 & Janus-Pro 7B 
 & 79.3 & 90.2 & 89.0 & 89.7 & 84.5 & 83.3 & 100\% \\

 & \cellcolor{gray!15}\quad + AttnReuse
 & \cellcolor{gray!15}79.2
 & \cellcolor{gray!15}87.5
 & \cellcolor{gray!15}87.6 
 & \cellcolor{gray!15}89.3 
 & \cellcolor{gray!15}88.6 
 & \cellcolor{gray!15}82.3
 & \cellcolor{gray!15}\textbf{75\%} \\

 & Lumina-GPT 2.0 
 & 69.9 & 65.9 & 71.7 & 69.4 & 70.9 & 52.3 & 100\% \\

 & \cellcolor{gray!15}\quad + AttnReuse
 & \cellcolor{gray!15}71.8
 & \cellcolor{gray!15}68.7
 & \cellcolor{gray!15}70.8
 & \cellcolor{gray!15}69.9
 & \cellcolor{gray!15}64.7
 & \cellcolor{gray!15}52.2
 & \cellcolor{gray!15}\textbf{75\%} \\

\midrule

\multirow{6}{*}{AR ($\tau=1$)}

 & Emu3 
 & 85.2 & 86.6 & 86.8 & 90.2 & 83.1 & 80.6 & 100\% \\

 & \cellcolor{gray!15}\quad + AttnReuse
 & \cellcolor{gray!15}84.5
 & \cellcolor{gray!15}86.5 
 & \cellcolor{gray!15}86.3 
 & \cellcolor{gray!15}89.7
 & \cellcolor{gray!15}84.2 
 & \cellcolor{gray!15}80.4
 & \cellcolor{gray!15}\textbf{75\%} \\

 & Janus-Pro 7B 
 & 86.9 & 88.9 & 89.4 & 89.3 & 89.4 & 84.1 & 100\% \\

 & \cellcolor{gray!15}\quad + AttnReuse
 & \cellcolor{gray!15}84.9
 & \cellcolor{gray!15}88.7 
 & \cellcolor{gray!15}87.1
 & \cellcolor{gray!15}90.1
 & \cellcolor{gray!15}88.9
 & \cellcolor{gray!15}82.0
 & \cellcolor{gray!15}\textbf{75\%} \\

 & Lumina-GPT 2.0 
 & 88.3 & 88.9 & 88.0 & 91.7 & 88.5 & 84.3 & 100\% \\

 & \cellcolor{gray!15}\quad + AttnReuse
 & \cellcolor{gray!15}88.4 
 & \cellcolor{gray!15}88.3 
 & \cellcolor{gray!15}87.4
 & \cellcolor{gray!15}90.8 
 & \cellcolor{gray!15}84.8 
 & \cellcolor{gray!15}83.7
 & \cellcolor{gray!15}\textbf{75\%} \\

\bottomrule
\end{tabular}
\caption{\textbf{DPG-Bench results.} Comparison of model backbones with and without AttnReuse (gray rows). Attention FLOPs are normalized to the standard CFG baseline ($100\%$), encompassing both conditional and unconditional computations. To balance efficiency and fidelity, AttnReuse is applied to middle layers (e.g., 16/32 for AR models), preserving early layers for low-level representations and late layers for final predictions. For SD3-Medium, reuse is further restricted to the final $90\%$ of denoising timesteps to maintain global layout. This strategy reduces relative attention FLOPs to $75\%$ for AR models and $87\%$ for SD3-Medium while maintaining competitive performance.}
\label{tab:dpg}
\vspace{-1.5em}
\end{table*}

\begin{figure}[t]
    \centering
    \includegraphics[width=0.8\linewidth]{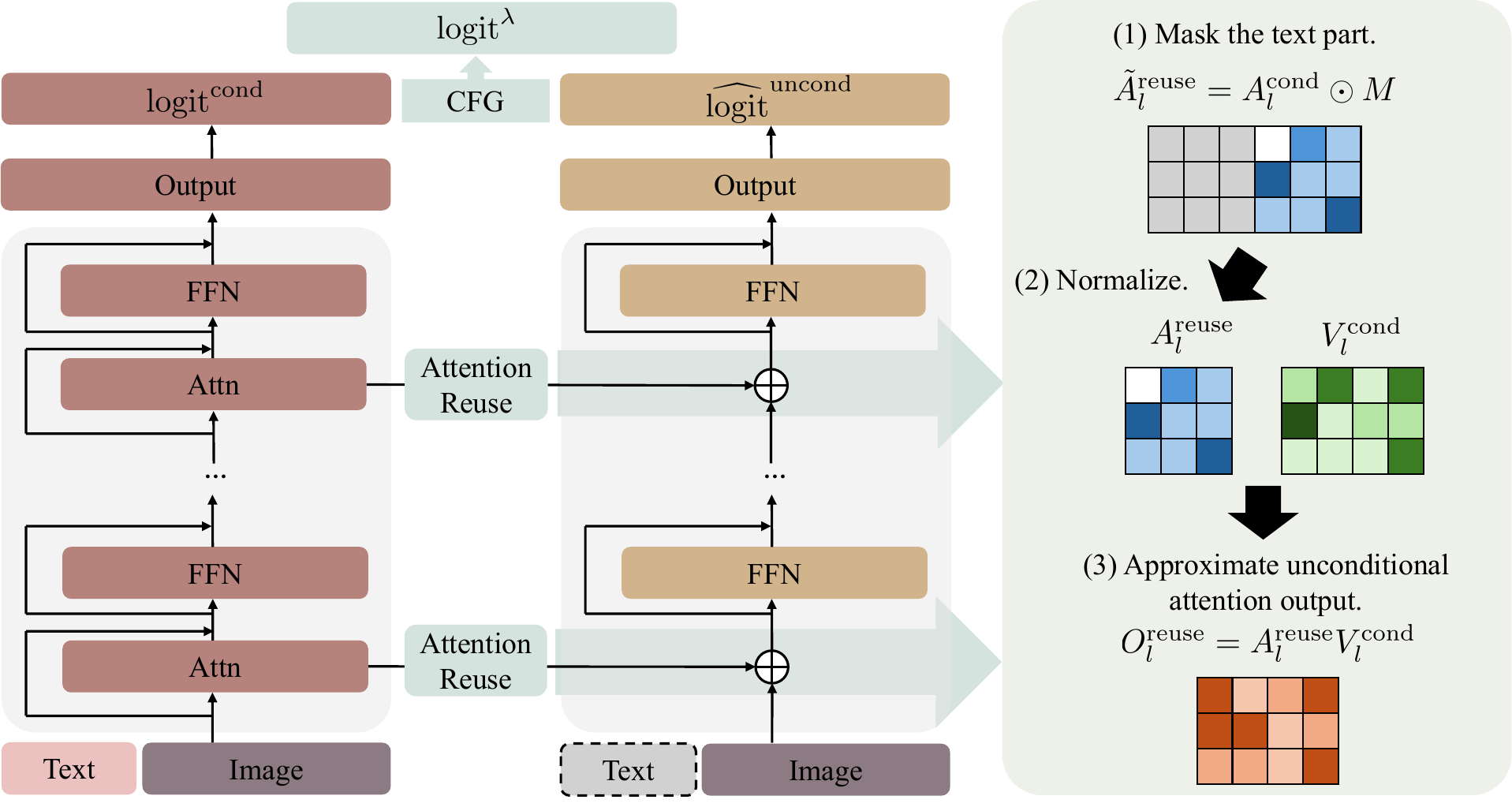}
    \caption{Illustration of the proposed AttnReuse method. In selected layers, we reuse conditional attention weights by masking out text-token entries and renormalizing over image tokens. We then apply these weights to the conditional value vectors to approximate the unconditional attention output.}
\label{fig:attnreuse-fig}
\vspace{-1em}
\end{figure}

Motivated by this observation, we propose \emph{Attention Weight Reuse} (AttnReuse), with the full procedure given in Algorithm~\ref{alg:attnreuse} in Appendix~\ref{sec:efficiency-appendix} and illustrated in Figure~\ref{fig:attnreuse-fig}. Instead of performing a separate unconditional forward pass, AttnReuse reuses the pre-computed conditional attention weights by masking out text-token entries and renormalizing each row over visual-token keys. The resulting distribution efficiently approximates the unconditional attention weights. In practice, we apply AttnReuse to about half of the layers for \ac{ar} models, with implementation details deferred to Appendix~\ref{sec:efficiency-appendix}.

We evaluate AttnReuse on three representative \ac{ar} image generation models: Emu3~\citep{wang2024emu3}, Lumina-GPT 2.0~\citep{xin2025lumina}, and Janus-Pro~\citep{chen2025janus}. To test its generality beyond \ac{ar} models, we also apply it to the diffusion-based Stable Diffusion 3 (SD3)~\citep{esser2024scaling}. We use \textbf{DPG-Bench}~\citep{hu2024ella}, which contains over 1,000 dense and structurally complex prompts for evaluating fine-grained semantic alignment. Results are reported under $\tau=0$ and $\tau=1$, corresponding to greedy decoding and stochastic sampling, respectively. The $\tau=0$ setting isolates the model's top-ranked semantics, while $\tau=1$ better reflects practical generation with sampling diversity. Detailed evaluation settings are shown in Appendix Table~\ref{tab:eval_settings}.

Quantitative results are summarized in Table~\ref{tab:dpg}, with qualitative examples shown in Appendix Fig.~\ref{fig:qual_main}. Across different \ac{ar} models, AttnReuse provides a high-fidelity proxy for the standard unconditional pass: it preserves structural faithfulness with only marginal performance degradation, while reducing attention-related FLOPs by approximately $25\%$. Consistent with these findings, qualitative results show no observable degradation across all evaluated models.

\section{Conclusion}\label{sec:conclusion}
Our work demystifies the mechanism underlying \ac{cfg} in \ac{ar} image generation. We show that texture information is primarily encoded in the difference between the conditional and unconditional logits, whereas both logits share previous-token repetition as their top-ranked semantics. We prove that this repetition semantics arises from texture sparsity rather than training failure, and that \ac{cfg} improves generation quality by rectifying this bias during inference. Motivated by this shared semantics, we propose \ac{awr}, which reduces about $25\%$ of attention computation with little performance degradation. A limitation of our work is its focus on image generation; we leave extensions to audio and video generation for future work.

\section*{Acknowledgments}
Dirk Bergemann gratefully acknowledges financial support from UK DSIT (AP-S2-100058), NSF (SES 2519400), and ONR MURI (N00014-24-1-2742). Zhuoran Yang gratefully acknowledges financial support from UK DSIT (AP-S2-100058) and NSF (DMS 2413243).

\newpage
\linespread{1.05}\selectfont
\bibliographystyle{ims}
\bibliography{reference}

\begin{thebibliography}{79}
\expandafter\ifx\csname natexlab\endcsname\relax\def\natexlab#1{#1}\fi
\expandafter\ifx\csname url\endcsname\relax
  \def\url#1{\texttt{#1}}\fi
\expandafter\ifx\csname urlprefix\endcsname\relax\def\urlprefix{}\fi

\bibitem[{Allen-Zhu and Li(2020)}]{allen2020towards}
\text{Allen-Zhu, Z.} and \text{Li, Y.} (2020).
\newblock Towards understanding ensemble, knowledge distillation and self-distillation in deep learning.
\newblock \textit{arXiv preprint arXiv:2012.09816}.

\bibitem[{Arivazhagan and Ganesan(2003)}]{arivazhagan2003texture}
\text{Arivazhagan, S.} and \text{Ganesan, L.} (2003).
\newblock Texture classification using wavelet transform.
\newblock \textit{Pattern recognition letters}, \textbf{24} 1513--1521.

\bibitem[{Arora et~al.(2019{\natexlab{a}})Arora, Du, Hu, Li and Wang}]{arora2019fine}
\text{Arora, S.}, \text{Du, S.}, \text{Hu, W.}, \text{Li, Z.} and \text{Wang, R.} (2019{\natexlab{a}}).
\newblock Fine-grained analysis of optimization and generalization for overparameterized two-layer neural networks.
\newblock In \textit{International conference on machine learning}. PMLR.

\bibitem[{Arora et~al.(2019{\natexlab{b}})Arora, Du, Hu, Li, Salakhutdinov and Wang}]{arora2019exact}
\text{Arora, S.}, \text{Du, S.~S.}, \text{Hu, W.}, \text{Li, Z.}, \text{Salakhutdinov, R.~R.} and \text{Wang, R.} (2019{\natexlab{b}}).
\newblock On exact computation with an infinitely wide neural net.
\newblock \textit{Advances in neural information processing systems}, \textbf{32}.

\bibitem[{Bradley and Nakkiran(2024)}]{bradley2024classifier}
\text{Bradley, A.} and \text{Nakkiran, P.} (2024).
\newblock Classifier-free guidance is a predictor-corrector.
\newblock \textit{arXiv preprint arXiv:2408.09000}.

\bibitem[{Cai et~al.(2026)Cai, Liu, Su and Wang}]{cai2026improving}
\text{Cai, J.-F.}, \text{Liu, H.}, \text{Su, Z.} and \text{Wang, C.} (2026).
\newblock Improving classifier-free guidance of flow matching via manifold projection.
\newblock \textit{arXiv preprint arXiv:2601.21892}.

\bibitem[{Cao et~al.(2022)Cao, Chen, Belkin and Gu}]{cao2022benign}
\text{Cao, Y.}, \text{Chen, Z.}, \text{Belkin, M.} and \text{Gu, Q.} (2022).
\newblock Benign overfitting in two-layer convolutional neural networks.
\newblock \textit{Advances in neural information processing systems}, \textbf{35} 25237--25250.

\bibitem[{Chambolle et~al.(2010)Chambolle, Caselles, Cremers, Novaga, Pock et~al.}]{chambolle2010introduction}
\text{Chambolle, A.}, \text{Caselles, V.}, \text{Cremers, D.}, \text{Novaga, M.}, \text{Pock, T.} \text{et~al.} (2010).
\newblock An introduction to total variation for image analysis.
\newblock \textit{Theoretical foundations and numerical methods for sparse recovery}, \textbf{9} 227.

\bibitem[{Chen et~al.(2018)Chen, Mishra, Rohaninejad and Abbeel}]{chen2018pixelsnail}
\text{Chen, X.}, \text{Mishra, N.}, \text{Rohaninejad, M.} and \text{Abbeel, P.} (2018).
\newblock Pixelsnail: An improved autoregressive generative model.
\newblock In \textit{International conference on machine learning}. PMLR.

\bibitem[{Chen et~al.(2025)Chen, Wu, Liu, Pan, Liu, Xie, Yu and Ruan}]{chen2025janus}
\text{Chen, X.}, \text{Wu, Z.}, \text{Liu, X.}, \text{Pan, Z.}, \text{Liu, W.}, \text{Xie, Z.}, \text{Yu, X.} and \text{Ruan, C.} (2025).
\newblock Janus-pro: Unified multimodal understanding and generation with data and model scaling.
\newblock \textit{arXiv preprint arXiv:2501.17811}.

\bibitem[{Chidambaram et~al.(2024)Chidambaram, Gatmiry, Chen, Lee and Lu}]{chidambaram2024does}
\text{Chidambaram, M.}, \text{Gatmiry, K.}, \text{Chen, S.}, \text{Lee, H.} and \text{Lu, J.} (2024).
\newblock What does guidance do? a fine-grained analysis in a simple setting.
\newblock \textit{Advances in Neural Information Processing Systems}, \textbf{37} 84968--85005.

\bibitem[{Chung et~al.(2024)Chung, Kim, Park, Nam and Ye}]{chung2024cfg++}
\text{Chung, H.}, \text{Kim, J.}, \text{Park, G.~Y.}, \text{Nam, H.} and \text{Ye, J.~C.} (2024).
\newblock Cfg++: Manifold-constrained classifier free guidance for diffusion models.
\newblock \textit{arXiv preprint arXiv:2406.08070}.

\bibitem[{Cui et~al.(2025)Cui, Chen, Deng, Huang, Li, Liu, Liu, Luo, Wang, Wang et~al.}]{cui2025emu3}
\text{Cui, Y.}, \text{Chen, H.}, \text{Deng, H.}, \text{Huang, X.}, \text{Li, X.}, \text{Liu, J.}, \text{Liu, Y.}, \text{Luo, Z.}, \text{Wang, J.}, \text{Wang, W.} \text{et~al.} (2025).
\newblock Emu3. 5: Native multimodal models are world learners.
\newblock \textit{arXiv preprint arXiv:2510.26583}.

\bibitem[{Deng et~al.(2009)Deng, Dong, Socher, Li, Li and Fei-Fei}]{deng2009imagenet}
\text{Deng, J.}, \text{Dong, W.}, \text{Socher, R.}, \text{Li, L.-J.}, \text{Li, K.} and \text{Fei-Fei, L.} (2009).
\newblock Imagenet: A large-scale hierarchical image database.
\newblock In \textit{2009 IEEE conference on computer vision and pattern recognition}. IEEE.

\bibitem[{Ding et~al.(2021)Ding, Yang, Hong, Zheng, Zhou, Yin, Lin, Zou, Shao, Yang and Tang}]{NEURIPS2021_a4d92e2c}
\text{Ding, M.}, \text{Yang, Z.}, \text{Hong, W.}, \text{Zheng, W.}, \text{Zhou, C.}, \text{Yin, D.}, \text{Lin, J.}, \text{Zou, X.}, \text{Shao, Z.}, \text{Yang, H.} and \text{Tang, J.} (2021).
\newblock Cogview: Mastering text-to-image generation via transformers.
\newblock In \textit{Advances in Neural Information Processing Systems} (M.~Ranzato, A.~Beygelzimer, Y.~Dauphin, P.~Liang and J.~W. Vaughan, eds.), vol.~34. Curran Associates, Inc.

\bibitem[{Du et~al.(2019)Du, Lee, Li, Wang and Zhai}]{du2019gradient}
\text{Du, S.}, \text{Lee, J.}, \text{Li, H.}, \text{Wang, L.} and \text{Zhai, X.} (2019).
\newblock Gradient descent finds global minima of deep neural networks.
\newblock In \textit{International conference on machine learning}. PMLR.

\bibitem[{Esser et~al.(2024)Esser, Kulal, Blattmann, Entezari, M{\"u}ller, Saini, Levi, Lorenz, Sauer, Boesel et~al.}]{esser2024scaling}
\text{Esser, P.}, \text{Kulal, S.}, \text{Blattmann, A.}, \text{Entezari, R.}, \text{M{\"u}ller, J.}, \text{Saini, H.}, \text{Levi, Y.}, \text{Lorenz, D.}, \text{Sauer, A.}, \text{Boesel, F.} \text{et~al.} (2024).
\newblock Scaling rectified flow transformers for high-resolution image synthesis.
\newblock In \textit{Forty-first international conference on machine learning}.

\bibitem[{Esser et~al.(2021)Esser, Rombach and Ommer}]{Esser_2021_CVPR}
\text{Esser, P.}, \text{Rombach, R.} and \text{Ommer, B.} (2021).
\newblock Taming transformers for high-resolution image synthesis.
\newblock In \textit{Proceedings of the IEEE/CVF Conference on Computer Vision and Pattern Recognition (CVPR)}.

\bibitem[{Fu et~al.(2024)Fu, Yang, Wang and Chen}]{fu2024unveil}
\text{Fu, H.}, \text{Yang, Z.}, \text{Wang, M.} and \text{Chen, M.} (2024).
\newblock Unveil conditional diffusion models with classifier-free guidance: A sharp statistical theory.
\newblock \textit{arXiv preprint arXiv:2403.11968}.

\bibitem[{Galashov et~al.(2025)Galashov, Pokle, Doucet, Gretton, Delbracio and De~Bortoli}]{galashov2025learn}
\text{Galashov, A.}, \text{Pokle, A.}, \text{Doucet, A.}, \text{Gretton, A.}, \text{Delbracio, M.} and \text{De~Bortoli, V.} (2025).
\newblock Learn to guide your diffusion model.
\newblock \textit{arXiv preprint arXiv:2510.00815}.

\bibitem[{Gao et~al.(2026)Gao, Zheng, Zou, Yang, Liu, Fan, Zhang, Zhang, Chen, Jiang et~al.}]{gao2026c}
\text{Gao, J.}, \text{Zheng, T.}, \text{Zou, J.}, \text{Yang, F.}, \text{Liu, S.}, \text{Fan, L.}, \text{Zhang, Z.}, \text{Zhang, H.}, \text{Chen, J.}, \text{Jiang, P.-T.} \text{et~al.} (2026).
\newblock C2fg: Control classifier-free guidance via score discrepancy analysis.
\newblock \textit{arXiv preprint arXiv:2603.08155}.

\bibitem[{Ghosh et~al.(2023)Ghosh, Hajishirzi and Schmidt}]{ghosh2023geneval}
\text{Ghosh, D.}, \text{Hajishirzi, H.} and \text{Schmidt, L.} (2023).
\newblock Geneval: An object-focused framework for evaluating text-to-image alignment.
\newblock \textit{Advances in Neural Information Processing Systems}, \textbf{36} 52132--52152.

\bibitem[{Golikov et~al.(2022)Golikov, Pokonechnyy and Korviakov}]{golikov2022neural}
\text{Golikov, E.}, \text{Pokonechnyy, E.} and \text{Korviakov, V.} (2022).
\newblock Neural tangent kernel: A survey.
\newblock \textit{arXiv preprint arXiv:2208.13614}.

\bibitem[{Grebenkina et~al.(2018)Grebenkina, Brachmann, Bertamini, Kaduhm and Redies}]{grebenkina2018edge}
\text{Grebenkina, M.}, \text{Brachmann, A.}, \text{Bertamini, M.}, \text{Kaduhm, A.} and \text{Redies, C.} (2018).
\newblock Edge-orientation entropy predicts preference for diverse types of man-made images.
\newblock \textit{Frontiers in Neuroscience}, \textbf{12} 678.

\bibitem[{Haralick et~al.(2007)Haralick, Shanmugam and Dinstein}]{haralick2007textural}
\text{Haralick, R.~M.}, \text{Shanmugam, K.} and \text{Dinstein, I.~H.} (2007).
\newblock Textural features for image classification.
\newblock \textit{IEEE Transactions on systems, man, and cybernetics} 610--621.

\bibitem[{Ho and Salimans(2022)}]{ho2022classifier}
\text{Ho, J.} and \text{Salimans, T.} (2022).
\newblock Classifier-free diffusion guidance.
\newblock \textit{arXiv preprint arXiv:2207.12598}.

\bibitem[{Hu et~al.(2024)Hu, Wang, Fang, Fu, Cheng and Yu}]{hu2024ella}
\text{Hu, X.}, \text{Wang, R.}, \text{Fang, Y.}, \text{Fu, B.}, \text{Cheng, P.} and \text{Yu, G.} (2024).
\newblock Ella: Equip diffusion models with llm for enhanced semantic alignment.
\newblock \textit{arXiv preprint arXiv:2403.05135}.

\bibitem[{Huang et~al.(2024)Huang, Wen, Chi and Liang}]{huang2024theoretical}
\text{Huang, Y.}, \text{Wen, Z.}, \text{Chi, Y.} and \text{Liang, Y.} (2024).
\newblock A theoretical analysis of self-supervised learning for vision transformers.
\newblock \textit{arXiv preprint arXiv:2403.02233}.

\bibitem[{Jacot et~al.(2018)Jacot, Gabriel and Hongler}]{jacot2018neural}
\text{Jacot, A.}, \text{Gabriel, F.} and \text{Hongler, C.} (2018).
\newblock Neural tangent kernel: Convergence and generalization in neural networks.
\newblock \textit{Advances in neural information processing systems}, \textbf{31}.

\bibitem[{Jelassi et~al.(2022)Jelassi, Sander and Li}]{jelassi2022vision}
\text{Jelassi, S.}, \text{Sander, M.} and \text{Li, Y.} (2022).
\newblock Vision transformers provably learn spatial structure.
\newblock \textit{Advances in Neural Information Processing Systems}, \textbf{35} 37822--37836.

\bibitem[{Jiang and Huang(2024)}]{jiang2024survey}
\text{Jiang, K.} and \text{Huang, J.} (2024).
\newblock A survey on vision autoregressive model.
\newblock \textit{arXiv preprint arXiv:2411.08666}.

\bibitem[{Jiao et~al.(2025)Jiao, Chen and Li}]{jiao2025towards}
\text{Jiao, Y.}, \text{Chen, Y.} and \text{Li, G.} (2025).
\newblock Towards a unified framework for guided diffusion models.
\newblock \textit{arXiv preprint arXiv:2512.04985}.

\bibitem[{Jin et~al.(2025)Jin, Shi and Gu}]{jin2025stage}
\text{Jin, C.}, \text{Shi, Q.} and \text{Gu, Y.} (2025).
\newblock Stage-wise dynamics of classifier-free guidance in diffusion models.
\newblock \textit{arXiv preprint arXiv:2509.22007}.

\bibitem[{Karras et~al.(2024)Karras, Aittala, Kynk{\"a}{\"a}nniemi, Lehtinen, Aila and Laine}]{karras2024guiding}
\text{Karras, T.}, \text{Aittala, M.}, \text{Kynk{\"a}{\"a}nniemi, T.}, \text{Lehtinen, J.}, \text{Aila, T.} and \text{Laine, S.} (2024).
\newblock Guiding a diffusion model with a bad version of itself.
\newblock \textit{Advances in Neural Information Processing Systems}, \textbf{37} 52996--53021.

\bibitem[{Kynk{\"a}{\"a}nniemi et~al.(2024)Kynk{\"a}{\"a}nniemi, Aittala, Karras, Laine, Aila and Lehtinen}]{kynkaanniemi2024applying}
\text{Kynk{\"a}{\"a}nniemi, T.}, \text{Aittala, M.}, \text{Karras, T.}, \text{Laine, S.}, \text{Aila, T.} and \text{Lehtinen, J.} (2024).
\newblock Applying guidance in a limited interval improves sample and distribution quality in diffusion models.
\newblock \textit{Advances in Neural Information Processing Systems}, \textbf{37} 122458--122483.

\bibitem[{Lehman~Pavasovic et~al.(2025)Lehman~Pavasovic, Verbeek, Biroli and Mezard}]{lehman2025understanding}
\text{Lehman~Pavasovic, K.}, \text{Verbeek, J.}, \text{Biroli, G.} and \text{Mezard, M.} (2025).
\newblock Understanding classifier-free guidance: High-dimensional theory and non-linear generalizations.
\newblock \textit{arXiv e-prints} arXiv--2502.

\bibitem[{Li and Jiao(2025)}]{li2025provable}
\text{Li, G.} and \text{Jiao, Y.} (2025).
\newblock Provable efficiency of guidance in diffusion models for general data distribution.
\newblock \textit{arXiv preprint arXiv:2505.01382}.

\bibitem[{Li et~al.(2025)Li, Wang and Qu}]{li2025towards}
\text{Li, X.}, \text{Wang, R.} and \text{Qu, Q.} (2025).
\newblock Towards understanding the mechanisms of classifier-free guidance.
\newblock \textit{arXiv preprint arXiv:2505.19210}.

\bibitem[{Liu et~al.(2024)Liu, Zhao, Zhuo, Lin, Xin, Li, Qin, Qiao, Li and Gao}]{liu2024lumina}
\text{Liu, D.}, \text{Zhao, S.}, \text{Zhuo, L.}, \text{Lin, W.}, \text{Xin, Y.}, \text{Li, X.}, \text{Qin, Q.}, \text{Qiao, Y.}, \text{Li, H.} and \text{Gao, P.} (2024).
\newblock Lumina-mgpt: Illuminate flexible photorealistic text-to-image generation with multimodal generative pretraining.
\newblock \textit{arXiv preprint arXiv:2408.02657}.

\bibitem[{Lu et~al.(1997)Lu, Chung and Chen}]{lu1997unsupervised}
\text{Lu, C.~S.}, \text{Chung, P.~C.} and \text{Chen, C.~F.} (1997).
\newblock Unsupervised texture segmentation via wavelet transform.
\newblock \textit{Pattern recognition}, \textbf{30} 729--742.

\bibitem[{Lu et~al.(2020)Lu, Ma, Lu, Lu and Ying}]{lu2020mean}
\text{Lu, Y.}, \text{Ma, C.}, \text{Lu, Y.}, \text{Lu, J.} and \text{Ying, L.} (2020).
\newblock A mean field analysis of deep resnet and beyond: Towards provably optimization via overparameterization from depth.
\newblock In \textit{International Conference on Machine Learning}. PMLR.

\bibitem[{Ma et~al.(2024)Ma, Zhou, Liang, Bai, Zhao, Chen and Jin}]{ma2024star}
\text{Ma, X.}, \text{Zhou, M.}, \text{Liang, T.}, \text{Bai, Y.}, \text{Zhao, T.}, \text{Chen, H.} and \text{Jin, Y.} (2024).
\newblock Star: Scale-wise text-to-image generation via auto-regressive representations.
\newblock \textit{arXiv preprint arXiv:2406.10797}, \textbf{3}.

\bibitem[{Malarz et~al.(2025)Malarz, Kasymov, Zi{\k{e}}ba, Tabor and Spurek}]{malarz2025classifier}
\text{Malarz, D.}, \text{Kasymov, A.}, \text{Zi{\k{e}}ba, M.}, \text{Tabor, J.} and \text{Spurek, P.} (2025).
\newblock Classifier-free guidance with adaptive scaling.

\bibitem[{Mallat(1999)}]{mallat1999wavelet}
\text{Mallat, S.} (1999).
\newblock \textit{A wavelet tour of signal processing}.
\newblock Elsevier.

\bibitem[{McAdams et~al.(2025)McAdams, Svobodova, Newman, Terry, Mather, Skelton and Franklin}]{mcadams2025edge}
\text{McAdams, P.}, \text{Svobodova, S.}, \text{Newman, T.-J.}, \text{Terry, K.}, \text{Mather, G.}, \text{Skelton, A.~E.} and \text{Franklin, A.} (2025).
\newblock The edge orientation entropy of natural scenes is associated with infant visual preferences and adult aesthetic judgements.
\newblock \textit{Plos one}, \textbf{20} e0316555.

\bibitem[{Mei et~al.(2021)Mei, Misiakiewicz and Montanari}]{mei2021learning}
\text{Mei, S.}, \text{Misiakiewicz, T.} and \text{Montanari, A.} (2021).
\newblock Learning with invariances in random features and kernel models.
\newblock In \textit{Conference on Learning Theory}. PMLR.

\bibitem[{Nichani et~al.(2024)Nichani, Damian and Lee}]{nichani2024transformers}
\text{Nichani, E.}, \text{Damian, A.} and \text{Lee, J.~D.} (2024).
\newblock How transformers learn causal structure with gradient descent.
\newblock \textit{arXiv preprint arXiv:2402.14735}.

\bibitem[{Otsu(1979)}]{otsu1979threshold}
\text{Otsu, N.} (1979).
\newblock A threshold selection method from gray-level histograms.
\newblock \textit{IEEE transactions on systems, man, and cybernetics}, \textbf{9} 62--66.

\bibitem[{Papalampidi et~al.(2025)Papalampidi, Wiles, Ktena, Shtedritski, Bugliarello, Kajic, Albuquerque and Nematzadeh}]{papalampidi2025dynamic}
\text{Papalampidi, P.}, \text{Wiles, O.}, \text{Ktena, I.}, \text{Shtedritski, A.}, \text{Bugliarello, E.}, \text{Kajic, I.}, \text{Albuquerque, I.} and \text{Nematzadeh, A.} (2025).
\newblock Dynamic classifier-free diffusion guidance via online feedback.
\newblock \textit{arXiv preprint arXiv:2509.16131}.

\bibitem[{Parmar et~al.(2018)Parmar, Vaswani, Uszkoreit, Kaiser, Shazeer, Ku and Tran}]{parmar2018image}
\text{Parmar, N.}, \text{Vaswani, A.}, \text{Uszkoreit, J.}, \text{Kaiser, L.}, \text{Shazeer, N.}, \text{Ku, A.} and \text{Tran, D.} (2018).
\newblock Image transformer.
\newblock In \textit{International conference on machine learning}. PMLR.

\bibitem[{Ramesh et~al.(2021)Ramesh, Pavlov, Goh, Gray, Voss, Radford, Chen and Sutskever}]{pmlr-v139-ramesh21a}
\text{Ramesh, A.}, \text{Pavlov, M.}, \text{Goh, G.}, \text{Gray, S.}, \text{Voss, C.}, \text{Radford, A.}, \text{Chen, M.} and \text{Sutskever, I.} (2021).
\newblock Zero-shot text-to-image generation.
\newblock In \textit{Proceedings of the 38th International Conference on Machine Learning} (M.~Meila and T.~Zhang, eds.), vol. 139 of \textit{Proceedings of Machine Learning Research}. PMLR.

\bibitem[{Redies et~al.(2017)Redies, Brachmann and Wagemans}]{redies2017high}
\text{Redies, C.}, \text{Brachmann, A.} and \text{Wagemans, J.} (2017).
\newblock High entropy of edge orientations characterizes visual artworks from diverse cultural backgrounds.
\newblock \textit{Vision Research}, \textbf{133} 130--144.

\bibitem[{Reed et~al.(2017)Reed, Oord, Kalchbrenner, Colmenarejo, Wang, Chen, Belov and Freitas}]{reed2017parallel}
\text{Reed, S.}, \text{Oord, A.}, \text{Kalchbrenner, N.}, \text{Colmenarejo, S.~G.}, \text{Wang, Z.}, \text{Chen, Y.}, \text{Belov, D.} and \text{Freitas, N.} (2017).
\newblock Parallel multiscale autoregressive density estimation.
\newblock In \textit{International conference on machine learning}. PMLR.

\bibitem[{Reed et~al.(2016)Reed, van~den Oord, Kalchbrenner, Bapst, Botvinick and De~Freitas}]{reed2016generating}
\text{Reed, S.}, \text{van~den Oord, A.}, \text{Kalchbrenner, N.}, \text{Bapst, V.}, \text{Botvinick, M.} and \text{De~Freitas, N.} (2016).
\newblock Generating interpretable images with controllable structure.

\bibitem[{Rojas et~al.(2026)Rojas, He, Lai, Takida, Mitsufuji and Tao}]{rojas2026improving}
\text{Rojas, K.}, \text{He, Y.}, \text{Lai, C.-H.}, \text{Takida, Y.}, \text{Mitsufuji, Y.} and \text{Tao, M.} (2026).
\newblock Improving classifier-free guidance in masked diffusion: Low-dim theoretical insights with high-dim impact.
\newblock In \textit{The Fourteenth International Conference on Learning Representations}.

\bibitem[{Rombach et~al.(2022)Rombach, Blattmann, Lorenz, Esser and Ommer}]{rombach2022high}
\text{Rombach, R.}, \text{Blattmann, A.}, \text{Lorenz, D.}, \text{Esser, P.} and \text{Ommer, B.} (2022).
\newblock High-resolution image synthesis with latent diffusion models.
\newblock In \textit{Proceedings of the IEEE/CVF conference on computer vision and pattern recognition}.

\bibitem[{Rudin et~al.(1992)Rudin, Osher and Fatemi}]{rudin1992nonlinear}
\text{Rudin, L.~I.}, \text{Osher, S.} and \text{Fatemi, E.} (1992).
\newblock Nonlinear total variation based noise removal algorithms.
\newblock \textit{Physica D: nonlinear phenomena}, \textbf{60} 259--268.

\bibitem[{Sadat et~al.(2025)Sadat, Vontobel, Salehi and Weber}]{sadat2025guidance}
\text{Sadat, S.}, \text{Vontobel, T.}, \text{Salehi, F.} and \text{Weber, R.~M.} (2025).
\newblock Guidance in the frequency domain enables high-fidelity sampling at low cfg scales.
\newblock \textit{arXiv preprint arXiv:2506.19713}.

\bibitem[{Salimans et~al.(2017)Salimans, Karpathy, Chen and Kingma}]{salimans2017pixelcnn++}
\text{Salimans, T.}, \text{Karpathy, A.}, \text{Chen, X.} and \text{Kingma, D.~P.} (2017).
\newblock Pixelcnn++: Improving the pixelcnn with discretized logistic mixture likelihood and other modifications.
\newblock \textit{arXiv preprint arXiv:1701.05517}.

\bibitem[{Shannon(1948)}]{shannon1948mathematical}
\text{Shannon, C.~E.} (1948).
\newblock A mathematical theory of communication.
\newblock \textit{The Bell system technical journal}, \textbf{27} 379--423.

\bibitem[{Sun et~al.(2024)Sun, Jiang, Chen, Zhang, Peng, Luo and Yuan}]{sun2024autoregressive}
\text{Sun, P.}, \text{Jiang, Y.}, \text{Chen, S.}, \text{Zhang, S.}, \text{Peng, B.}, \text{Luo, P.} and \text{Yuan, Z.} (2024).
\newblock Autoregressive model beats diffusion: Llama for scalable image generation.
\newblock \textit{arXiv preprint arXiv:2406.06525}.

\bibitem[{Tang et~al.(2024)Tang, Wu, Yang, Xie, Chen, Chen, Zhang, Cai, Lu and Han}]{tang2024hart}
\text{Tang, H.}, \text{Wu, Y.}, \text{Yang, S.}, \text{Xie, E.}, \text{Chen, J.}, \text{Chen, J.}, \text{Zhang, Z.}, \text{Cai, H.}, \text{Lu, Y.} and \text{Han, S.} (2024).
\newblock Hart: Efficient visual generation with hybrid autoregressive transformer.
\newblock \textit{arXiv preprint arXiv:2410.10812}.

\bibitem[{Tian et~al.(2024)Tian, Jiang, Yuan, Peng and Wang}]{tian2024visual}
\text{Tian, K.}, \text{Jiang, Y.}, \text{Yuan, Z.}, \text{Peng, B.} and \text{Wang, L.} (2024).
\newblock Visual autoregressive modeling: Scalable image generation via next-scale prediction.
\newblock \textit{Advances in neural information processing systems}, \textbf{37} 84839--84865.

\bibitem[{Van~den Oord et~al.(2016)Van~den Oord, Kalchbrenner, Espeholt, Vinyals, Graves et~al.}]{van2016conditional}
\text{Van~den Oord, A.}, \text{Kalchbrenner, N.}, \text{Espeholt, L.}, \text{Vinyals, O.}, \text{Graves, A.} \text{et~al.} (2016).
\newblock Conditional image generation with pixelcnn decoders.
\newblock \textit{Advances in neural information processing systems}, \textbf{29}.

\bibitem[{van~den Oord et~al.(2017)van~den Oord, Vinyals and kavukcuoglu}]{NIPS2017_7a98af17}
\text{van~den Oord, A.}, \text{Vinyals, O.} and \text{kavukcuoglu, k.} (2017).
\newblock Neural discrete representation learning.
\newblock In \textit{Advances in Neural Information Processing Systems} (I.~Guyon, U.~V. Luxburg, S.~Bengio, H.~Wallach, R.~Fergus, S.~Vishwanathan and R.~Garnett, eds.), vol.~30. Curran Associates, Inc.

\bibitem[{Vaswani et~al.(2017)Vaswani, Shazeer, Parmar, Uszkoreit, Jones, Gomez, Kaiser and Polosukhin}]{vaswani2017attention}
\text{Vaswani, A.}, \text{Shazeer, N.}, \text{Parmar, N.}, \text{Uszkoreit, J.}, \text{Jones, L.}, \text{Gomez, A.~N.}, \text{Kaiser, {\L}.} and \text{Polosukhin, I.} (2017).
\newblock Attention is all you need.
\newblock \textit{Advances in neural information processing systems}, \textbf{30}.

\bibitem[{Wang et~al.(2026{\natexlab{a}})Wang, Liu, Chi, Liu, Xue and Duan}]{wang2026cfg}
\text{Wang, H.}, \text{Liu, Y.}, \text{Chi, J.}, \text{Liu, F.}, \text{Xue, R.} and \text{Duan, Y.} (2026{\natexlab{a}}).
\newblock Cfg-ctrl: Control-based classifier-free diffusion guidance.
\newblock \textit{arXiv preprint arXiv:2603.03281}.

\bibitem[{Wang et~al.(2026{\natexlab{b}})Wang, Cui, Wang, Zhang, Wang, Zhang, Luo, Sun, Li, Wang et~al.}]{wang2024emu3}
\text{Wang, X.}, \text{Cui, Y.}, \text{Wang, J.}, \text{Zhang, F.}, \text{Wang, Y.}, \text{Zhang, X.}, \text{Luo, Z.}, \text{Sun, Q.}, \text{Li, Z.}, \text{Wang, Y.} \text{et~al.} (2026{\natexlab{b}}).
\newblock Multimodal learning with next-token prediction for large multimodal models.
\newblock \textit{Nature} 1--7.

\bibitem[{Wang et~al.(2024)Wang, Dufour, Andreou, Cani, Abrevaya, Picard and Kalogeiton}]{wang2024analysis}
\text{Wang, X.}, \text{Dufour, N.}, \text{Andreou, N.}, \text{Cani, M.-P.}, \text{Abrevaya, V.~F.}, \text{Picard, D.} and \text{Kalogeiton, V.} (2024).
\newblock Analysis of classifier-free guidance weight schedulers.
\newblock \textit{arXiv preprint arXiv:2404.13040}.

\bibitem[{Wen et~al.(2024)Wen, Zhang, Lin and Zhang}]{wen2024sparse}
\text{Wen, K.}, \text{Zhang, H.}, \text{Lin, H.} and \text{Zhang, J.} (2024).
\newblock From sparse dependence to sparse attention: unveiling how chain-of-thought enhances transformer sample efficiency.
\newblock \textit{arXiv preprint arXiv:2410.05459}.

\bibitem[{Wu et~al.(2024)Wu, Chen, Li, Wang and Wei}]{wu2024theoretical}
\text{Wu, Y.}, \text{Chen, M.}, \text{Li, Z.}, \text{Wang, M.} and \text{Wei, Y.} (2024).
\newblock Theoretical insights for diffusion guidance: A case study for gaussian mixture models.
\newblock \textit{arXiv preprint arXiv:2403.01639}.

\bibitem[{Xia et~al.(2025)Xia, Xue, Shen, Yi, Gong and Liu}]{xia2025rectified}
\text{Xia, M.}, \text{Xue, N.}, \text{Shen, Y.}, \text{Yi, R.}, \text{Gong, T.} and \text{Liu, Y.-J.} (2025).
\newblock Rectified diffusion guidance for conditional generation.
\newblock In \textit{Proceedings of the Computer Vision and Pattern Recognition Conference}.

\bibitem[{Xiao et~al.(2018)Xiao, Bahri, Sohl-Dickstein, Schoenholz and Pennington}]{xiao2018dynamical}
\text{Xiao, L.}, \text{Bahri, Y.}, \text{Sohl-Dickstein, J.}, \text{Schoenholz, S.} and \text{Pennington, J.} (2018).
\newblock Dynamical isometry and a mean field theory of cnns: How to train 10,000-layer vanilla convolutional neural networks.
\newblock In \textit{International conference on machine learning}. PMLR.

\bibitem[{Xin et~al.(2025)Xin, Yan, Qin, Li, Liu, Li, Huang, Zhou, Zhang, Zhuo et~al.}]{xin2025lumina}
\text{Xin, Y.}, \text{Yan, J.}, \text{Qin, Q.}, \text{Li, Z.}, \text{Liu, D.}, \text{Li, S.}, \text{Huang, V. S.-J.}, \text{Zhou, Y.}, \text{Zhang, R.}, \text{Zhuo, L.} \text{et~al.} (2025).
\newblock Lumina-mgpt 2.0: Stand-alone autoregressive image modeling.
\newblock \textit{arXiv preprint arXiv:2507.17801}.

\bibitem[{Xiong et~al.(2024)Xiong, Liu, Huang, Wu, Wu, Mu, Yao, Shen, Wan, Huang et~al.}]{xiong2024autoregressive}
\text{Xiong, J.}, \text{Liu, G.}, \text{Huang, L.}, \text{Wu, C.}, \text{Wu, T.}, \text{Mu, Y.}, \text{Yao, Y.}, \text{Shen, H.}, \text{Wan, Z.}, \text{Huang, J.} \text{et~al.} (2024).
\newblock Autoregressive models in vision: A survey.
\newblock \textit{arXiv preprint arXiv:2411.05902}.

\bibitem[{Ye et~al.(2025)Ye, Kevin and Molei}]{ye2025exactly}
\text{Ye, H.}, \text{Kevin, R.} and \text{Molei, T.} (2025).
\newblock What exactly does guidance do in masked discrete diffusion models.
\newblock \textit{arXiv preprint arXiv:2506.10971}.

\bibitem[{Yu et~al.(2022)Yu, Xu, Koh, Luong, Baid, Wang, Vasudevan, Ku, Yang, Ayan et~al.}]{yu2022scaling}
\text{Yu, J.}, \text{Xu, Y.}, \text{Koh, J.~Y.}, \text{Luong, T.}, \text{Baid, G.}, \text{Wang, Z.}, \text{Vasudevan, V.}, \text{Ku, A.}, \text{Yang, Y.}, \text{Ayan, B.~K.} \text{et~al.} (2022).
\newblock Scaling autoregressive models for content-rich text-to-image generation.
\newblock \textit{arXiv preprint arXiv:2206.10789}, \textbf{2} 5.

\bibitem[{Yuan et~al.(2025)Yuan, Li, Zhang, Zhang, Li and Gao}]{yuanmlep}
\text{Yuan, L.}, \text{Li, X.}, \text{Zhang, Y.}, \text{Zhang, J.}, \text{Li, H.} and \text{Gao, X.} (2025).
\newblock Mlep: Multi-granularity local entropy patterns for generalized ai-generated image detection.
\newblock In \textit{The Thirty-ninth Annual Conference on Neural Information Processing Systems}.

\bibitem[{Zhao and Schwing(2026)}]{zhao2026studying}
\text{Zhao, X.} and \text{Schwing, A.} (2026).
\newblock Studying classifier (-free) guidance from a classifier-centric perspective.
\newblock In \textit{Proceedings of the AAAI Conference on Artificial Intelligence}, vol.~40.

\end{thebibliography}

\newpage 

\appendix

\section{Experimental Details}\label{app:exp_detail}

\subsection{Experimental Details and Additional Results of Image Generation in Section~\ref{sec:semantic}}
\label{sec:gen}

In this section, we describe the image generation setup used in our Section~\ref{sec:semantic} and provide more results. 

We generate images using LlamaGen (0.7B) \citep{sun2024autoregressive}, Lumina-mGPT 2.0 (7B) \citep{xin2025lumina}, and Emu3 (8B) \citep{wang2024emu3}. We employ the GenEval benchmark \citep{ghosh2023geneval}, which consists of 553 prompts of varying lengths. Generations are performed via greedy decoding ($\tau = 0$) using three logit configurations: $\text{logit}^\text{cond}$, $\text{logit}^\lambda$, and the difference $\text{logit}^\text{cond} - \text{logit}^\text{uncond}$. For $\text{logit}^\lambda$, we set $\lambda = 10$ across all models to ensure generation quality under greedy settings. Aside from the decoding strategy and guidance scale, all other hyperparameters follow the standard configurations of the respective models. Image resolutions are set to $512 \times 512$ for LlamaGen and Emu3, and $768 \times 768$ for Lumina-mGPT 2.0. During the image generation process, we also save the generated token sequences prior to decoding. By saving these token sequences, we can efficiently retrieve the corresponding logits via prefilling in any subsequent experiments. Our experiments are run on an A100 Nvidia GPU with $40$G memory. The evaluation time used for DPG-Bench in Section~\ref{sec:efficiency} follows the standard setting of this benchmark.

\begin{figure}[t]
    \centering
    \includegraphics[width=0.99\linewidth]{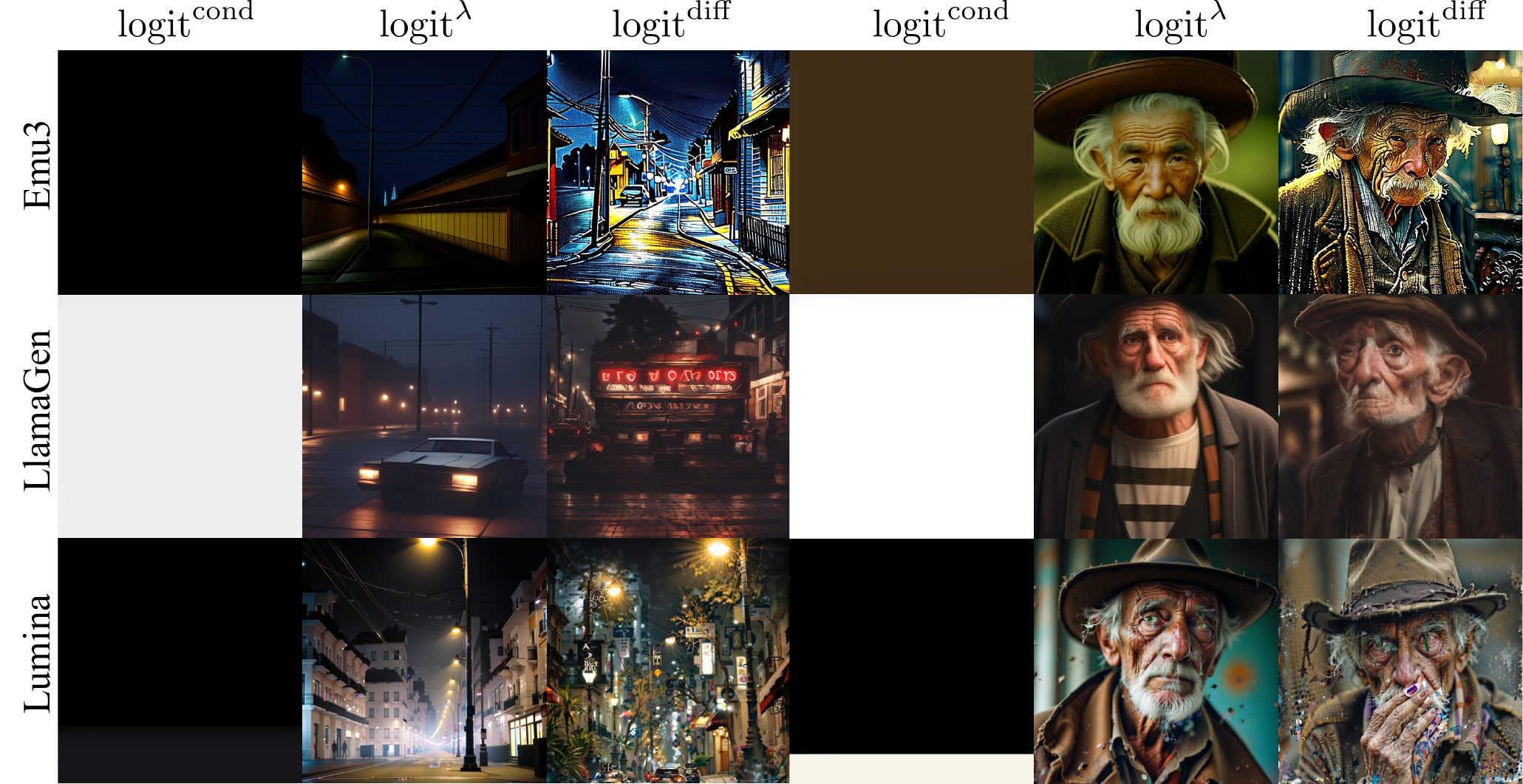}
    \caption{Generated images from $\logit^\cond$, $\logit^\lambda$, and $\logit^\diff = \logit^\cond-\logit^\uncond$, shown from left to right in each grid. All images are generated by greedy decoding ($\tau=0$). Prompts: \textbf{Left:} A street at night is lit up by streetlights. \textbf{Right:} An old man is wearing an odd hat. Images generated from $\logit^\cond$ are nearly pure-color, whereas those from $\logit^\diff$ show richer textures.}
\label{fig:vis-figure-2}
\end{figure}

More generated examples are provided in Figure~\ref{fig:vis-figure-2}, which exhibit similar phenomena to those observed in Figure~\ref{fig:vis-figure} (Section~\ref{sec:semantic}).

\subsection{Experimental Details of Texture Complexity Metrics in Section~\ref{sec:semantic}}

In this section, we provide formal definitions and implementation details for the Texture Complexity metrics analyzed in Section~\ref{sec:semantic}. Let an image be denoted by $I \in \{0, \dots, 255\}^{C \times H \times W}$, where $I_{c,i,j}$ represents the pixel value at location $(i, j)$ in channel $c \in \{1, \dots, C\}$. In our experiments, we use standard RGB images ($C=3$). We evaluate these metrics across images generated by $\logit^\cond$, $\logit^\lambda$ and $\logit^\cond - \logit^\uncond$. We report metrics as averages across images within each of the generation groups established in Section~\ref{sec:gen}.

\paragraph{Total Variation ($\mathrm{TV}$).} To measure the spatial smoothness of an image, we define the Total Variation as the average of normalized $\ell_1$-norm of discrete gradients:
\begin{align*}
\operatorname{TV}(I) = \frac{1}{C} \sum_{c=1}^{C} \left( \frac{1}{(H-1)W} \sum_{i=1}^{H-1} \sum_{j=1}^{W} |\nabla_i I_{c,i,j}| + \frac{1}{H(W-1)} \sum_{i=1}^{H} \sum_{j=1}^{W-1} |\nabla_j I_{c,i,j}| \right),
\end{align*}
where $\nabla_i I_{c,i,j} = I_{c,i+1,j} - I_{c,i,j}$ and $\nabla_j I_{c,i,j} = I_{c,i,j+1} - I_{c,i,j}$ represent horizontal and vertical forward differences, respectively.

\paragraph{Pixel Entropy ($\mathrm{Entropy}$).} This metric captures the global richness of the intensity distribution. For each channel $c$, let $p_c(v) = (HW)^{-1} \sum_{i,j} \mathbb{I}\{I_{c,i,j} = v\}$ be the empirical intensity distribution for $v \in \{0, \dots, 255\}$. The pixel entropy is computed as:
\begin{equation*}
\mathrm{Entropy}(I) = \frac{1}{C} \sum_{c=1}^{C} H(p_c) = \frac{1}{C} \sum_{c=1}^{C} \left( -\sum_{v=0}^{255} p_c(v) \log_2 p_c(v) \right).
\end{equation*}

\paragraph{Orientation Entropy ($\mathrm{OriEntropy}$).}
To characterize structural regularity, we measure the uncertainty of gradient orientations within the image. For each channel $c$, we compute the Sobel gradients $(G_{x, i,j}^{(c)}, G_{y, i,j}^{(c)})$ at position $(i, j)$, which capture the local intensity variations along the horizontal and vertical axes, respectively, and the corresponding magnitude $M_{i,j}^{(c)} = \big((G_{x, i,j}^{(c)})^2 + (G_{y, i,j}^{(c)})^2\big)^{1/2}$. The orientation is defined as $\theta_{i,j}^{(c)} = \operatorname{atan2}(G_{y, i,j}^{(c)}, G_{x, i,j}^{(c)})$, mapped to the undirected range $[0, \pi)$. To focus on salient structures, we only consider the set of pixels $\mathcal{M}_c = \{(i,j) \mid M_{i,j}^{(c)} > \bar{M}^{(c)}\}$, where $\bar{M}^{(c)}$ is the mean magnitude of the channel. We construct a magnitude-weighted orientation distribution $q_c$ over $K=18$ discrete bins $\{\Theta_k\}_{k=1}^K$ partitioning $[0, \pi)$:
\begin{equation*}
q_c(k) = \frac{1}{Z_c} \sum_{(i,j) \in \mathcal{M}_c} M_{i,j}^{(c)} \cdot \mathbb{I}\{\theta_{i,j}^{(c)} \in \Theta_k\}, \quad Z_c = \sum_{(i,j) \in \mathcal{M}_c} M_{i,j}^{(c)}.
\end{equation*}
The $\mathrm{OriEntropy}$ of an image is defined as the average Shannon entropy across channels, which means $\mathrm{OriEntropy}(I) = C^{-1} \sum_{c=1}^C H(q_c)$.

\subsection{Experimental Details of the Similarity Metrics in Section~\ref{sec:overlap}}
\label{sec:overlap-app}

In this section, we provide the experimental details in Section~\ref{sec:overlap} when we measure the similarity between $\logit_t^{\cond}$ and $\logit_t^{\uncond}$ in logit and pixel spaces.

For all models, we collect these logits and their corresponding predicted tokens by prefilling the models with token sequences generated via $\logit^\lambda$ (as described in Section~\ref{sec:gen}). We then compute the difference $\logit^\cond - \logit^\uncond$. The logit-space similarities ($\alpha^\diff, \alpha^\cond, \alpha^\uncond$) are calculated with $k=5$ for all three models. In the pixel space, we decode the predicted tokens to obtain $\beta^\diff, \beta^\cond$, and $\beta^\uncond$. Finally, all metrics are averaged across images generated from the prompts defined in Section~\ref{sec:gen}.

\begin{figure}[t]
    \centering
    \includegraphics[width=0.99\linewidth]{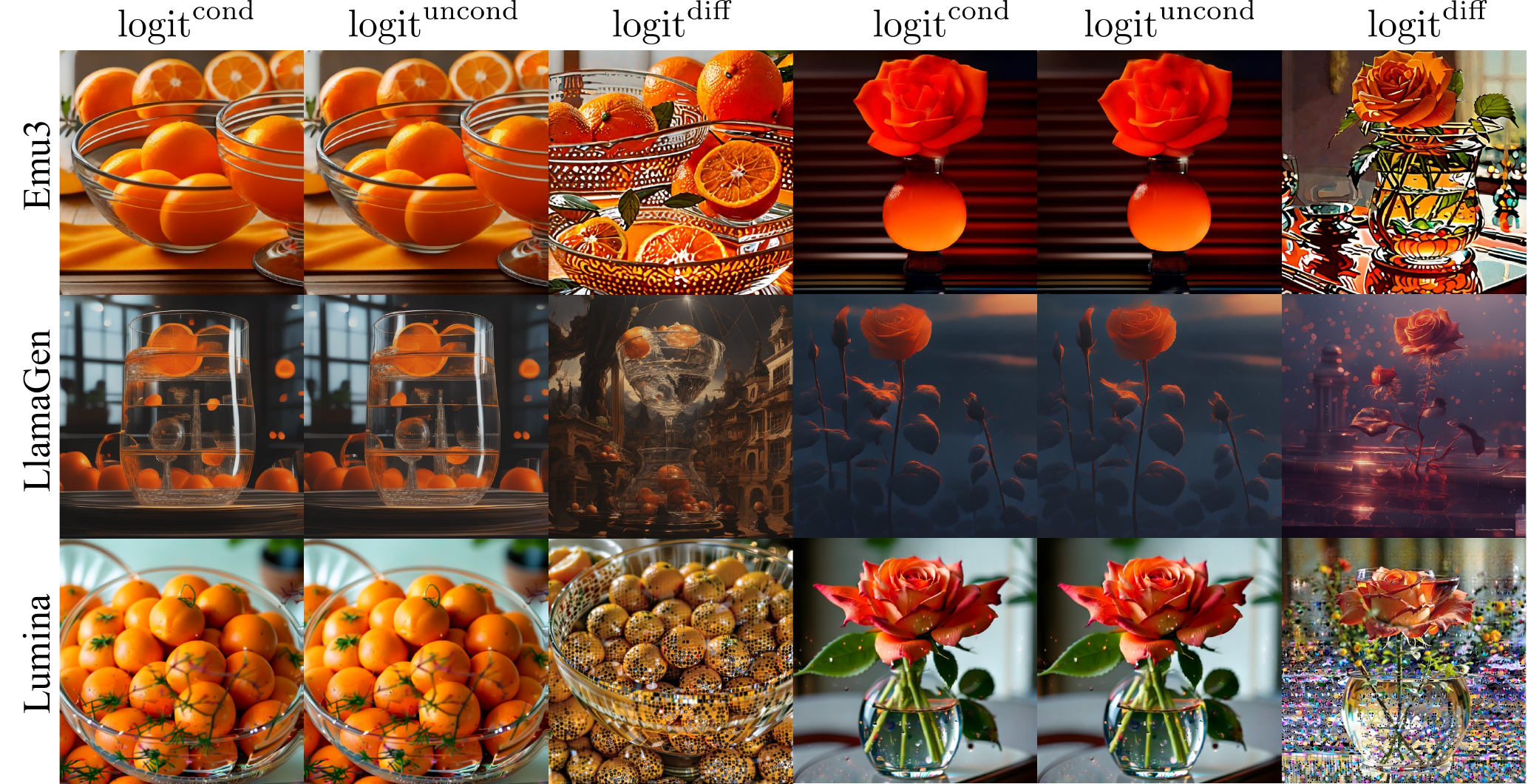}
    \caption{Decoded images from $\logit^\cond$, $\logit^\uncond$, and $\logit^\diff = \logit^\cond-\logit^\uncond$ collected in Section~\ref{sec:overlap-app}, shown from left to right in each grid. Prompts: \textbf{Left:} A glass bowl filled with oranges on a table. \textbf{Right:} An orange-reddish rose in a vase filled with water on top of a table. On the same visual prefixes, the two branches give nearly identical images, while their logit difference emphasizes textures.}
\label{fig:vis-figure-sim-1}
\end{figure}

\begin{figure}[t]
    \centering
    \includegraphics[width=0.99\linewidth]{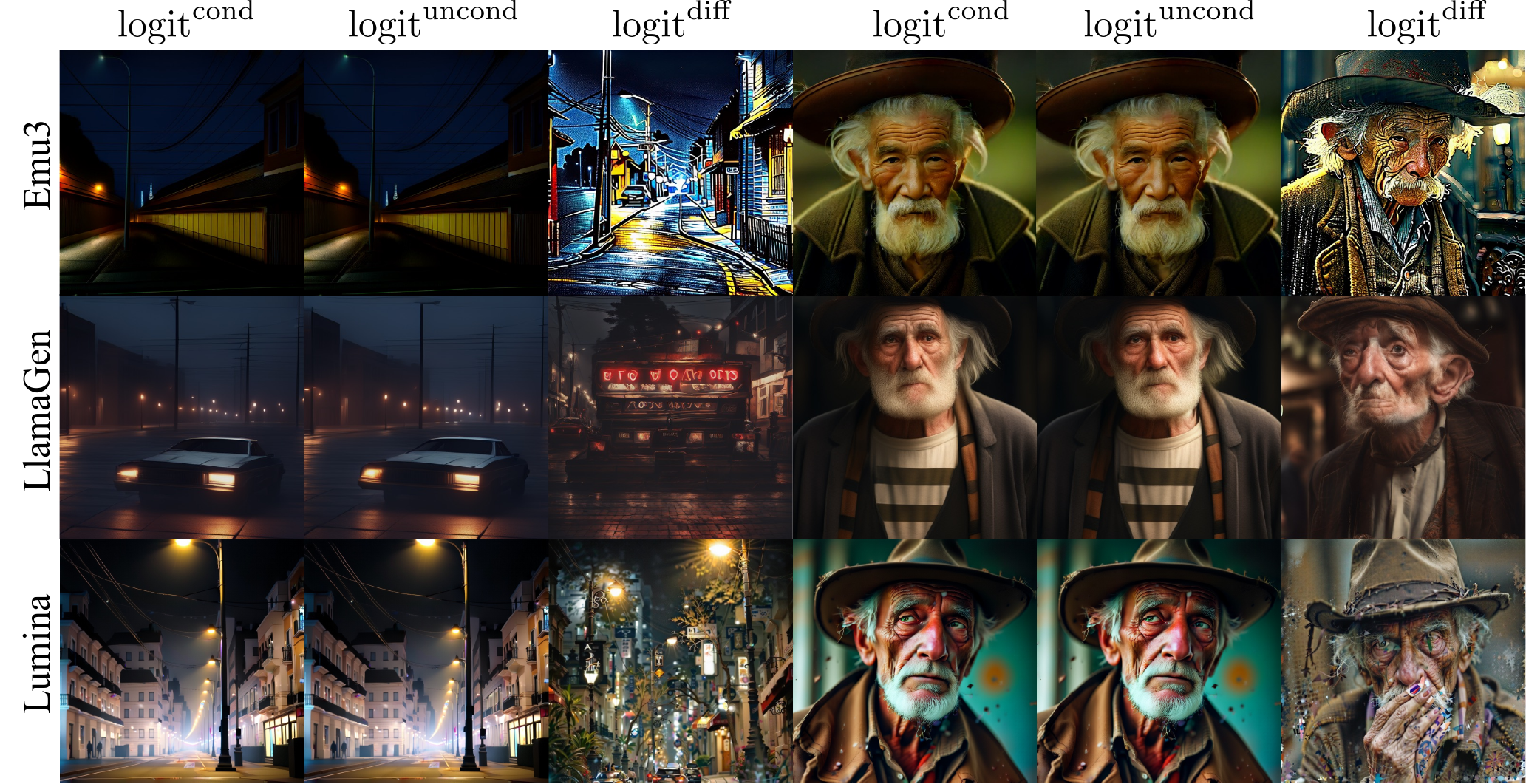}
    \caption{Decoded images from $\logit^\cond$, $\logit^\uncond$, and $\logit^\diff = \logit^\cond-\logit^\uncond$ collected in Section~\ref{sec:overlap-app}, shown from left to right in each grid. Prompts: \textbf{Left:} A street at night is lit up by streetlights. \textbf{Right:} An old man is wearing an odd hat. On the same visual prefixes, the two branches give nearly identical images, while their logit difference emphasizes textures.}
\label{fig:vis-figure-sim-2}
\end{figure}

To provide a qualitative comparison, we present sampled images decoded from $\logit^\cond$ and $\logit^\uncond$ in Figures~\ref{fig:vis-figure-sim-1} and~\ref{fig:vis-figure-sim-2}. We observe that the images generated by $\logit^\cond$ and $\logit^\uncond$ are highly similar across all cases, a finding that aligns with the results in Figure~\ref{fig:prefill-exp} and the theoretical analysis in Section~\ref{sec:overlap}.


\subsection{Experimental Details and Additional Results of Section~\ref{sec:repeat}}
\label{sec:repeat-appendix}

In this section, we provide experimental details of quantifying repetition behavior in Section~\ref{sec:repeat}, and additional results of Emu3 and LlamaGen models.

In our experiments, we first identify high-frequency visual tokens and use them to construct prefixes. For each evaluated model, we pool the tokens generated across all images to compute their occurrence frequency. We define the set of common visual tokens as the top $1\%$ most frequent tokens for each model, which comprises more than $100$ tokens. Then, for each prompt in the GenEval benchmark \cite{ghosh2023geneval}, we randomly sample one such token to construct a prefix covering $20\%$ of the visual token positions, as described in Section~\ref{sec:repeat}. Specifically, the prefix occupies the first $20\%$ of rows in the 2D latent space: 7 rows for LlamaGen ($32\times32$), 13 for Emu3 ($64\times64$), and 20 for Lumina-mGPT ($96\times96$). Following generation conditioned on the prompts and prefixes, we compute $r_t^{\rep}$ at every position and average these values across all non-prefix positions to obtain $r^{\rep}$. Finally, all results are averaged across the generated images.

\begin{figure}[t]
    \centering
    \begin{subfigure}[t]{0.48\linewidth}
        \centering
        \includegraphics[width=\linewidth]{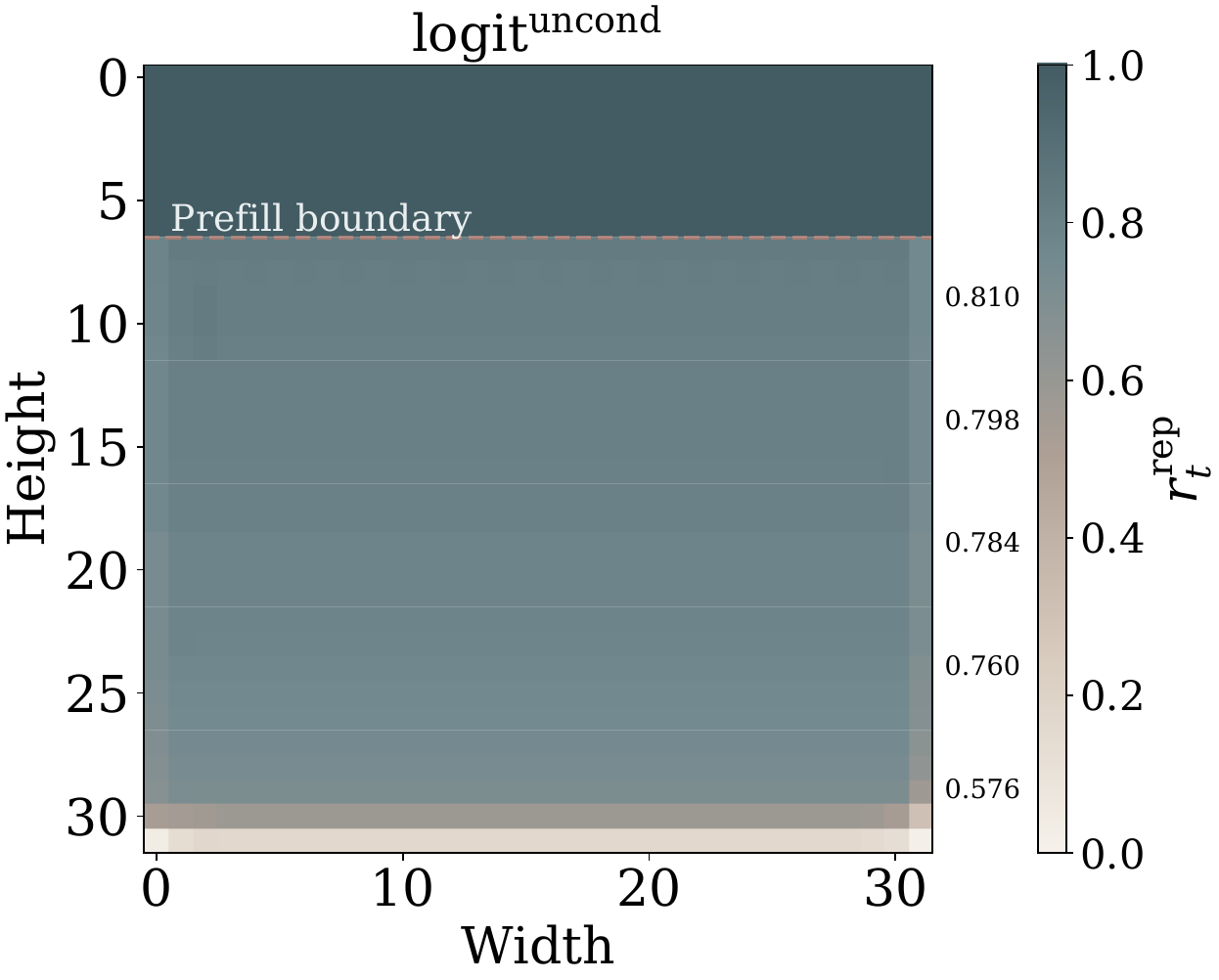}
    \end{subfigure}
    \hfill
    \begin{subfigure}[t]{0.48\linewidth}
        \centering
        \includegraphics[width=\linewidth]{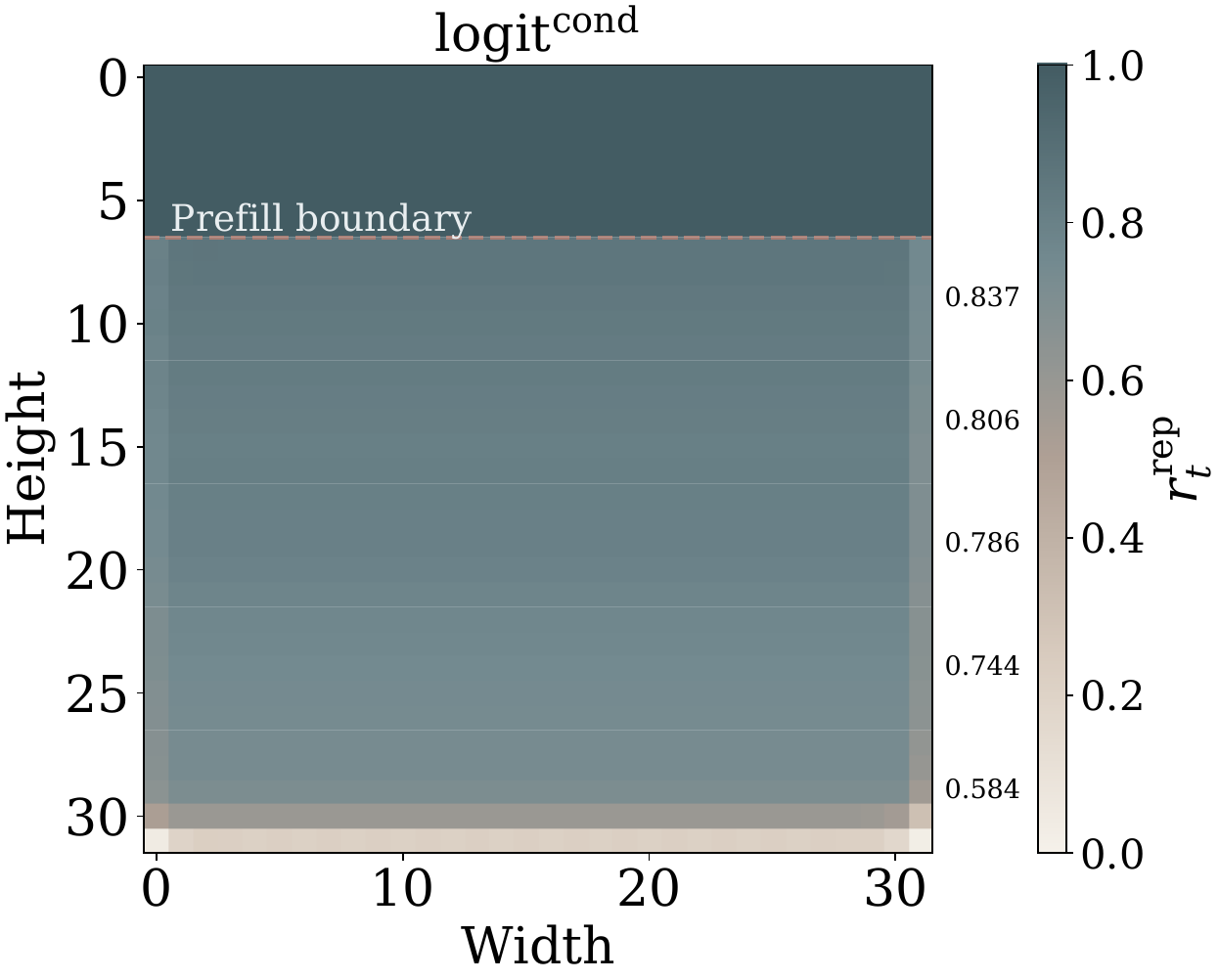}
    \end{subfigure}
    \caption{Position-wise repetition ratio $r_t^\rep$ of LlamaGen, averaged over samples. The ratio measures how often a generated token matches the prescribed prefix token; numbers beside the heatmaps are averages over horizontal bands. Both $\logit^{\cond}$ and $\logit^{\uncond}$ show strong repetition of the prescribed prefix token.}
    \label{fig:add_line_llamagen}
\end{figure}

\begin{figure}[t]
    \centering
    \begin{subfigure}[t]{0.48\linewidth}
        \centering
        \includegraphics[width=\linewidth]{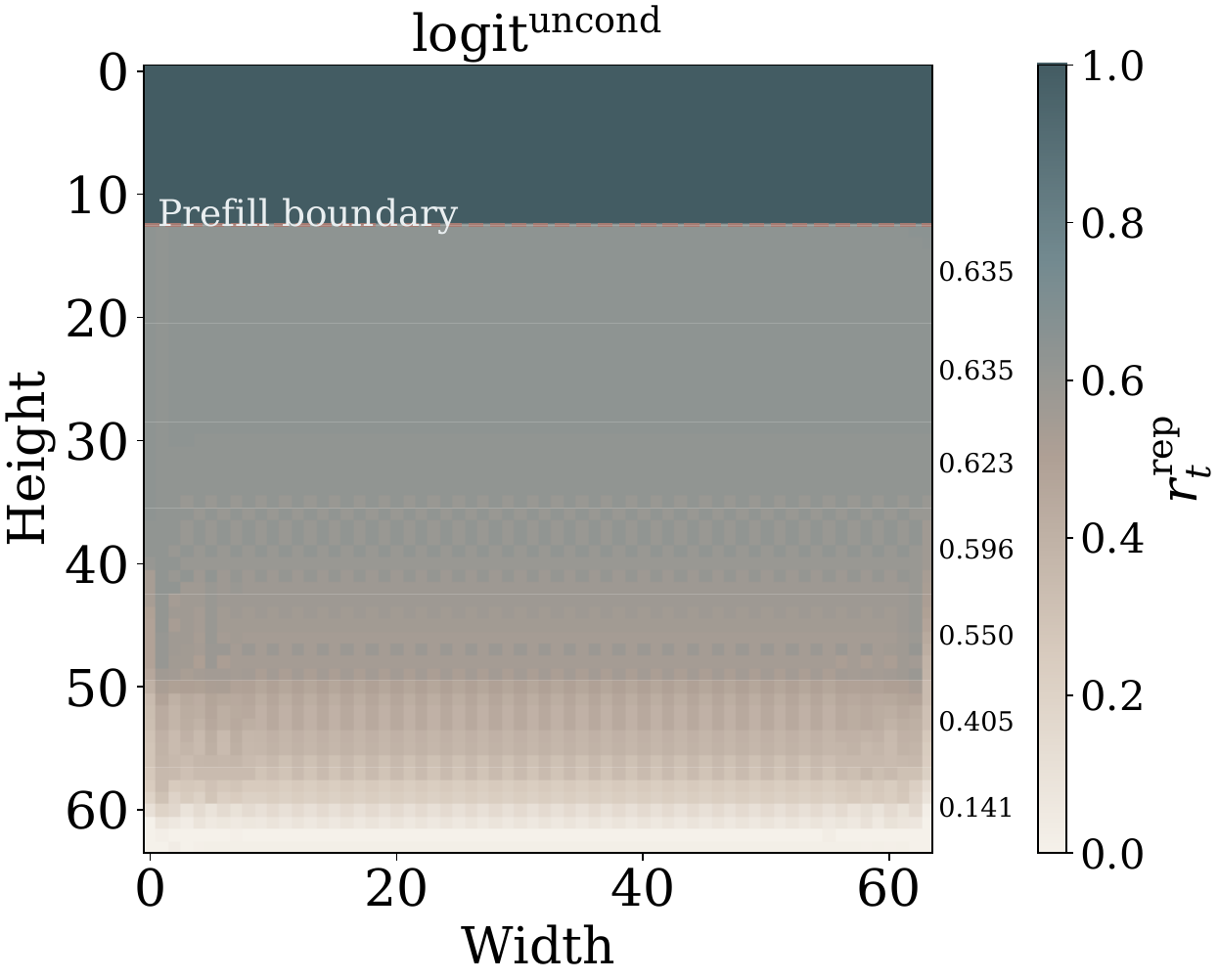}
    \end{subfigure}
    \hfill
    \begin{subfigure}[t]{0.48\linewidth}
        \centering
        \includegraphics[width=\linewidth]{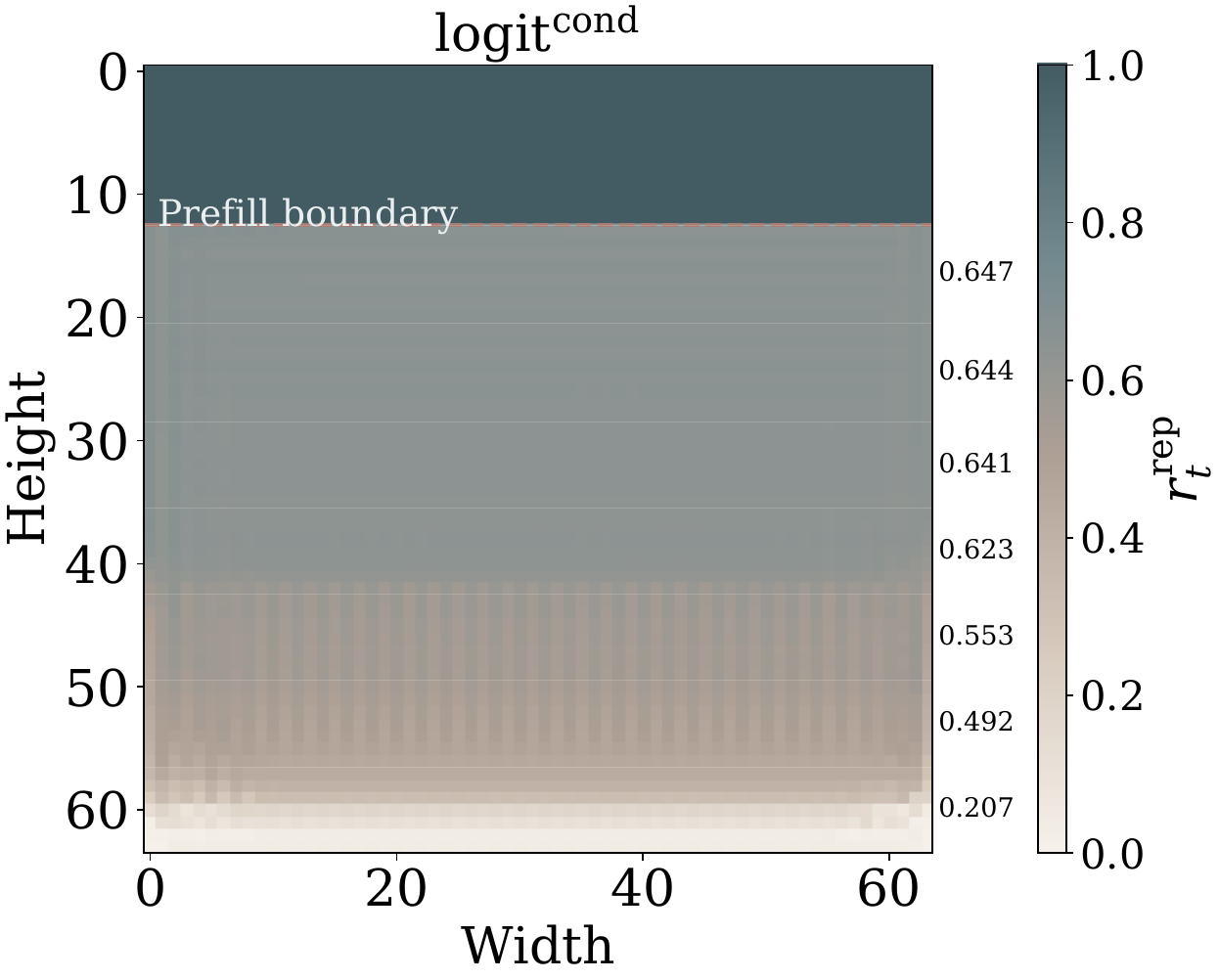}
    \end{subfigure}
    \caption{Position-wise repetition ratio $r_t^\rep$ of Emu3, averaged over samples. The ratio measures how often a generated token matches the prescribed prefix token; numbers beside the heatmaps are averages over horizontal bands. Both $\logit^{\cond}$ and $\logit^{\uncond}$ show strong repetition of the prescribed prefix token.}
    \label{fig:add_line_emu3}
\end{figure}

Figures~\ref{fig:add_line_llamagen} and~\ref{fig:add_line_emu3} present the average $r_t^\rep$ across images for LlamaGen and Emu3. The results are consistent with the findings in Section~\ref{sec:repeat}, demonstrating that tokens generated from $\logit^\cond$ and $\logit^\uncond$ tend to repeat the prefix tokens.

\subsection{Experimental Details and Additional Results of Section~\ref{sec:efficiency}}
\label{sec:efficiency-appendix}
In this section, we provide additional details on AttnReuse. We first present the complete AttnReuse algorithm in Algorithm~\ref{alg:attnreuse}, which describes how attention weights are reused in selected layers while the remaining layers are computed normally. We then summarize the evaluation settings for each model in Table~\ref{tab:eval_settings}. Finally, we present qualitative results in Fig.~\ref{fig:qual_main}, showing that AttnReuse preserves the semantic content and overall visual structure of the generated images with no noticeable degradation compared to standard \ac{cfg}.

In Table~\ref{tab:eval_settings}, we list the evaluation settings for each models. For both \ac{ar} and diffusion models, AttnReuse is applied only to a selected subset of layers during generation to balance efficiency and fidelity. For diffusion models, where computation is additionally distributed across denoising timesteps, we further apply AttnReuse to selected timesteps. 

For the three evaluated \ac{ar} models, each backbone contains 32 Transformer layers. We apply AttnReuse to the middle 16 layers and keep the early and late layers unchanged. This design is motivated by the observation that early layers are more sensitive to low-level token representations, while late layers more directly affect the final prediction logits. Therefore, applying AttnReuse to the middle layers provides a better trade-off between computational efficiency and output fidelity.

For SD3-Medium, which contains 24 Transformer layers, we similarly apply AttnReuse to middle layers. In addition, since the early denoising timesteps largely determine the coarse semantics and global layout of the final image, we keep these timesteps unchanged and apply AttnReuse only to the last $90\%$ of denoising timesteps. This timestep selection reduces computation while preserving the semantic structure established in the early generation stage.

For the stochastic setting $\tau=1$, we follow the officially released inference configuration of each model. For the greedy setting $\tau=0$, directly using the official CFG scale often produces semantically incomplete outputs. We therefore search the CFG scale under $\tau=0$ for each model and report the selected setting used in our evaluation. Across all settings, AttnReuse introduces no additional learnable parameters and is fully compatible with pretrained checkpoints.

\begin{algorithm}[t]
\caption{Attention Weight Reuse (AttnReuse) for \ac{ar} models}
\label{alg:attnreuse}
\begin{algorithmic}[1]

\Require Text-token length $T_{\txt}$; current sequence length $t$; reuse-layer set $\mathcal{L}_{\reuse}$; CFG scale $\lambda$.
\Ensure Guided logits $\logit_t^{\lambda}$.

\For{each decoding step $t$}
    \State Run the conditional branch and cache the conditional attention weights $\{A_{\ell}^{\cond}\}_{\ell\in\mathcal{L}_{\reuse}}$.

    \For{each layer $\ell$}
        \If{$\ell \in \mathcal{L}_{\reuse}$}
        \State Construct a binary key mask $M\in\{0,1\}^{t}$:
        \[
        M_j =
        \begin{cases}
        0, & j \leq T_{\txt},\\
        1, & T_{\txt}<j\leq t.
        \end{cases}
        \]

        \State Remove the text-token contribution from the conditional attention weights:
        \[
        \widetilde{A}_{\ell}^{\reuse}
        =
        A_{\ell}^{\cond}\odot M .
        \]

        \State Renormalize over the key dimension:
        \[
        A_{\ell}^{\reuse}
        =
        \frac{
        \widetilde{A}_{\ell}^{\reuse}
        }{
        \sum_{j=1}^{t}\widetilde{A}_{\ell,:,j}^{\reuse}+\epsilon
        } .
        \]

        \State Approximate the unconditional attention output:
        \[
        O_{\ell}^{\reuse}
        =
        A_{\ell}^{\reuse}V_{\ell}^{\cond}.
        \]

        \State Replace the unconditional attention output at layer $\ell$ with $O_{\ell}^{\reuse}$.
        \Else
            \State Compute the unconditional attention output with standard attention:
            \[
            O_{\ell}^{\uncond}
            =
            \operatorname{Attn}
            \left(
            Q_{\ell}^{\uncond},
            K_{\ell}^{\uncond},
            V_{\ell}^{\uncond}
            \right).
            \]
        \EndIf
    \EndFor

    \State Obtain $\logit_t^{\cond}$ from the conditional branch and approximate $\widehat{\logit}_t^{\uncond}$ from the resulting unconditional branch.

    \State Compute the guided logits:
    \[
    \logit_t^{\lambda}
    =
    \widehat{\logit}_t^{\uncond}
    +
    \lambda
    \left(
    \logit_t^{\cond}
    -
    \widehat{\logit}_t^{\uncond}
    \right).
    \]

    \State Decode the next visual token from $\logit_t^{\lambda}$.

\EndFor

\end{algorithmic}
\end{algorithm}

\begin{table}[t]
\centering
\setlength{\tabcolsep}{6pt}
\begin{tabular}{lccccc}
\toprule
\textbf{Model} 
& \textbf{Type}
& \textbf{$\tau$} 
& \textbf{CFG} 
& \textbf{AttnReuse Layers} 
& \textbf{Timesteps} \\
\midrule

Emu3 
& AR 
& $\{0,1\}$
& $\{14.0,\,7.0\}$ 
& 8--23 
& -- \\

Janus-Pro 7B
& AR 
& $\{0,1\}$ 
& 5.0 
& 8--23 
& -- \\

Lumina-GPT 2.0 
& AR 
& $\{0,1\}$ 
& $\{20.0,\,4.0\}$ 
& 8--23 
& -- \\

\midrule

Stable Diffusion 3 
& Diffusion 
& -- 
& 7.0 
& 9--17 
& Last $90\%$ \\

\bottomrule
\end{tabular}
\vspace{0.5em}
\caption{
Evaluation settings across models. 
For \ac{ar} models, AttnReuse is applied to a subset of attention layers (Layers 8--23). 
For diffusion models, it is applied to selected layers (Layers 9--17) and across the last $90\%$ of timesteps.
}
\label{tab:eval_settings}
\end{table}

\begin{figure*}[t]
    \centering
    \setlength{\tabcolsep}{2pt}
    
    \begin{subfigure}{0.90\textwidth}
        \centering
        \includegraphics[width=\linewidth]{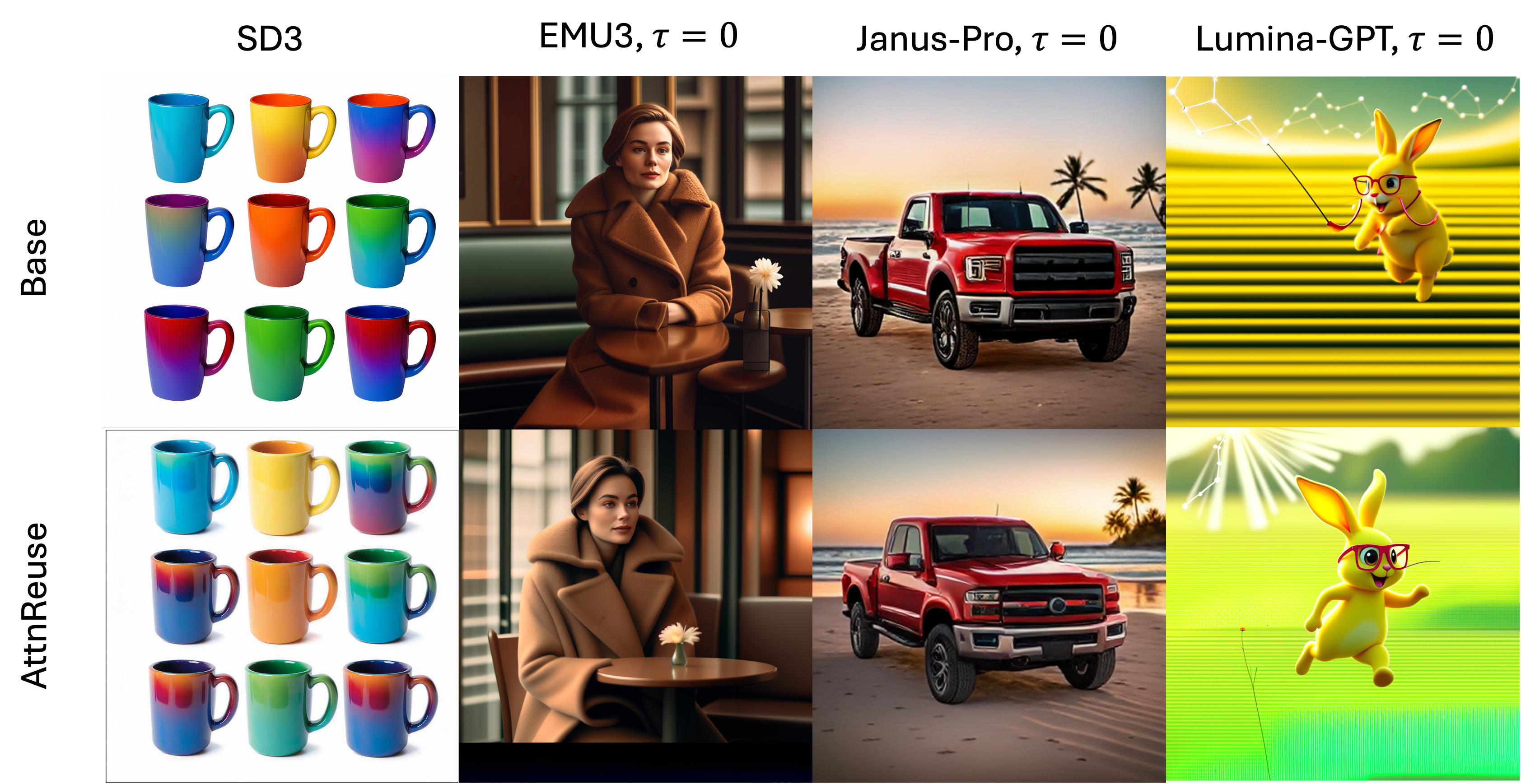}
        \label{fig:qual_tau0}
    \end{subfigure}
    \hfill
    \begin{subfigure}{0.90\textwidth}
        \centering
        \includegraphics[width=\linewidth]{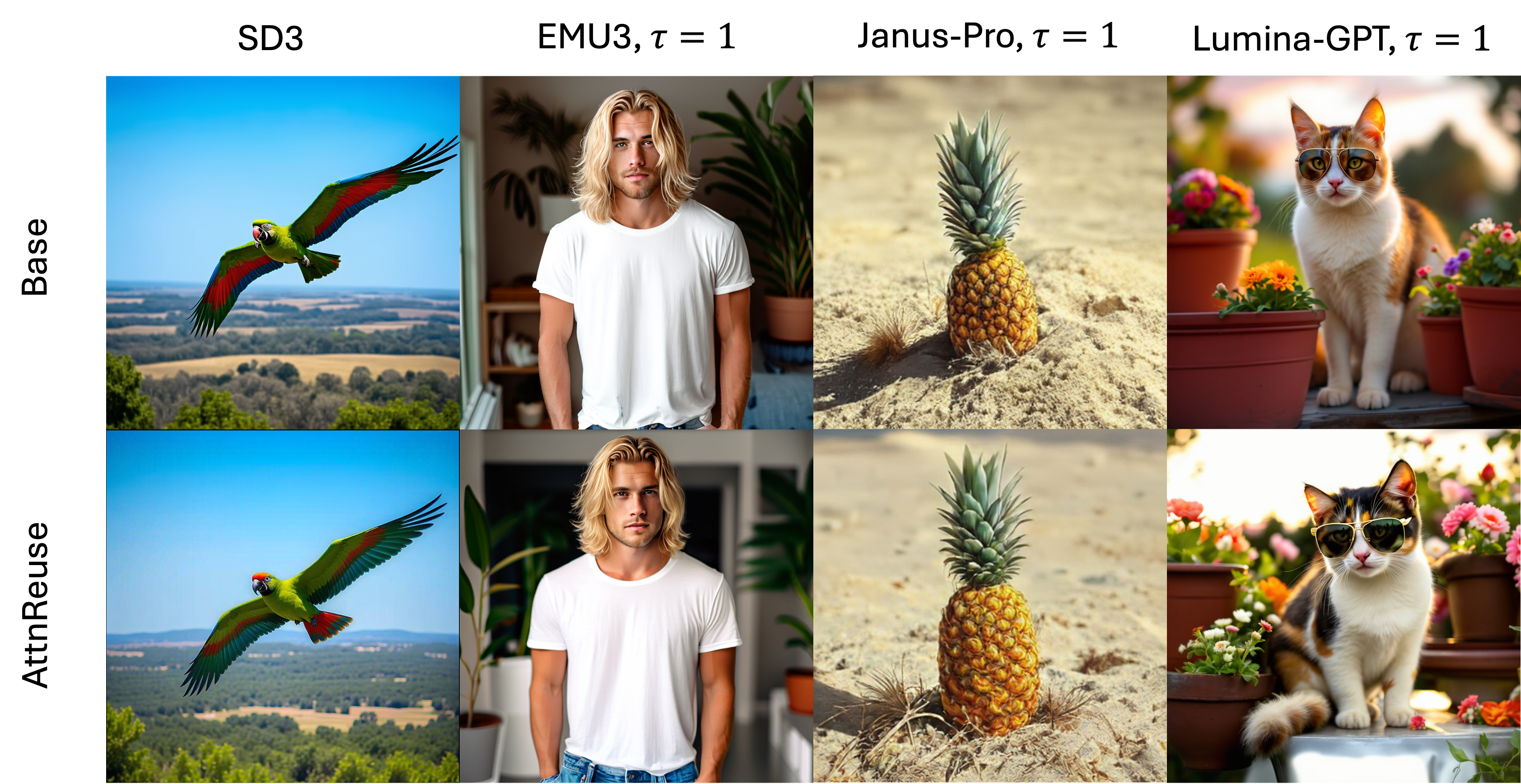}
        \label{fig:qual_tau1}
    \end{subfigure}

    \caption{
    Qualitative comparison between \textbf{Base} and \textbf{AttnReuse} across different models.
    AttnReuse preserves structural and semantic fidelity under both greedy ($\tau=0$) and stochastic ($\tau=1$) decoding, while reducing attention computation.
    }
    \label{fig:qual_main}
\end{figure*}

\section{Detailed Training Algorithm in Section~\ref{sec:theory}}\label{app:algo}
We train the model by (two-step) population gradient descent on the data distribution defined above, as specified in Algorithm~\ref{alg:training}. 
The population cross-entropy loss is
\begin{align*}
\cL_{\vis}(\theta)
=\bbE_{(z_1, x_{1:T_{\vis}})\sim P_{\text{data}}}\Bigg[
T_{\vis}^{-1}\sum_{t=0}^{T_{\vis}-1} 
-\log P_\theta(x_{t+1} \mid z_1, x_{1:t})
\Bigg].
\end{align*}
To train the parameters $\theta=(W_V,W_{QK})$, we initialize them as
$\theta^{(0)}=(0,0)$. We first update $W_V$ by one gradient step, and then
optimize $W_{QK}$ for one step while keeping $W_V$ fixed. This two-stage
training protocol follows prior analyses of transformer training dynamics
and is used for theoretical tractability~\citep{wen2024sparse,nichani2024transformers}.

\begin{algorithm}[H]
\caption{Two-step population gradient descent}
\label{alg:training}
\begin{algorithmic}[1]
\Require Transformer $P_\theta$; learning rates $\eta_V,\eta_{QK}>0$;
         population loss $\mathcal{L}_{\vis}(\cdot)$.
\State $\theta^{(0)}\gets(0,\,0)$
\State $W_V^{(1)}\gets-\eta_V\,\nabla_{W_V}\mathcal{L}_{\vis}(\theta^{(0)})$;
       \quad $\theta^{(1)}\gets\bigl(W_V^{(1)},\,0\bigr)$
\State $W_{QK}^{(1)}\gets-\eta_{QK}\,\nabla_{W_{QK}}\mathcal{L}_{\vis}(\theta^{(1)})$;
       \quad $\theta^{(2)}\gets\bigl(W_V^{(1)},\,W_{QK}^{(1)}\bigr)$
\State \Return $\theta^{(2)}$
\end{algorithmic}
\end{algorithm}
\section{Technical Preliminaries for the Proof}
\label{Sectionpre-proof}

In this section, we introduce the notation and technical results that will be used in the proof of the main theorem. 
We clarify the notation and make the intermediate outputs of the model explicit to facilitate the subsequent analysis.
We then derive explicit expressions for the gradients with respect to the model parameters, which will play a central role in the analysis.

\paragraph{Transformer Computation Decomposition}
To analyze the computational flow of the Transformer defined in Section~\ref{sec:theory}, we first decompose the attention mechanism under the given parameter constraints. For notational simplicity, we denote the initial text token $z_1$ as $x_0$ and fix its position at $t=0$. Let $s_{tl}$ and $\alpha_{tl}$ denote the pre-softmax attention score and the corresponding weight from query position $t$ to key position $l$. The attention output $\tilde{o}_t$ and the resulting logit $\text{logit}_t$ at position $t$ are computed as follows:
\begin{equation*}
s_{tl} = h_t^\top \widetilde{W}_{QK} h_l,\quad
\alpha_{tl} = \frac{\exp(s_{tl})}{\sum_{s=0}^{t}\exp(s_{ts})},\quad
\tilde{o}_t = \sum_{l=0}^{t}\alpha_{tl}\,\widetilde{W}_V h_l,\quad \logit_t = \widetilde{W}_{\lm}^\top\tilde{o}_t,
\end{equation*}
where for simplicity, we denote $\logit_t$ for $\logit_t^\cond$. Because the parameterization is restricted to
\begin{align*}
\widetilde{W}_{QK} = \begin{bmatrix} 0 & 0 \\ W_{QK} & 0 \end{bmatrix}, \quad
\widetilde{W}_V = \begin{bmatrix} W_V & 0 \\ 0 & 0 \end{bmatrix}, \quad
\widetilde{W}_{\lm} = \begin{bmatrix} W_{\lm} \\ 0 \end{bmatrix},
\end{align*}
and input satisfies $h_t = [f_{x_t}^\top, e_t^\top]^\top$, we can simplify as
\begin{align*}
s_{tl}= e_t^\top W_{QK} f_{x_l},\quad
\tilde{o}_t =
\begin{bmatrix}
o_t\\
0
\end{bmatrix},\quad o_t = \sum_{l=0}^{t}\alpha_{tl} W_V f_{x_l}, \quad \logit_t = W_{\lm}^\top o_t.
\end{align*}

For notational convenience, we use the superscript $\uc$ to denote variables associated with the unconditional case ($\uncond$). Correspondingly, the attention weights $\alpha_{tl}^\uc$, attention outputs $\tilde{o}_t^\uc$, and logits $\logit_t^\uc$ for the unconditional model (simplified version of $\logit_t^\uncond$) are defined as:
\begin{align*}
\alpha_{tl}^\uc = \frac{\exp(s_{tl})}{\sum_{s=1}^{t}\exp(s_{ts})},\quad
\tilde{o}_t^\uc = \sum_{l=1}^{t}\alpha_{tl}^\uc\,\widetilde{W}_V h_l,\quad \logit_t^\uc = \widetilde{W}_{\lm}^\top\tilde{o}_t^\uc.
\end{align*}

Similarly, by exploiting the previously defined block-matrix parameterization, these expressions can be simplified in terms of the token-level representations:
\begin{align*}
\tilde{o}_t^\uc =
\begin{bmatrix}
o_t^\uc\\
0
\end{bmatrix},\quad o_t^\uc = \sum_{l=1}^{t}\alpha_{tl}^\uc W_V f_{x_l}, \quad \logit_t^\uc = W_{\lm}^\top o_t^\uc
\end{align*}

For clarity, we denote the $i$-th element of $\logit_t\in\RR^{K+J}$ as $\logit_t[i]$, which also applies for $\logit_t^\uc$ and $\logit_t^\lambda$.

\paragraph{Equivalent Form of CFG on Output}

Given that the output matrix $W_{\lm}$ is fixed, we can establish an operational equivalence between applying CFG at the logit level and the attention output level. At position $t$, the CFG-modified logits with guidance scale $\lambda$ are formulated as:
\begin{align}
\label{eq:equivalence-cfg}
\logit_t^{\lambda}&=\logit_t+\lambda\,(\logit_t-\logit_t^{\uc}) = W_{\lm}^\top \big(o_t + \lambda (o_t - o_t^\uc)\big),
\end{align}
where the first equation follows from the definition in Eqn.~\eqref{eq:cfg}, and the second relies on the simplified representation introduced previously.

\paragraph{Compact Notation for Sequences}

To simplify the exposition, we adopt the superscript $\times L$ to denote a token repeated $L$ times. For instance, a representative sample from our data distribution is denoted as $\big(K+J+k_1, (K+j_1)^{\times \tau}, (k_1)^{\times L}, (K+j_1)^{\times (T - \tau - L)}\big)$. This sequence consists of an initial text token $K+J+k_1$, followed by a block of $\tau$ background tokens $K+j_1$, a segment of $L$ feature tokens $k_1$, and a final padding of $T - \tau - L$ background tokens.

\paragraph{Per-sequence loss and Population loss} To analyze the training dynamics, we derive a closed-form expression for the population loss. We begin by defining the loss for an individual sequence $(x_0, x_1, \dots, x_T) \in \cI_{k,j}$, where $k \in [K]$ and $j \in [J]$ denote the visual feature and background classes, respectively. The cross-entropy loss for this sequence is given by:
\begin{align}\label{eq:loss-kj}
\cL(k,j;\theta)= -\sum_{t=1}^{T - 1}\log P_\theta(x_{t+1} | x_0, x_1, \cdots x_t), \quad \text{where}\quad (x_0, x_1, \cdots, x_T) = I_{k, j}.
\end{align}
The population loss under the data distribution $P_{\text{data}}$ (defined in Section~\ref{sec:theory}) is then formulated as the expected loss over all sequences:
\begin{align}
\label{eq:pop-loss}
\cL_{\vis}(\theta)= \bbE_{(z_1, x_{1:T})\sim P_{\text{data}}}\Bigg[
\sum_{t=1}^{T-1} 
-\log P_\theta(x_{t+1} \mid x_0, x_{1:t})\Bigg] = \sum_{k=1}^{K}p_k\sum_{j=1}^{J}q_j\,\cL(k,j;\theta),
\end{align}
where the first equation is the definition of population loss in Section~\ref{sec:theory}, and the second equation holds because the probability of sampling visual feature class $k$ is $p_k$, and sampling visual background class $j$ is $q_j$.

\subsection{Gradient Formulas}
\label{sec:grad-formulas}

To facilitate the subsequent analysis of gradient dynamics, we define several auxiliary quantities that characterize the prediction error and its projection back into the hidden representation space. 

Let $P_t = P_\theta(x_{t+1} | x_{0:t}) = \sm(\logit_t)\in \RR^{K+J}$ be the predicted probability distribution over the vocabulary at position $t$. We define the prediction residual $\delta_t$ and its pullback $\tilde{g}_t$ through the language modeling head $\widetilde{W}_{\text{LM}}$ as follows:
\begin{align*}
y_t &= \bone_{x_{t+1}} \in \RR^{K+J}, \quad
\delta_t = P_t - y_t \in \RR^{K+J}, \quad
\tilde{g}_t = \widetilde{W}_{\lm} \delta_t 
=
\begin{bmatrix}
g_t \\ 
0 
\end{bmatrix}  \in \RR^D, \quad g_t = W_{\lm} \delta_t\in \RR^d,
\end{align*}
where $\bone_i$ is the $i$-th standard basis vector in $\RR^{K+J}$, and here $\bone_{x_{t+1}}$ acts as indicator vector for class $x_{t+1}$. The structure of $\tilde{g}_t$ comes from the simplified structure of parameters.

\begin{lemma}[Single-sequence gradients]\label{lem:gradients}
Fix any token sequence $(x_0,\dots,x_T)$ and parameters $\theta=(W_{V},W_{QK})$.
The gradients of the single-sequence loss $\cL$ in Eqn.~\eqref{eq:loss-kj} with respect to $W_V$ and $W_{QK}$ are
\begin{align}
\label{eq:grad-WV}
\nabla_{W_V}\cL
&=
\sum_{t=1}^{T-1}\sum_{l=0}^{t}
\alpha_{tl}\,g_t\,f_{x_l}^\top,\\[4pt]
\label{eq:grad-WQK}
\nabla_{W_{QK}}\cL
&=
\sum_{t=1}^{T-1}\sum_{l=0}^{t}\alpha_{tl}\big[g_t^\top(W_V f_{x_l} - o_t)\big]\,
e_t f_{x_l}^\top.
\end{align}
\end{lemma}

\begin{proof}
\textbf{Gradient w.r.t.\ $W_V$.}
Write the per-step loss as $\cL_t=-\log P_t(x_{t+1} | x_{0:t})$. Because $P_t = \sm(\logit_t) = \sm(W_{\lm}^\top o_t)$. Applying the standard gradient property of the softmax cross-entropy loss and the chain rule via $\logit_t = W_{\lm}^\top o_t$, we obtain
\begin{align*}
  \frac{\partial\cL_t}{\partial \logit_t} = P_t - y_t = \delta_t,\quad \frac{\partial\cL_t}{\partial o_t} = W_{\lm}\delta_t = g_t.
\end{align*}
Since $o_t=\sum_{l=0}^t\alpha_{tl}W_V f_{x_l}$ and $\alpha_{tl}$ does not depend on $W_V$,
differentiating entry-wise gives
\begin{align*}
  \frac{\partial o_t}{\partial(W_V)_{ab}}
  = \sum_{l=0}^t\alpha_{tl}\,(f_{x_l})_b\,\bone_a,
\end{align*}
where $\bone_a$ is the $a$-th standard basis vector in $\RR^{K+J}$. Applying the chain rule from $\frac{\partial\cL_t}{\partial o_t}, \frac{\partial o_t}{\partial(W_V)_{ab}}$ and
collecting into matrix form, we have
\begin{align*}
  \frac{\partial\cL_t}{\partial W_V}
  = g_t \big(\sum_{l=0}^t\alpha_{tl}\,f_{x_l}\big)^{\!\top}
  = \sum_{l=0}^t\alpha_{tl}\,g_t\,f_{x_l}^\top.
\end{align*}

Summing over $t=1,\dots,T-1$, we have
\begin{align*}
\nabla_{W_V}\cL
&=
\sum_{t=1}^{T-1}\frac{\partial\cL_t}{\partial W_V}
=
\sum_{t=1}^{T-1}\sum_{l=0}^{t}
\alpha_{tl}\,g_t\,f_{x_l}^\top.
\end{align*}

\textbf{Gradient w.r.t.\ $W_{QK}$.}
We decompose the chain via the intermediate variables
$\alpha_{tj}$ and $s_{tl}$.

Since $o_t=\sum_{l=0}^t\alpha_{tl}W_V f_{x_l}$, varying $\alpha_{tj}$ while keeping all other attention weights fixed gives $\partial o_t/\partial\alpha_{tj}=W_V f_{x_j}$.
Combined with $\partial\cL_t/\partial o_t= g_t$, by chain rule, we have
\begin{align*}
  \frac{\partial\cL_t}{\partial\alpha_{tj}}
  = g_t^\top W_V f_{x_j}.
\end{align*}

From the softmax formula $\alpha_{tj}=\exp(s_{tj})/\!\sum_{s'=0}^t\exp(s_{ts'})$,
standard differentiation yields the Jacobian
\begin{align*}
  \frac{\partial\alpha_{tj}}{\partial s_{tl}}
  = \alpha_{tj} \big(\bbI\{j=l\}-\alpha_{tl}\big).
\end{align*}

By the chain rule over $j$, we have
\begin{align*}
\frac{\partial\cL_t}{\partial s_{tl}}
&= \sum_{j=0}^t\frac{\partial\cL_t}{\partial\alpha_{tj}}
   \cdot\frac{\partial\alpha_{tj}}{\partial s_{tl}}\\
&= \sum_{j=0}^t\big(g_t^\top W_V f_{x_j}\big)\,\alpha_{tj}\!\left(\bbI\{j=l\}-\alpha_{tl}\right)\\
&= \alpha_{tl}\,g_t^\top W_V f_{x_l}
   -\alpha_{tl}\sum_{j=0}^t\alpha_{tj}\,g_t^\top W_V f_{x_j}\\
&= \alpha_{tl}\,g_t^\top(W_V f_{x_l} - o_t),
\end{align*}
where the second and third equalities follow from previous results of $\frac{\partial\cL_t}{\partial\alpha_{tj}}, \frac{\partial\alpha_{tj}}{\partial s_{tl}}$ and direct expansion, the last step invokes the structural form of $o_t = \sum_{j=0}^t \alpha_{tj}g_t^\top W_V f_{x_j}$.

From the simplification, we have $s_{tl}=e_t^\top W_{QK} f_{x_l}$, the differentiation gives $\partial s_{tl}/\partial W_{QK}=e_t f_{x_l}^\top$. Combining the previous results and applying the chain rule, we obtain
\begin{align*}
  \frac{\partial\cL_t}{\partial W_{QK}}
  = \sum_{l=0}^t\frac{\partial\cL_t}{\partial s_{tl}}\cdot e_t f_{x_l}^\top
  = \sum_{l=0}^t\alpha_{tl}\, g_t^\top(W_V f_{x_l} - o_t)\,e_t f_{x_l}^\top,
\end{align*}
where the first equation comes from chain rule expansion, and the second equation follow from previous results of $\frac{\partial\cL_t}{\partial s_{tl}}, \frac{\partial\cL_t}{\partial W_{QK}}$.

Summing over $t=1,\dots,T-1$, we have
\begin{align*}
\nabla_{W_{QK}}\cL
&=
\sum_{t=1}^{T-1}\frac{\partial\cL_t}{\partial W_{QK}}
=
\sum_{t=1}^{T-1}\sum_{l=0}^{t}\alpha_{tl}\big[g_t^\top(W_V f_{x_l} - o_t)\big]\,
e_t f_{x_l}^\top.
\end{align*}
Thus, we conclude the proof of Lemma~\ref{lem:gradients}.
\end{proof}

\section{Proof of Theorem~\ref{thm:main}}
\label{sec:proof}
In the proof of Theorem~\ref{thm:main}, we follow the training protocol in
Appendix~\ref{app:algo} and analyze the approximation errors of
$\theta^{(1)}$ and $\theta^{(2)}$ sequentially. For ease of notation, we will write $T_\vis$ as $T$ in the proof. Before the proof, we will first state the formal version of Theorem~\ref{thm:main} as follows.

\begin{theorem}
\label{thm:formal}
Under Assumptions~\ref{asm:length-f} and ~\ref{asm:ortho}, suppose the following parameter regime holds as $T_\vis\to\infty$: the start positions of features satisfy $\tau_k = \tau=\Theta(T_\vis^\beta)$ for all $k\in[K]$, and some fixed $\beta\in(1/2,1)$; the number of visual background tokens satisfies $J=\bigl\lfloor\tau\log T_\vis/cL\bigr\rfloor$ for a constant $c$ with $1<c< 1 / (2-2\beta)$; the number of visual feature tokens satisfies $K=\Theta(1)$; the distribution of images satisfies $p_k=\Theta(1/K)$,\; $q_j=1/J$; the learning rates satisfy $\eta_V=\Theta(1)$,\; $\eta_{QK}=\Theta(T_\vis^\beta)$.
Let $\theta^{(2)}=(W_V^{(1)},W_{QK}^{(1)})$ be the output of
Algorithm~\ref{alg:training} applied to the Transformer model $P_\theta$
Then the following three statements hold.
\begin{enumerate}
  \item \textup{(Training loss.)}
        The population loss is asymptotically small, with
        \begin{align*}
          \mathcal{L}_{\vis}(\theta^{(2)})
          = \Theta \Big(\frac{(\log T_\vis)^2}{T_\vis}\Big).
        \end{align*}
  \item \textup{(Conditional generation fails and repeats.)}
        With guidance scale $\lambda=0$ (pure conditional generation),
        the generated sequence is not a valid image and repeats backgrounds for any query:
        \begin{align*}
          \hat{I}^{0}_{k,j}(\theta^{(2)})=(K+j, \cdots, K+j)\neq I_{k,j},
          \qquad\text{for all }k\in[K],\;j\in[J].
        \end{align*}
  \item \textup{(CFG recovers valid images.)}
        There exists $\lambda^*=\Theta(1)$ with $\lambda^*>0$ such that
        CFG at guidance scale $\lambda^*$ generates a valid image for every query:
        \begin{align*}
          \hat{I}^{\lambda^*}_{k,j}(\theta^{(2)})=I_{k,j},
          \qquad\text{for all }k\in[K],\;j\in[J].
        \end{align*}
    \item \textup{(Alignment of conditional and unconditional prediction.)}
        The conditional model and unconditional model has the same prediction when generating image with guidance scale $\lambda^*$, which means
        \begin{align*}
          \arg\max_{x\in\calX_\vis} \logit_t^\cond = \arg\max_{x\in\calX_\vis} \logit_t^\uncond
        \end{align*}
\end{enumerate}
\end{theorem}

\textbf{Proof Roadmap.} We begin by outlining the roadmap for the proof of Theorem~\ref{thm:formal}. Following the two-stage training procedure in Algorithm~\ref{alg:training}, we first analyze the update of $W_V^{(1)}$ and then characterize the subsequent update of $W_{QK}^{(1)}$ based on the first-stage results. We then demonstrate that the training loss is asymptotically minimized. Finally, we establish the model's generation behavior: while the model successfully generates target images using CFG with a suitable guidance scale, it fails when restricted to conditional decoding. We also prove that the predictions of the conditional and unconditional branches remain aligned.

Steps~1--4 establish the first-stage parameter estimates: Step~1 computes the population gradient of $W_V$ at initialization; Step~2 derives the asymptotic order of its coefficients; Step~3 verifies a pivotal asymptotic property induced by the parameter regime; and Step~4 converts these results into an explicit one-step approximation for $W_V^{(1)}$, analyzing how embeddings are projected into the value space. Step~5 examines the logits and probabilities induced by $W_V^{(1)}$, laying the groundwork for the second stage. Steps~6--7 analyze the population gradient and one-step update of $W_{QK}$, yielding an explicit approximation for $W_{QK}^{(1)}$ and characterizing the computation of position-feature correlations. Step~8 leverages these estimates to analyze CFG behavior across different time regimes and defines the valid range for the guidance scale. Step~9 proves that the training loss is asymptotically small. Steps~10--11 establish our two generation claims: CFG-based decoding correctly generates the target image, whereas pure conditional decoding collapses into repetitive background patterns. Step~12 proves the alignment between conditional and unconditional predictions. Finally, Step~13 synthesizes these intermediate results to complete the proof of Theorem~\ref{thm:formal}.

\subsection{Step~1: Population Gradient of $W_V$ at Initialization}
\label{sec:grad-wv-pop}

To compute the first update in Algorithm~\ref{alg:training}, we first derive $\nabla_{W_V}\mathcal{L}_{\vis}(\theta^{(0)})$ at initialization, where $\theta^{(0)} = (W_V^{(0)}, W_{QK}^{(0)}) = (0, 0)$.

\begin{lemma}[Population gradient of $W_V$ at initialization]
\label{lem:grad-wv}
At initialization of Algorithm~\ref{alg:training}, where $\theta^{(0)} = (W_V^{(0)}, W_{QK}^{(0)}) = (0, 0)$, the population gradient is
\begin{align*}
\nabla_{W_V}\mathcal{L}_{\vis}(\theta^{(0)})
=
\sum_{k=1}^K p_k\sum_{j=1}^J q_j \nabla_{W_V}\cL(k,j;\theta^{(0)}),
\end{align*}

where for each $(k,j)$,
\begin{equation}
\label{eq:Gtt-decomposition}
\begin{aligned}
\nabla_{W_V}\cL(k,j;\theta^{(0)})
&=
\bar f\Big(A_{K+J+k} f_{K+J+k}^\top+A_k f_k^\top+A_{K+j} f_{K+j}^\top\Big)\\
&\quad
-f_{K+j}\Big(B_{K+J+k} f_{K+J+k}^\top+B_k f_k^\top+B_{K+j} f_{K+j}^\top\Big)\\
&\quad
-f_k\Big(C_{K+J+k} f_{K+J+k}^\top+C_k f_k^\top+C_{K+j} f_{K+j}^\top\Big),
\end{aligned}
\end{equation}
for some positive constant $A_{K+J+k}, A_k, A_{K+j}, B_{K+J+k}, B_k, B_{K+j}, C_{K+J+k}, C_k, C_{K+j}$.
\end{lemma}

\begin{proof}
\textbf{Uniform attention and uniform prediction at initialization.}
When $W_{QK}=0$, because $s_{tl} = e_t^\top W_{QK} f_{x_l}$, all pre-softmax attention scores $s_{tl}$ are zero, so the attention score $\alpha_{tl}=1/(t+1)$ for all $l=0,\dots,t$.
When $W_V=0$, $o_t=\sum_{l=0}^t\alpha_{tl} W_V f_{x_l}=0$ for all $t$, hence
\begin{align*}
P_t&=\sm(
\begin{bmatrix}
\mathbf{0}
\end{bmatrix}
)=\frac{1}{K+J}\mathbf{1},\quad
g_t=W_{\lm}\delta_t=\bar f-f_{x_{t+1}},
\quad \bar f=\frac{1}{K+J}\sum_{a=1}^{K+J}f_a,
\end{align*}
where $\mathbf{1}\in\RR^{K+J}$ denotes all one vector, $\mathbf{0}\in\RR^{K+J}$ denotes all zero vector, and the second equations holds because of the fixed languages model head $W_{\lm}=[f_1, \cdots. f_{K+J}]$. The piecewise structure of $f_{x_{t+1}}$ determined by the data model given visual feature class $k\in[K]$ and visual background class $j\in[J]$ is
\begin{equation}
\label{eq:Itplus1-piecewise}
f_{x_{t+1}}=
\begin{cases}
f_{K+j}, & 1\le t\le \tau-2,\\
f_k, & \tau-1\le t\le \tau+L-2,\\
f_{K+j}, & \tau+L-1\le t\le T-1.
\end{cases}
\end{equation}
Applying Eqn.~\eqref{eq:grad-WV} in Lemma~\ref{lem:gradients} with $\alpha_{tl}=1/(t+1)$, the gradient of the single-sequence loss $\cL(k,j;\theta)$ in Eqn.~\eqref{eq:loss-kj} at initialization is
\begin{equation}
\label{eq:Gtt-master}
\nabla_{W_V}\cL(k,j;\theta^{(0)})
=\sum_{t=1}^T\sum_{l=0}^t \alpha_{tl}g_t f_{x_l}^\top=
\sum_{t=1}^{T-1}\frac{1}{t+1}\; g_t \big(\sum_{l=0}^t f_{x_l}\big)^\top
\end{equation}

\textbf{Token prefix sum.}
Define the number of feature in prefix at position $t$ as
\begin{equation}
\label{eq:nf}
n_F(t)=\big|\{\tau,\dots,\tau+L-1\}\cap\{1,\dots,t\}\big|=
\begin{cases}
0,& t<\tau,\\
t-\tau+1,& \tau\le t\le \tau+L-1,\\
L,& t\ge \tau+L-1.
\end{cases}
\end{equation}
Since position $t=0$ is the text token $x_0=K+J+k$ for visual feature class $k$ and all non-feature positions carry the same background token $x_t=K+j$ for visual background class $j$, the prefix sum is
\begin{equation}
\label{eq:Stok-explicit}
\sum_{l=0}^t f_{x_l}
=
f_{K+J+k} + n_F(t)\,f_k + \big(t-n_F(t)\big)\,f_{K+j}.
\end{equation}

\textbf{Expansion.}
Substituting Eqn.~\eqref{eq:Stok-explicit} into Eqn.~\eqref{eq:Gtt-master} and expand $g_t=\bar f-f_{x_{t+1}}$ gives
\begin{align*}
\nabla_{W_V}\cL(k,j;\theta^{(0)})
&\!=\!
\sum_{t=1}^{T-1}\frac{1}{t + 1}\bar{f}f_{K+J+k}^\top +
\sum_{t=1}^{T-1}\frac{n_F(t)}{t + 1} \bar{f}f_k^\top +
\sum_{t=1}^{T-1}\frac{t-n_F(t)}{t + 1} \bar{f}f_{K+j}^\top \\
&\quad
\!-\!\sum_{t=1}^{T-1}\frac{1}{t + 1} f_{x_{t+1}} f_{K+J+k}^\top
\!-\!\sum_{t=1}^{T-1}\frac{n_F(t)}{t + 1} f_{x_{t+1}} f_k^\top
\!-\!\sum_{t=1}^{T-1}\frac{t-n_F(t)}{t + 1} f_{x_{t+1}} f_{K+j}^\top,
\end{align*}
where we decompose and group the terms into several sums of outer products. To investigate $f_{x_{t+1}}$ within the summation, we then split each resulting time sum according to the next token:
\begin{equation}
\label{eq:back-feature}
\mathcal{B}=\{1,\dots,T-1\}\setminus\{\tau-1,\dots,\tau+L-2\},
\quad
\mathcal{F}=\{\tau-1,\dots,\tau+L-2\},
\end{equation}
where on $\mathcal{B}$ we have $f_{x_{t+1}}=f_{K+j}$, while on
$\mathcal{F}$ we have $f_{x_{t+1}}=f_k$.  Grouping coefficients of the same outer products yields
\begin{align*}
\sum_{t=1}^{T-1}\frac{1}{t+1}\, f_{x_{t+1}} f_{K+J+k}^\top
&=
\Big(\sum_{t\in\mathcal{B}}\frac{1}{t+1}\Big) f_{K+j} f_{K+J+k}^\top
+
\Big(\sum_{t\in\mathcal{F}}\frac{1}{t+1}\Big) f_k f_{K+J+k}^\top,\\
\sum_{t=1}^{T-1}\frac{n_F(t)}{t+1}\, f_{x_{t+1}} f_k^\top
&=
\Big(\sum_{t\in\mathcal{B}}\frac{n_F(t)}{t+1}\Big) f_{K+j} f_k^\top
+
\Big(\sum_{t\in\mathcal{F}}\frac{n_F(t)}{t+1}\Big) f_k f_k^\top,\\
\sum_{t=1}^{T-1}\frac{t-n_F(t)}{t+1}\, f_{x_{t+1}} f_{K+j}^\top
&=
\Big(\sum_{t\in\mathcal{B}}\frac{t-n_F(t)}{t+1}\Big) f_{K+j} f_{K+j}^\top
+
\Big(\sum_{t\in\mathcal{F}}\frac{t-n_F(t)}{t+1}\Big) f_k f_{K+j}^\top.
\end{align*}
So we plug group the terms in the summation according to their left factors in the outer product and get
\begin{align*}
\nabla_{W_V}\cL(k,j;\theta^{(0)})
&=
\bar f\Big(A_{K+J+k} f_{K+J+k}^\top+A_k f_k^\top+A_{K+j} f_{K+j}^\top\Big)\\
&\quad
-f_{K+j}\Big(B_{K+J+k} f_{K+J+k}^\top+B_k f_k^\top+B_{K+j} f_{K+j}^\top\Big)\\
&\quad
-f_k\Big(C_{K+J+k} f_{K+J+k}^\top+C_k f_k^\top+C_{K+j} f_{K+j}^\top\Big),
\end{align*}
where the coefficients can be interpreted as the learned association strengths from specific input features to target features, mediated by the linear transformation $W_V$ prior to their aggregation via the attention mechanism. $A_{K+J+k}, A_k, A_{K+j}$ means the contribution from $f_{K+J+k}, f_k, f_{K+j}$ to $\bar f$. $B_{K+J+k}, B_k, B_{K+j}$ means the contribution from $f_{K+J+k}, f_k, f_{K+j}$ to $f_{K+j}$. $C_{K+J+k}, C_k, C_{K+j}$ means the contribution from $f_{K+J+k}, f_k, f_{K+j}$ to $f_{k}$. We can further derive the closed form of these coefficient as follows:
\begin{equation}
\label{eq:coeff-real}
\begin{aligned}
A_{K+J+k}&=\sum_{t=1}^{T-1}\frac{1}{t+1}, \quad A_{k}=\sum_{t=1}^{T-1}\frac{n_F(t)}{t+1}, \quad A_{K+j}=\sum_{t=1}^{T-1}\frac{t-n_F(t)}{t+1},\\
B_{K+J+k}&=\sum_{t\in\mathcal{B}}\frac{1}{t+1},\quad B_k = \sum_{t\in\mathcal{B}}\frac{n_F(t)}{t+1}, \quad B_{K+j}=\sum_{t\in\mathcal{B}}\frac{t-n_F(t)}{t+1},\\
\quad C_{k+J+k} &= \sum_{t\in\mathcal{F}}\frac{1}{t+1},\quad C_k = \sum_{t\in\mathcal{F}}\frac{n_F(t)}{t+1}, \quad C_{K+j}=\sum_{t\in\mathcal{F}}\frac{t-n_F(t)}{t+1},
\end{aligned}
\end{equation}
which is a results of direct computation from the previous expansion.

From the closed form in Eqn.~\eqref{eq:pop-loss}, we have
\begin{align*}
\nabla_{W_V}\mathcal{L}_{\vis}(\theta^{(0)})
&= \sum_{k=1}^{K}p_k\sum_{j=1}^{J}q_j\,\nabla_{W_V}\cL(k,j;\theta^{(0)}).
\end{align*}
Thus, we conclude the proof of Lemma~\ref{lem:grad-wv}.
\end{proof}

\subsection{Step~2: Asymptotic Oder of Coefficients}
\label{sec:coefficients}

In this step, we derive the asymptotic order of the embedding coefficients in Lemma~\ref{lem:grad-wv}, which is crucial for our subsequent analysis.

\begin{proposition}[Asymptotic Order of $B$, $C$ coefficients]
\label{prop:coefficients}
Under assumptions in Theorem~\ref{thm:formal}, the
asymptotic order of coefficients $B_{K+J+k}, B_k, B_{K+j}, C_{K+J+k}, C_k, C_{K+j}$ in Lemma~\ref{lem:grad-wv} are
\begin{equation}
\label{eq:coeff-orders}
\begin{aligned}
B_{K+J+k}&=\Theta(\log T), \quad B_k=\Theta((\log T)^2), \quad B_{K+j}=\Theta(T),\\
C_{K+J+k}&=\Theta\!\left(\tfrac{\log T}{T^\beta}\right), \quad
C_k = \Theta\!\left(\tfrac{(\log T)^2}{T^\beta}\right), \quad
C_{K+j}=\Theta(\log T).
\end{aligned}
\end{equation}
\end{proposition}

\begin{proof}
Let $H_n = \sum_{i=1}^{n} 1/i$ denote the $n$-th harmonic number, and $H_n = \log n + \gamma + O(1/n)$.

\medskip
\noindent\textbf{Order of $C_{K+J+k}$.}
From the closed form of coefficient in Eqn.~\eqref{eq:coeff-real}, we have
\begin{align*}
C_{K+J+k} = \sum_{t\in\mathcal{F}}\frac{1}{t+1} = \sum_{r=0}^{L-1}\frac{1}{\tau+r} = \frac{L}{\tau}(1+o(1)) = \Theta\Big(\frac{\log T}{T^\beta}\Big),
\end{align*}
where the second equation comes from expansion using Eqn.~\eqref{eq:back-feature}, the third and fourth equation come from assumption $L = \Theta(\log T), \tau = \Theta(T^\beta)$.

\noindent\textbf{Order of $B_{K+J+k}$.}
From the closed form in Eqn.~\eqref{eq:coeff-real}, we have
\begin{align*}
B_{K+J+k} &= \sum_{t\in\mathcal{B}}\frac{1}{t+1} = \sum_{t=1}^{T-1} \frac{1}{t+1} - \sum_{t=\tau-1}^{\tau + L - 2} \frac{1}{t+1} = (H_{\tau - 1} - H_1) + (H_T - H_{\tau + L - 1}),
\end{align*}
where the second equation comes from Eqn.~\eqref{eq:back-feature}, and the third equation rewrites the terms as a harmonic sum. Expand it via the harmonic approximation,
\begin{align*}
B_{K+J+k} &= \log \tau + O(1) + \log T - \log (\tau + L - 1) + O(1) = \Theta(\log T),
\end{align*}
where the last equation comes from assumption $L = \Theta(\log T), \tau = \Theta(T^\beta)$.

\noindent\textbf{Order of $C_k$.}
From the closed form in Eqn.~\eqref{eq:coeff-real}, we have
\begin{align*}
C_k &= \sum_{t\in\mathcal{F}}\frac{n_F(t)}{t+1} = \sum_{t=\tau}^{\tau + L -2}\frac{t-\tau + 1}{t+1} = (L - 1) - \tau\sum_{r=1}^{L-1}\frac{1}{\tau + r},
\end{align*}
where the second equation uses the results in Eqn.~\eqref{eq:nf} and Eqn.~\eqref{eq:back-feature}, the third equation is a direct expansion. Using Taylor expansion $1/(\tau+r) = 1/\tau - r/\tau^2 + O(r^2/\tau^3)$, we have
\begin{align*}
\sum_{r=1}^{L-1}\frac{1}{\tau + r} = \frac{L-1}{\tau} - \frac{L(L-1)}{2\tau^2} + O\!\left(\frac{L^3}{\tau^3}\right),
\end{align*}
and plug it in the previous results, we have
\begin{align*}
C_k = (L - 1) - \tau\sum_{r=1}^{L-1}\frac{1}{\tau + r} = \frac{L(L-1)}{2\tau} + O\!\left(\frac{L^3}{\tau^2}\right)
= \Theta\!\left(\frac{(\log T)^2}{T^\beta}\right),
\end{align*}
where the last equation comes from assumption $L = \Theta(\log T), \tau = \Theta(T^\beta)$.

\noindent\textbf{Order of $B_k$.}
From the closed form in Eqn.~\eqref{eq:coeff-real}, we have
\begin{align*}
B_k &= \sum_{t\in\mathcal{B}}\frac{n_F(t)}{t+1} = \sum_{t=\tau+L-1}^{T-1}\frac{L}{t+1} = L(H_{T} - H_{\tau + L - 1}) = L\big(\log T - \log (\tau + L - 1) + O(1)\big),
\end{align*}
where the second equation uses the results in Eqn.~\eqref{eq:nf} and Eqn.~\eqref{eq:back-feature}, the third equation rewrites the terms as harmonic sum, and the fourth equation uses the approximation results of harmonic sum. Using the assumption that $L = \Theta(\log T), \tau = \Theta(T^\beta)$, we have $\log (\tau + L - 1) = \beta \log T + O(1)$, so plug it in the previous results, we have
\begin{align*}
B_k = L\big((1 - \beta)\log T + O(1)\big) = \Theta\big((\log T)^2\big).
\end{align*}

\noindent\textbf{Order of $C_{K+j}$.}
From the closed form in Eqn.~\eqref{eq:coeff-real}, we have
\begin{align*}
C_{K+j} \!=\! \sum_{t\in\mathcal{F}}\frac{t - n_F(t)}{t+1} = \!\!\sum_{t=\tau - 1}^{\tau + L - 2}\frac{t - n_F(t)}{t+1} \!= \!\frac{\tau - 1}{\tau + 1} +\!\! \sum_{t=\tau}^{\tau + L - 2}\frac{t - t + \tau - 1}{t+1} \!=\! (\tau - 1)\!\!\!\sum_{t=\tau - 1}^{\tau + L - 2}\frac{1}{t+1},
\end{align*}
where the second equation uses the results in Eqn.~\eqref{eq:back-feature}, the third equation uses the results in Eqn.~\eqref{eq:nf} the fourth equation is a direct computation. Reorganize the summation, we have
\begin{align*}
C_{K+j} = (\tau - 1)\sum_{r=0}^{L-1}\frac{1}{\tau + r} = (\tau - 1)\cdot\frac{L}{\tau}(1 + o(1)) = \Theta(L) = \Theta(\log T),
\end{align*}
where the last two equations use the assumption that $L = \Theta(\log T), \tau = \Theta(T^\beta)$.

\noindent\textbf{Order of $B_{K+j}$.}
From the closed form in Eqn.~\eqref{eq:coeff-real}, we have
\begin{align*}
B_{K+j} &= \sum_{t\in\mathcal{B}}\frac{t - n_F(t)}{t+1}\\
&= \sum_{t=1}^{\tau-2}\frac{t}{t+1} + \sum_{t=\tau+L-1}^{T-1}\frac{t - L}{t+1}\\
&= (T - L - 1) - (H_{\tau-1} - H_1) - (L+1)(H_T - H_{\tau + L - 1})\\
&= \Theta(T)
\end{align*}
where the second equation uses results in Eqn.~\eqref{eq:nf} and Eqn.~\eqref{eq:back-feature}, and the third equation rewrites the terms as harmonic sum. The fourth equation uses assumption $L = \Theta(\log T), \tau = \Theta(T^\beta)$, to get $T - L - 1 = \Theta(T)$, $H_{\tau - 1} - H_1 = \Theta(\log T)$, and $(L+1)(H_T - H_{\tau + L - 1}) = \Theta\big((\log T)^2\big)$, thus we can conclude that $B_{K+j} = \Theta(T)$.

Thus, we conclude the proof of Proposition~\ref{prop:coefficients}.
\end{proof}

\subsection{Step~3: Asymptotic Sandwich Property}
\label{sec:param-regime}

We now verify that, under the parameter regime of Theorem~\ref{thm:formal}, $q_j$ is asymptotically sandwiched between the two critical ratios $C_{K+J+k}/B_{K+J+k}$ and $C_k/B_k$. Specifically, we show that $q_j B_{K+J+k} > C_{K+J+k}$ and $q_j B_k < C_k$ for sufficiently large $T$. These conditions are pivotal for the subsequent analysis, as they ensure that the projection via $W_V^{(1)}$ maps $f_{K+J+k}$ predominantly to background tokens, while preserving $f_{k}$ as itself. 

\begin{proposition}[Asymptotic sandwich on $q_j$]
\label{prop:param-valid}
Under the assumptions of Theorem~\ref{thm:formal}, we know that $q_j = 1/J$, with $J=\bigl\lfloor\tau\log T/cL\bigr\rfloor$ for some constant $c\in\bigl(1,\,\frac{L-1}{2(1-\beta)L}\bigr)$. Then, for sufficiently large $T$,
\[
\frac{C_{K+J+k}}{B_{K+J+k}} < q_j < \frac{C_k}{B_k}.
\]
\end{proposition}

\begin{proof}
Using the exact expressions from Proposition~\ref{prop:coefficients}, we compute the two ratio bounds:
\begin{align*}
&\frac{C_{K+J+k}}{B_{K+J+k}} = \frac{\frac{L}{\tau}(1 + o(1))}{\log T + O(1)} = \frac{L}{\tau\log T} + o(1),\\
&\frac{C_{k}}{B_k} = \frac{L(L-1)/(2\tau) + O(L^3/\tau^2)}{L(1-\beta)\log T + O(1)} = \frac{L-1}{2(1-\beta)\tau\log T} + o(1),
\end{align*}
where for both row, the first equations come from the closed form in proof of Proposition~\ref{prop:coefficients}, and the second equations use the asymptotic results in Eqn.~\eqref{eq:coeff-orders}. Since $q_j = 1/J$, with $J=\bigl\lfloor\tau\log T/cL\bigr\rfloor$, when $c\in\bigl(1,\,\frac{L-1}{2(1-\beta)L}\bigr)$, we know that
\begin{align}
\label{eq:qj-region}
    \frac{L}{\tau\log T}  < \frac{cL}{\tau\log T} < \frac{L-1}{2(1-\beta)L}\frac{L}{\tau\log T} = \frac{L - 1}{2(1 - \beta)\tau \log T},
\end{align}
so $C_{K+J+k}/B_{K+J+k} < q_j < C_k/B_k$ holds for sufficiently large $T$. Equivalently, we know $q_j B_{K+J+k} > C_{K+J+k}$ and $ q_j B_k < C_k$.

Because when $\beta\in(1/2,1)$, the upper bound satisfies $\frac{L-1}{2(1-\beta)L}\to\frac{1}{2(1-\beta)}>1$ as $L\to\infty$, such a constant $c$ exists. Because $L=\Theta(\log T)$, the interval is non-empty for sufficiently large $T$.

Thus, we conclude the proof of Proposition~\ref{prop:coefficients}.
\end{proof}

\subsection{Step~4: One-Step Update of $W_V$}
\label{sec:wv-update}

Now we compute $W_V^{(1)}$, and analyze the projection of embeddings into the value space via it.

\begin{lemma}[Projection of embeddings via $W_V^{(1)}$]
\label{lem:wv-action}
After the first step update in Algorithm~\ref{alg:training}, the value matrix is
\begin{align*}
W_V^{(1)}
=
-\eta_V
\sum_{k=1}^K p_k\sum_{j=1}^J q_j \nabla_{W_V}\cL(k,j;\theta^{(0)}).
\end{align*}
Projecting $f_{K+j_1}$, $f_{k_1}$, and $f_{K+J+k_1}$, with feature class $k_1\in[K]$, background class $j_1\in[J]$, into the value space via $W_V^{(1)}$ yields:
\begin{align*}
W_{V}^{(1)}f_{K+j_1} &= \eta_V\!\big[-q_{j_1}A_{K+j_1}\bar{f} + q_{j_1} B_{K+j_1}f_{K+j_1} + q_{j_1} C_{K+j_1}\sum_k p_k f_{k}\big],\\
W_{V}^{(1)}f_{k_1} &= \eta_V\!\big[-p_{k_1} A_{k_1}\bar{f} + p_{k_1}\sum_{j} q_j B_{k_1}f_{K+j} + p_{k_1} C_{k_1}f_{k_1}\big],\\
W_{V}^{(1)}f_{K+J+k_1} &= \eta_V\!\big[-p_{k_1} A_{K+J+k_1} \bar{f} + p_{k_1}\sum_{j} q_j B_{K+J+k_1}f_{K+j} + p_{k_1}C_{K+J+k_1}f_{k_1}\big].
\end{align*}
\end{lemma}

\begin{proof}
The first step gradient descent in Algorithm~\ref{alg:training} using population loss will help us get
\begin{align*}
W_V^{(1)} = 0 - \eta_V \nabla_{W_V}\cL_{\vis}(\theta^{(0)}) = -\eta_V
\sum_{k=1}^K p_k\sum_{j=1}^J q_j \nabla_{W_V}\cL(k,j;\theta^{(0)}),
\end{align*}
which is a direct expansion using the results in Lemma~\ref{lem:grad-wv}. For $k_1\in[K], j_1\in[J]$, We can analyze the projection of an embedding into the value space via the weight matrix $W_V^{(1)}$.

For $W_{V}^{(1)}f_{K+j_1}$, we have
\begin{align*}
W_{V}^{(1)}f_{K+j_1}
&= -\eta_V\sum_kp_k\sum_jq_j \nabla_{W_V}\cL(k,j;\theta^{(0)})f_{K+j_1}\\
&= \eta_V\!\left[-q_{j_1}A_{K+j_1}\bar f
+ q_{j_1} B_{K+j_1}f_{K+j_1}
+ q_{j_1} C_{K+j_1}\textstyle\sum_k p_k f_k\right],
\end{align*}
where the second equality comes from expansion of $\nabla_{W_V}\cL(k,j;\theta^{(0)})$ using Lemma~\ref{lem:grad-wv}, and direct computation using the orthonormality of embeddings and property $\sum_{k=1}^K p_k = 1, \sum_{j=1}^J q_j = 1$.

Similarly, we apply the same technique to expand $W_{V}^{(1)}f_{k_1}, W_{V}^{(1)}f_{K+J+k_1}$ and can get
\begin{align*}
W_{V}^{(1)}f_{k_1} &= \eta_V[-p_{k_1} A_{k_1}\bar{f} + p_{k_1}\sum_{j} q_j B_{k_1}f_{K+j} + p_{k_1} C_{k_1}f_{k_1}],\\
W_{V}^{(1)}f_{K+J+k_1} &= \eta_V[-p_{k_1} A_{K+J+k_1} \bar{f} + p_{k_1}\sum_{j} q_j B_{K+J+k_1}f_{K+j} + p_{k_1}C_{K+J+k_1}f_{k_1}].
\end{align*}
Thus, we conclude the proof of Lemma~\ref{lem:wv-action}.
\end{proof}

\subsection{Step~5: Logits and Probabilities under $W_V^{(1)}$}
\label{sec:pt}

After knowing how $W_V^{(1)}$ works, we can analyze the logits and probabilities of the model after the frist update, which is crucial for the analysis of the second update.

\begin{lemma}[Logits of tokens after First step update]
\label{lem:logit-first-step}
With the parameter after the first step optimization $\theta^{(1)} = (W_V^{(1)}, 0)$, when the input is an image with visual feature class $k_1\in[K]$ and visual background class $j_1\in[J]$, i.e. $x_0 = K+J+k_1$ and $x_{1:T} \in \cI_{k_1, j_1}$, $\logit_t\in \RR^{K+J}$ satisfies
\begin{align*}
\logit_t[k] &=\tilde{A}(t) + \frac{\eta_{V}}{t+1}p_k q_{j_1} \big(t - n_F(t)\big)C_{K+j_1}, \quad k\in[K], k\neq k_1,\\
\logit_t[k_1]&=\tilde{A}(t) +\frac{\eta_{V}}{t+1}p_{k_1}\Big( q_{j_1} \big(t - n_F(t)\big)C_{K+j_1}+C_{K+J+k_1} + n_F(t) C_{k_1}\Big),\\
\logit_t[K+j] &=\tilde{A}(t) +\frac{\eta_{V}}{t+1}q_j \Big[p_{k_1} \big(B_{K+J+k_1} + n_F(t) B_{k_1} \big)\Big], \quad j \in [J], j \neq j_1,\\
\logit_t[K+j_1] &\!=\!\tilde{A}(t) \!+\!\frac{\eta_{V}}{t+1}q_{j_1} \Big[p_{k_1}\! \big(B_{K+J+k_1} + n_F(t) B_{k_1} \big)\Big] \!+\! \frac{\eta_{V}}{t+1}(t \!-\! n_F(t)) q_{j_1} B_{K+j_1},
\end{align*}
where the shared terms above also has a closed form
\begin{align*}
\tilde{A}(t) &= \frac{\eta_V}{(t+1)(K+J)}\Big[ p_{k_1} \big(A_{K+J+k_1} + n_F(t) A_{k_1} \big) + (t - n_F(t)) q_{j_1} A_{K+j_1} \Big].
\end{align*}
The coefficients $A_i, B_i, C_i$, for $i\in[K+J]$, comes from Lemma~\ref{lem:grad-wv}.
\end{lemma}

\begin{proof}
With the input sequence $x_{0:T}$, because $W_{QK}=0$, $s_{tl} = e_t^\top W_{QK} f_{x_l=0}$ for $0\leq l \leq t$, so $\alpha_{tl}=1/(t+1)$, using the definition of $o_t$ with $W_V = W_V^{(1)}$, we have
\begin{align*}
    o_t
    &=\sum_{l=0}^t \frac{1}{t+1} W_V^{(1)}f_{x_l}\\
    &= \frac{1}{t+1} W_V^{(1)} \Big(f_{K+J+k_1} + n_F(t) f_{k_1} + \big(t - n_F(t)\big)\Big)\\
    &= \frac{\eta_{V}}{t+1}\Big(- p_{k_1} A_{K+J+k_1} \bar{f} + p_{k_1}\sum_{j} q_j B_{K+J+k_1}f_{K+j} + p_{k_1}C_{K+J+k_1}f_{k_1}\\
    &\quad + n_F(t)(-p_{k_1} A_{k_1}\bar{f} + p_{k_1}\sum_{j} q_j B_{k_1}f_{K+j} + p_{k_1} C_{k_1}f_{k_1})\\
    &\quad +\big(t - n_F(t)\big)\big(- q_{j_1}A_{K+j_1}\bar{f} + (\sum_k p_kq_{j_1} B_{K+j_1})f_{K+j_1} + \sum_k p_kq_{j_1} C_{K+j_1}f_{k}\big) \Big),
\end{align*}
where the second equation use the results in Eqn.~\eqref{eq:Stok-explicit}, and the third equation use the results of $W_V^{(1)}f_{k_1}, W_V^{(1)}f_{K+J+k_1}, W_V^{(1)}f_{K+j_1}$ in Lemma~\ref{lem:wv-action}.
Reorganize it with Eqn.~\eqref{eq:nf}, we have
\begin{align*}
o_t
&= \frac{\eta_{V}}{t+1} \Bigg\{ 
 \bar{f} \bigg( - \Big[ p_{k_1} \big(A_{K+J+k_1} + n_F(t) A_{k_1} \big) + \big(t - n_F(t)\big) q_{j_1} A_{K+j_1} \Big] \bigg) \\
& + \sum_k f_k \bigg( p_k \Big[ \bbI\{k=k_1\} \big(  C_{K+J+k_1} + n_F(t) C_{k_1} \big) + (t - n_F(t)) q_{j_1} C_{K+j_1}\Big] \bigg) \\
& + \sum_j f_{K+j} \bigg( q_j \Big[ p_{k_1} \big(  B_{K+J+k_1} + n_F(t) B_{k_1} \big) \Big] + \bbI\{j=j_1\} \big(t - n_F(t)\big) q_{j_1} B_{K+j_1} \bigg) \Bigg\},
\end{align*}
where we group the terms according to the embeddings. Because $\bar f = \sum_{i=1}^{K+J} f_i /(K+J)$, we can further group the coefficient of embeddings and rewrite $o_t$ as linear combination of embeddings. Due to the weight tying $W_{\lm}$, we have $o_t = \sum_{i=1}^{K+J}\logit_t[i] f_i$, where
\begin{equation*}
\begin{aligned}
\logit_t[k] &\!=\!\tilde{A}(t) \!+\! \frac{\eta_{V}}{t+1}p_k q_{j_1} \big(t - n_F(t)\big)C_{K+j_1}, \quad k\in[K], k\neq k_1,\\
\logit_t[k_1]&\!=\!\tilde{A}(t) \!+\!\frac{\eta_{V}}{t+1}p_{k_1}\Big( q_{j_1} \big(t - n_F(t)\big)C_{K+j_1}+C_{K+J+k_1} + n_F(t) C_{k_1}\Big),\\
\logit_t[K+j] &\!=\!\tilde{A}(t) \!+\!\frac{\eta_{V}}{t+1}q_j \Big[p_{k_1} \big(B_{K+J+k_1} + n_F(t) B_{k_1} \big)\Big], \quad j \in [J], j \neq j_1,\\
\logit_t[K+j_1] &\!=\!\tilde{A}(t) \!+\!\frac{\eta_{V}}{t+1}q_{j_1} \!\Big[p_{k_1} \big(B_{K+J+k_1} + n_F(t) B_{k_1} \big)\!\Big] \!+\! \frac{\eta_{V}}{t+1}(t - n_F(t)) q_{j_1} B_{K+j_1},
\end{aligned}
\end{equation*}
where we directly decompose the factors and $\bbI\{k=k_1\}, \bbI\{j=j_1\}$. The common factor of every embedding is provided by coefficient of $\bar f$, which is
\begin{align*}
\tilde{A}(t) = \frac{\eta_V}{(t+1)(K+J)}\Big[ p_{k_1} \big(A_{K+J+k_1} + n_F(t) A_{k_1} \big) + (t - n_F(t)) q_{j_1} \sum_k p_k A_{K+j_1} \Big],
\end{align*}
which will be eliminated in the computation of softmax function.
Thus, we conclude the proof of Lemma~\ref{lem:logit-first-step}.
\end{proof}

\begin{lemma}[Concentration of probability on the background token after first step]
\label{lem:pt-concentration}
Under assumptions in Theorem~\ref{thm:formal}, after the first step optimization, for all $t\in\{1,\dots,T-1\}$
and all three phases of $n_F(t)$ in Eqn.~\eqref{eq:nf}, with an input sequence with visual feature class $k_1\in[K]$ and visual background class $j_1\in[J]$, i.e. $x_0 = K+J+k_1, x_{1:T}\in\cI_{k_1, j_1}$, the probability satisfies
\begin{align*}
P_t[i]=
\begin{cases}
e^{-\Theta(T^{1-\beta})}, & i\neq K+j_1,\\
1 - \Theta\!\big(T^\beta e^{-\Theta(T^{1-\beta})}\big), & i=K+j_1.
\end{cases}
\end{align*}
The dominant logit gap is driven by $\cC_{K+j_1}=\Theta(T^{1-\beta})$, while all other
logits are $\Theta((\log T)^3/T^\beta)$ or smaller.
\end{lemma}

\begin{proof}
From Lemma~\ref{lem:logit-first-step}, we know the closed form of logits after the first step with an input sequence with feature $k_1$ and background $j_1$ in Lemma~\ref{lem:logit-first-step}. Note that $\tilde{A}(t)$ is the same for every logits, so when we feed $\logit_t^{(1)}$ into softmax, the term $\tilde{A}(t)$ will be eliminated, which means
\begin{align*}
    \sm(\logit_t)[a]
    &=\sm\Big(\big[\logit_t[1]-\tilde{A}(t), \cdots, \logit_t[K+J]-\tilde{A}(t)\big]^\top\Big)[a],
\end{align*}
which holds because we divide $\tilde{A}(t)$ from the numerator and denominator, and then use the definition of softmax function. Therefore, the probability $P_t[i]$ only determined by $\logit_t[i] - \tilde{A}(t)$. Now we analysis their order at different region using the results in Eqn.~\eqref{eq:coeff-orders}.

When $t < \tau$, from Eqn.~\eqref{eq:nf}, $n_F(t) = 0$, we plug them in results in Lemma~\ref{lem:logit-first-step} and get
\begin{align*}
\logit_t[k]-\tilde{A}(t) &= \frac{\eta_{V}}{t+1}p_k q_{j_1} tC_{K+j_1}, \quad k\in[K], k\neq k_1,\\
\logit_t[k_1]-\tilde{A}(t)&=\frac{\eta_{V}}{t+1}p_{k_1}( q_{j_1}tC_{K+j_1}+C_{K+J+k_1}),\\
\logit_t[K+j]-\tilde{A}(t)&=\frac{\eta_{V}}{t+1}q_j p_{k_1} B_{K+J+k_1}, \quad j \in [J], j \neq j_1,\\
\logit_t[K+j_1]-\tilde{A}(t)&=\frac{\eta_{V}}{t+1}q_{j_1} p_{k_1} B_{K+J+k_1} + \frac{\eta_{V}}{t+1}t q_{j_1} B_{K+j_1}.
\end{align*}
Then we plug in the order of the coefficients computed in Eqn.~\eqref{eq:coeff-orders}, and use the assumption in Theorem~\ref{thm:formal} with direct computation, finally we get
\begin{align*}
    \logit_t[i] - \tilde{A}(t)=
    \begin{cases}
        \Theta\big(\frac{\log T}{T^\beta}\big), &i=k\in [K], k\neq k_1,\\
        \Theta\big(\frac{\log T}{T^\beta}\big), &i=k_1,\\
        \Theta\big(\frac{\log T}{(t+1)T^\beta}\big), &i=K+j, j\in[J], j\neq j_1,\\
        \Theta(T^{1-\beta}), &i=K+j_1.
    \end{cases}
\end{align*}

Similarly, from Eqn.~\eqref{eq:nf}, when $\tau \leq t \leq \tau + L - 1$, $t - n_F(t) = \tau - 1=\Theta(1)$, we plug them in Lemma~\ref{lem:logit-first-step} and use the coefficient order in Eqn.~\eqref{eq:coeff-orders} with assumption in Theorem~\ref{thm:formal}, we can get
\begin{align*}
    \logit_t[i]  - \tilde{A}(t)=
    \begin{cases}
        \Theta\big(\frac{\log T}{T^\beta}\big), &i=k\in [K], k\neq k_1,\\
        \Theta\big(\frac{\log T}{T^\beta}\big), &i=k_1,\\
        \Big[\Theta\big(\frac{(\log T)^2}{T^{2\beta}}\big), \Theta\big(\frac{(\log T)^3}{T^{2\beta}}\big)\Big], &i=K+j, j\in[J], j\neq j_1,\\
        \Theta(T^{1-\beta}), &i=K+j_1,
    \end{cases}
\end{align*}
where those for $i=K+j, j\neq j_1$ lies in an interval because from Eqn.~\ref{eq:nf}, when $\tau \leq t \leq \tau + L - 1$, $n_F(t) \in [1, L]$, and $L = \Theta(\log T)$.

Similarly, from Eqn.~\eqref{eq:nf}, when $t \geq \tau +L-1$, $n_F(t) = L$, $t - n_F = t - L$, $t$ can be at most $T$, so we also  plug them in Lemma~\ref{lem:logit-first-step} and use the coefficient order in Eqn.~\eqref{eq:coeff-orders} with assumption in Theorem~\ref{thm:formal}, we can get
\begin{align*}
    \logit_t[i]  - \tilde{A}(t)=
    \begin{cases}
        \Theta\big(\frac{\log T}{T^\beta}\big), &i=k\in [K], k\neq k_1,\\
        \Theta\big(\frac{\log T}{T^\beta}\big), &i=k_1,\\
        \Theta\big(\frac{(\log T)^3}{(t+1)T^\beta}\big), & i=K+j,j\in[J], j\neq j_1,\\
        \Theta(T^{1-\beta}), &i=K+j_1.
    \end{cases}
\end{align*}
Now feed them into softmax. Firstly, we can find that in different time region, $K+j_1$ always dominates, but we also need to consider that $J = \bigl\lfloor\tau\log T/cL\rfloor$. Therefore, for all $t$, we apply softmax function and get
\begin{align*}
\sm(\logit_t)[K+j_1]
&= \frac{1}{K\exp\big(\!\!-\!\Theta(T^{1-\beta})\big) \!+\! (J\!-\!1)\exp\big(\!\!-\!\Theta(T^{1-\beta})\big) \!+\! 1}\!=\! 1 \!-\! \Theta(T^\beta e^{-\Theta(T^{1-\beta})}),
\end{align*}
where the second equation is a results of divide $\exp\big(\logit_t[K+j_1] -\tilde{A}(t)\big)$ from the numerator and denominator, and $\logit_t[K+j_1]  - \tilde{A}(t) = \Theta(T^{1-\beta})$ for every $t\in[1, T-1]$ and dominate others. The third equation comes from direct computation. So for $i \neq K+j_1$, we have $P_t[i]=e^{-\Theta(T^{1-\beta})}$. Conclude the above results, for $t \in [1, T-1]$ we have
\begin{align*}
    P_t[i]=
    \begin{cases}
        e^{-\Theta(T^{1-\beta})}, & i\neq K+j_1\\
        1 - \Theta\big(T^\beta e^{-\Theta(T^{1-\beta})}\big), &i=K+j_1.
    \end{cases}
\end{align*}
Thus, we conclude the proof of Lemma~\ref{lem:logit-first-step}.
\end{proof}

\subsection{Step~6: Population Gradient of $W_{QK}$ after first Update}
\label{sec:grad-wqk}

In this step, we can compute the population gradient of $W_{QK}$, which will be used in the second update.

\begin{lemma}[Population gradient of $W_{QK}$]
\label{lem:grad-wqk}
At $\theta^{(1)} = (W_V^{(1)},0)$, the negative gradient of $W_{QK}$ on a sample with visual feature class $k_1\in[K]$ and visual background class $j_1\in[J]$ is
\begin{align*}
&-\nabla_{W_{QK}}\cL(k_1, j_1; \theta^{(1)}) \\
&\quad = \sum_{t = \tau - 1}^{\tau + L - 2}
\gamma(k_1,j_1, t, K+J+k_1)\, e_t f_{K+J+k_1}^\top
+ \sum_{t = \tau - 1}^{\tau + L - 2}(\tau - 1)\,\gamma(k,j, t, K+j)\, e_t f_{K+j}^\top\\
&\quad\qquad  +\sum_{t = \tau - 1}^{\tau + L - 2}(t-\tau+1)\,\gamma(k_1,j_1, t, k_1)\, e_t f_{k_1}^\top + \sum_{t=1}^{T-1}\sum_{l=0}^t \xi(k_1, j_1, t, x_l) e_tf_{x_l}^\top,
\end{align*}
where the scalar $\gamma(k_1, j_1, t, a)$ parameterd by class $k_1, j_1$, position $t$ and token $a$ satisfies
\begin{align*}
&\gamma(k_1,j_1,t,K+j_1)\\
&\quad = \frac{1}{(t+1)^2}\Big(p_{k_1}q_{j_1} B_{K+J+k_1} - p_{k_1}C_{K+J+k_1}\Big) +\frac{n_F(t)}{(t+1)^2} \Big(p_{k_1}q_{j_1}B_{k_1} - p_{k_1}C_{k_1}\Big)\\
&\quad\qquad -\frac{1 + n_F(t)}{(t+1)^2} \Big(q_{j_1} B_{K+j_1} - p_{k_1}q_{j_1} C_{K+j_1}\Big),\\
&\gamma(k_1,j_1,t,k_1)\\
&\quad = \frac{1}{(t+1)^2}\Big(p_{k_1}q_{j_1} B_{K+J+k_1} - p_{k_1}C_{K+J+k_1}\Big)-\frac{t+1-n_F(t)}{(t+1)^2}\Big(p_{k_1}q_{j_1}B_{k_1} - p_{k_1}C_{k_1}\Big)\\
&\quad\qquad +\frac{t - n_F(t)}{(t+1)^2}\Big(q_{j_1} B_{K+j_1} - p_{k_1}q_{j_1} C_{K+j_1}\Big),\\
&\gamma(k_1,j_1,t,K+J+k_1)\\
&\quad = -\frac{t}{(t+1)^2}\Big(p_{k_1}q_{j_1} B_{K+J+k_1} - p_{k_1}C_{K+J+k_1}\Big)+\frac{n_F(t)}{(t+1)^2}\Big(p_{k_1}q_{j_1}B_{k_1} - p_{k_1}C_{k_1}\Big)\\
&\quad\qquad +\frac{t - n_F(t)}{(t+1)^2}\Big(q_{j_1} B_{K+j_1} - p_{k_1}q_{j_1} C_{K+j_1}\Big).
\end{align*}
Another group of scaler $\xi(k_1, j_1, t, a)$ parameterd by class $k_1, j_1$, position $t$ and token $a$ satisfies: $\xi(k_1, j_1, t, K+j_1)=O(\frac{T}{t+1}e^{-\Theta(T^{1-\beta})})$, $\xi(k_1, j_1, t, K+J+k_1)=O(\frac{T}{t+1}e^{-\Theta(T^{1-\beta})})$, $\xi(k_1, j_1, t, k_1)=O(\frac{T}{t+1}e^{-\Theta(T^{1-\beta})})$, which are exponentially small.
\end{lemma}

\begin{proof}
Recall the deinition in Section~\ref{sec:grad-formulas}, for each position $t$, we have $\delta_t=P_t-y_t$ and $g_t=W_{\lm}\delta_t$. For a sequence with visual feature class $k_1\in [K]$, and visual background class $j_1\in [J]$, by Lemma~\ref{lem:pt-concentration}, we know that the probability $P_t^{(1)}$ concentrates on $K+j_1$, which means
\begin{align*}
\delta_t=P_t-y_t = (1 - \epsilon_{K + j_1})\bone_{K+j_1} + \sum_{i\neq K+j_1}\epsilon_i \bone_i - \bone_{x_{t+1}},
\end{align*}
where $\epsilon_{K+j_1} = \Theta(T^\beta e^{-\Theta(T^{1-\beta})})$, and $\epsilon_i = e^{-\Theta(T^{1-\beta})}, i\neq K+j_1$, and $\sum_{i=1}^{K+J} \epsilon_i = 1$. So we have
\begin{align}
\label{eq:gt-case}
g_t &=
\begin{cases}
f_{K+j_1} - f_{k_1} + \sum_{i\neq K+j_1}\epsilon_i f_i - \epsilon_{K+j_1} f_{K+j_1}, & x_{t+1} = k_1,\\
\sum_{i\neq K+j_1}\epsilon_i f_i - \epsilon_{K+j_1} f_{K+j_1}, & x_{t+1} = K + j_1,
\end{cases}
\end{align}
which is direct expansion using weight tying language model head $W_{\lm}$. Because all the $\epsilon_{i}, i\in[K+J]$ are exponentially small, when $x_{t+1} = k_1$, $g_t \approx f_{K+j_1} - f_{k_1}$; when $x_{t+1} = K+j_1$, $g_t \approx 0$, for sufficiently large $T$. We can divide them into the main part $f_{K+j_1} - f_{k_1}$ and the residual part $\sum_{i\neq K+j_1}\epsilon_i f_i - \epsilon_{K+j_1} f_{K+j_1}$.

Because $s_{tl}=e_t^\top W_{QK} f_{x_l} = 0$, the attention score is uniform, i.e. $\alpha_{tl}=1 / (t+1)$, for all $0\leq l \leq t$. So from Eqn.~\eqref{eq:grad-WQK}, we know that, at $(W_V, W_{QK})=(W_V^{(1)},0)$, the negative of the exact per-sample gradient is
\begin{align}
\label{eq:neg-wqk}
-\nabla_{W_{QK}}\cL
&=
\sum_{t=1}^{T-1}\sum_{l=0}^t
\frac{1}{t+1}\,
\Big[g_t^{\top}\big(o_t - W_V^{(1)}f_{x_l}\big)\Big]\,
e_t f_{x_l}^\top,
\end{align}
which is a simple reverse of sign in the Eqn.~\eqref{eq:grad-WQK}. To compute the closed form of the gradient of $W_{QK}$, we should consider to decompose the term $g_t^{\top}\big(o_t - W_V^{(1)}f_{x_l}\big)$. Frisly, we expand the $o_t - W_V^{(1)}f_{x_l}$ part by considering 3 kinds of embeddings respectively, so we can get
\begin{align*}
    o_t - W_{V}^{(1)} f_{K + j_1} &= W_{V}^{(1)} \Big(\frac{1}{t+1}f_{K+J+k_1} +\frac{n_F(t)}{t+1} f_{k_1} - \frac{1 + n_F(t)}{t+1}f_{K+j_1}\Big),\\
    o_t - W_{V}^{(1)} f_{k_1} &= W_{V}^{(1)} \Big(\frac{1}{t+1}f_{K+J+k_1} -\frac{t+1-n_F(t)}{t+1} f_{k_1} + \frac{t - n_F(t)}{t+1}f_{K+j_1}\Big),\\
    o_t - W_{V}^{(1)} f_{K+J+k_1} &= W_{V}^{(1)} \Big(-\frac{t}{t+1}f_{K+J+k_1} +\frac{n_F(t)}{t+1} f_{k_1} + \frac{t - n_F(t)}{t+1}f_{K+j_1}\Big),
\end{align*}
where the equation all comes from the direct computation using definition $o_t = \sum_{l=0}^t \alpha_{tl}W_V^{(1)}f_{x_l}$ with uniform attention $\alpha_{tl}=1/(t+1)$ and the results in Eqn.~\eqref{eq:Stok-explicit}.

Then we should consider the inner product $g_t^{\top}\big(o_t - W_V^{(1)}f_{x_l}\big)$. But due to the structure of $g_t$ in Eqn.~\eqref{eq:gt-case}, we need to consider the main part and the residual part seperately, and then merge them.

\textbf{Main Part:} From the structure of $g_t$ in Eqn.~\eqref{eq:gt-case}, the main part $f_{K+j_1} - f_{k_1}$ exists when $t \in \cF$ in Eqn.~\eqref{eq:back-feature}, because $x_{t+1} = k_1$. We firsly consider the quadratic form when $f_{K + j_1} - f_{k_1}$ work on $W_V^{(1)}f_{K+j_1}$ to simplify the computation. Combine the results in Lemma~\ref{lem:wv-action}, we have
\begin{align*}
    &(f_{K+j_1} - f_{k_1})^\top W_{V}^{(1)}f_{K+j_1} \\
    &\quad = -\eta_V (f_{K+j_1} - f_{k_1})^\top\sum_k p_k q_{j_1}A_{K+j_1}\bar{f}+ \eta_V (\sum_k p_kq_{j_1} B_{K+j_1})(f_{K+j_1} - f_{k_1})^\top f_{K+j_1}\\
    &\quad\qquad + \eta_V \sum_k p_kq_{j_1} C_{K+j_1}(f_{K+j_1} - f_{k_1})^\top f_{k}\\
    &\quad = \eta_Vq_{j_1} B_{K+j_1} - \eta_V p_{k_1}q_{j_1} C_{K+j_1},
\end{align*}
where the first equation is a direct expansion of the $W_V^{(1)}f_{K+j_1}$, and the second equation comes from the orthonormality of the embeddings. Similarly, using the same technique, we can get 
\begin{align*}
    (f_{K+j_1} - f_{k_1})^\top W_{V}^{(1)}f_{k_1} &= \eta_V p_{k_1}q_{j_1}B_{k_1} - \eta_V p_{k_1}C_{k_1},\\
    (f_{K+j_1} - f_{k_1})^\top W_{V}^{(1)}f_{K+J+k_1} &= \eta_V p_{k_1}q_{j_1} B_{K+J+k_1} - \eta_V p_{k_1}C_{K+J+k_1}.
\end{align*}

To simplify the computation, we denote the scaler $\gamma(k_1, j_1, t, a)=(f_{K+j}-f_k)^\top(o_t-W_V^{(1)}f_a)/(t+1)$ to pack the term befor $e_t f_{x_l}^\top$ in Eqn.~\eqref{eq:neg-wqk} for a sequence with feature class $k_1$ and background class $j_1$ when computing the loss, and $x_l = a$. When $x_l = K+j_1$, we have
\begin{align*}
    \gamma(k_1, j_1, t, K+j_1)&\!=\!\frac{1}{t+1}(f_{K+j_1} - f_{k_1})^\top(o_t - W_{V}^{(1)} f_{K + j_1})\\ 
    &\!=\! \frac{1}{(t+1)^2}(p_{k_1}q_{j_1} B_{K+J+k_1}\! -\! p_{k_1}C_{K+J+k_1})\!+\!\frac{n_F(t)}{(t+1)^2} (p_{k_1}q_{j_1}B_{k_1} \!-\! p_{k_1}C_{k_1})\\
    &\quad -\frac{1 + n_F(t)}{(t+1)^2} (q_{j_1} B_{K+j_1} - p_{k_1}q_{j_1} C_{K+j_1}),
\end{align*}
where the second equation use the results of $o_t - W_{V}^{(1)} f_{K + j_1}$ and expand them with the closed form of quadratic form $(f_{K+j_1} - f_{k_1})^\top W_{V}^{(1)}f_{K+j_1}$, $(f_{K+j_1} - f_{k_1})^\top W_{V}^{(1)}f_{k_1}$ and $(f_{K+j_1} - f_{k_1})^\top W_{V}^{(1)}f_{K+J+k_1}$. Similarly, we can use the same technique to derive those form when $x_l = k_1$ and $x_l=K+J+k_1$ as follows
\begin{align*}
    &\gamma(k_1, j_1, t, k_1)\\
    &\quad = \frac{1}{(t+1)^2}(p_{k_1}q_{j_1} B_{K+J+k_1} - p_{k_1}C_{K+J+k_1})-\frac{t+1-n_F(t)}{(t+1)^2}(p_{k_1}q_{j_1}B_{k_1} - p_{k_1}C_{k_1})\\
    &\quad\qquad +\frac{t - n_F(t)}{(t+1)^2}(q_{j_1} B_{K+j_1} - p_{k_1}q_{j_1} C_{K+j_1}),\\
    &\gamma(k_1, j_1, t, K+J+k_1)\\
    &\quad = -\frac{t}{(t+1)^2}(p_{k_1}q_{j_1} B_{K+J+k_1} - p_{k_1}C_{K+J+k_1})+\frac{n_F(t)}{(t+1)^2}(p_{k_1}q_{j_1}B_{k_1} - p_{k_1}C_{k_1})\\
    &\quad\qquad+\frac{t - n_F(t)}{(t+1)^2}(q_{j_1} B_{K+j_1} - p_{k_1}q_{j_1} C_{K+j_1}).
\end{align*}

\textbf{Residual Part:} Because $\epsilon_i$ are exponentially small, so for any $i\in [K+J]$, we have $\epsilon_if_i^\top W_{V}^{(1)}f_{K+J+k_1} = o(1)$, $\epsilon_i f_i^\top W_{V}^{(1)}f_{k_1} = o(1)$, and $\epsilon_i f_i^\top W_{V}^{(1)}f_{K+j_1} = o(1)$. We explicitly describe these in the part. For the residuals $\sum_{i\neq K+j_1}\epsilon_i f_i - \epsilon_{K+j_1} f_{K+j_1}$, we also compute the quadratic form made by it and $W_V^{(1)}f_{K+j_1}, W_V^{(1)}f_{K+J+k_1}, W_V^{(1)}f_{k_1}$. For $W_V^{(1)}f_{K+j_1}$, we have
\begin{align*}
    &(\sum_{i\neq K+j_1}\epsilon_i f_i - \epsilon_{K+j_1} f_{K+j_1})^\top W_{V}^{(1)}f_{K+j_1}\\
    &\qquad= -\eta_V (\sum_{i\neq K+j_1}\epsilon_i f_i - \epsilon_{K+j_1} f_{K+j_1})^\top\sum_k p_k q_{j_1}A_{K+j_1}\bar{f}\\
    &\qquad\qquad+ \eta_V (\sum_k p_kq_{j_1} B_{K+j_1})(\sum_{i\neq K+j_1}\epsilon_i f_i - \epsilon_{K+j_1} f_{K+j_1})^\top f_{K+j_1}\\
    &\qquad\qquad+ \eta_V \sum_k p_kq_{j_1} C_{K+j_1}(\sum_{i\neq K+j_1}\epsilon_i f_i - \epsilon_{K+j_1} f_{K+j_1})^\top f_{k}\\
    &\qquad = -\epsilon_{K+j_1}\eta_V(\sum_k p_kq_{j_1} B_{K+j_1}) + \eta_V\sum_k \epsilon_{k}p_{k}q_{j_1} C_{K+j_1}\\
    &\qquad = -\Theta(T e^{-\Theta(T^{1-\beta})}),
\end{align*}
where the first equation comes from direct expansion of $W_V^{(1)}f_{K+j_1}$ using results in Lemma~\ref{lem:wv-action}, the second equation uses the orthonormality of embeddings, and the third equation use the assumption in Theorem~\ref{thm:formal} and the order of the coefficient in Eqn.~\eqref{eq:coeff-orders}. Use the same technique, we have
\begin{align*}
(\sum_{i\neq K+j_1}\epsilon_i f_i - \epsilon_{K+j_1} f_{K+j_1})^\top W_{V}^{(1)}f_{k_1}&=\Theta\big((\log T)^2e^{-\Theta(T^{1-\beta})}\big),\\
(\sum_{i\neq K+j_1}\epsilon_i f_i - \epsilon_{K+j_1} f_{K+j_1})^\top W_{V}^{(1)}f_{K+J+k_1}&=\Theta\big(\frac{\log T}{T^{\beta}}e^{-\Theta(T^{1-\beta})}\big).
\end{align*}
To simplify the computation, we denote a scaler similar to the previous $\gamma(k_1, j_1, t, a)$, which is $\xi(k_1, j_1, t, a)=(\sum_{i\neq K+j_1}\epsilon_i f_i - \epsilon_{K+j_1} f_{K+j_1})^\top(o_t-W_V^{(1)}f_a)/(t+1)$ to pack the term befor $e_t f_{x_l}^\top$ in Eqn.~\eqref{eq:neg-wqk} for a sequence with feature class $k_1$ and background class $j_1$ when computing the loss, and $x_l = a$. And for different $x_l$, we have
\begin{align*}
    \xi(k_1, j_1, t, K+j_1)&=\frac{1}{t+1}(\sum_{i\neq K+j_1}\epsilon_i f_i - \epsilon_{K+j_1} f_{K+j_1})^\top (o_t - W_{V}^{(1)} f_{K + j_1})\\
    &= O\big(\frac{T}{t+1}e^{-\Theta(T^{1-\beta})}\big),
\end{align*}
where the second equation uses the previous expansion results of $o_t - W_{V}^{(1)} f_{K + j_1}$ and plug in the results of the order of the previous quadratic form of the residual part $\sum_{i\neq K+j_1}\epsilon_i f_i - \epsilon_{K+j_1} f_{K+j_1}$. Use similar technique for $x_l = k_1$ and $x_l=K+J+k_1$, we have
\begin{align*}
    \xi(k_1, j_1, t, k_1) = O\big(\frac{T}{t+1}e^{-\Theta(T^{1-\beta})}\big),\quad \xi(k_1, j_1, t, K+J+k_1)=O(\frac{T}{t+1}e^{-\Theta(T^{1-\beta})}).
\end{align*}

For the negative gradient for general per sequence in Eqn.~\eqref{eq:neg-wqk}, we can fixed the input sequence with a visual feature class $k_1$ and visual background class $j_1$, and expand the summation using the structure of $g_t$ in Eqn.~\eqref{eq:gt-case} and the next-token structure in Eqn.~\eqref{eq:Itplus1-piecewise}, we have
\begin{align*}
-\nabla_{W_{QK}}\cL(k_1,j_1, \theta^{(1)}) &= \sum_{t=1}^{T-1}\sum_{l=0}^t
\frac{1}{t+1}\,
\big[g_t^\top(o_t - W_{V}^{(1)}f_{x_l})\big]\,
e_t f_{x_l}^\top\\
&= \sum_{t=\tau-1}^{\tau + L - 2}\sum_{l=0}^t \big(\gamma(k_1, j_1, t, x_l) + \xi(k_1, j_1, t, x_l) \big) e_t f_{x_l}^\top\\
&\quad + \sum_{t=1}^{\tau-2}\sum_{l=0}^t \xi(k_1, j_1, t, x_l)e_t f_{x_l}^\top + \sum_{t=\tau+L-1}^{T - 1}\sum_{l=0}^t \xi(k_1, j_1, t, x_l)e_t f_{x_l}^\top\\
&= \sum_{t=\tau-1}^{\tau + L - 2}\sum_{l=0}^t \gamma(k_1, j_1, t, x_l) e_tf_{x_l}^\top + \sum_{t=1}^{T-1}\sum_{l=0}^t \xi(k_1, j_1, t, x_l) e_tf_{x_l}^\top,
\end{align*}
where the second equation is a direct expansion using Eqn.~\eqref{eq:Itplus1-piecewise} and Eqn.~\eqref{eq:gt-case}. We can further expand the formula using the data model structure and get
\begin{align*}
&-\nabla_{W_{QK}}\cL(k_1,j_1, \theta^{(1)})\\
&\quad = \sum_{t = \tau - 1}^{\tau + L - 2}
\gamma(k_1,j_1, t, K+J+k_1)\, e_t f_{K+J+k_1}^\top
+ \sum_{t = \tau - 1}^{\tau + L - 2}(\tau - 1)\,\gamma(k_1,j_1, t, K+j_1)\, e_t f_{K+j_1}^\top\\
&\quad\qquad +\sum_{t = \tau - 1}^{\tau + L - 2}(t-\tau+1)\,\gamma(k_1,j_1, t, k_1)\, e_t f_{k_1}^\top + \sum_{t=1}^{T-1}\sum_{l=0}^t \xi(k_1, j_1, t, x_l) e_t f_{x_l}^\top,
\end{align*}
where the equation comes from group of $x_l$, because at $\tau-1 \leq t\leq \tau + L-2$, the prefix contain $1$ $K+J+k_1$ token, $\tau - 1$ $K+j_1$ token, and $t - \tau + 1$ $k_1$ token. The $x_l$ in the last term is the $l$-th token in the sequence with feature and background $k_1, j_1$. We don't explicitly expand those because we will show that their effect is exponentially small.

Thus, we conclude the proof of Lemma~\ref{lem:grad-wqk}.
\end{proof}

\subsection{Step~7: One-Step Update of $W_{QK}$}
\label{sec:wqk-update}

With the population gradient of $W_{QK}$, we can compute the updated version $W_{QK}^{(1)}$ and analyze its properties.

\begin{proposition}[Position-Feature correlation computed by $W_{QK}^{(1)}$]
\label{prop:wqk}
After the second update in Algorithm~\ref{alg:training}, the query-key matrix $W_{QK}^{(1)}$ will have the following properties when computes the position-feature correlation on a sampled sequence with visual feature class $k_1\in[K]$ and visual background class $j_1\in[J]$:
\begin{align*}
e_t^\top  W_{QK}^{(1)} f_{K+j_1} &= \eta_{QK}\sum_{k=1}^K p_{k}q_{j_1}\;
(\tau - 1)\,\gamma(k,j_1, t, K+j_1)\bone\{t\in\cF\}+o(1),\\
e_t^\top  W_{QK}^{(1)} f_{k_1} &= \eta_{QK}\, p_{k_1}\sum_{j=1}^J q_{j}\;
(t - \tau + 1)\,\gamma(k_1, j, t, k_1)\bone\{t\in\cF\} + o(1),\\
e_t^\top  W_{QK}^{(1)} f_{K+J+k_1} &= \eta_{QK}\, p_{k_1}\sum_{j=1}^J q_{j}\;
\gamma(k_1, j, t, K+J+k_1)\bone\{t\in\cF\} + o(1),
\end{align*}
where $\cF$ is the set of positions whose subsequent token is a feature, as defined in Eqn.~ \ref{eq:back-feature}, and the $o(1)$ term is specifically of order $O\big(T^{2+\beta}e^{-\Theta(T^{1-\beta})}\big)$.
\end{proposition}

\begin{proof}
From the closed form of population loss in Eqn.~\eqref{eq:pop-loss}, the population gradient is
\begin{equation}
\label{eq:gradWQK-pop}
\nabla_{W_{QK}}\mathcal{L}_{\vis}(\theta^{(1)})
=
\sum_{k=1}^K p_k\sum_{j=1}^J q_j\;
\nabla_{W_{QK}}\cL(k,j;\theta^{(1)}).
\end{equation}

After the second update in Algorithm~\ref{alg:training}, and plug in the results in Lemma~\ref{lem:grad-wqk}, we can have the following form
\begin{align*}
W_{QK}^{(1)}&=-\eta_{QK}\sum_{k=1}^K p_k\sum_{j=1}^J q_j\nabla_{W_{QK}}\cL(k,j;\theta^{(1)})
\gamma(k,j, t, K+J+k)\, e_t f_{K+J+k}^\top\\
&\quad + \sum_{t = \tau - 1}^{\tau + L - 2}(\tau - 1)\,\gamma(k,j, t, K+j)\, e_t f_{K+j}^\top +\sum_{t = \tau - 1}^{\tau + L - 2}(t-\tau+1)\,\gamma(k,j, t, k)\, e_t f_{k}^\top\\
&\quad + \sum_{t=1}^{T-1}\sum_{l=0}^t \xi(k, j, t, x_l) e_t f_{x_l}^\top\Big],
\end{align*}

where the second step use the results in Lemma~\ref{lem:grad-wqk}. Suppose the input sequence has feature $k_1$ and background $j_1$, from $e_t$, we find the attention between feature $f_{k_1}, f_{K+j_1}, f_{K+J+k_1}$. 

For $e_t^\top  W_{QK}^{(1)} f_{K+j_1}$, we directly plug in the expansion in Lemma~\ref{lem:grad-wqk} and get
\begin{align*}
&e_t^\top  W_{QK}^{(1)} f_{K+j_1}\\
&\quad= \eta_{QK}\sum_{k=1}^K p_{k}q_{j_1}\;
(\tau - 1)\,\gamma(k,j_1, t, K+j_1)\bone\{t\in\cF\}+ \eta_{QK}\sum_{k=1}^K p_{k}q_{j_1} O(T^2e^{-\Theta(T^{1-\beta})})\\
&\quad= \eta_{QK}\sum_{k=1}^K p_{k}q_{j_1}\;
(\tau - 1)\,\gamma(k,j_1, t, K+j_1)\bone\{t\in\cF\}+O(T^2e^{-\Theta(T^{1-\beta})}),
\end{align*}
where the first equation also use the orthornormality of embeddings and the last step use the assumptions in Theorem~\ref{thm:formal}, which are $\eta_{QK} = \Theta(T^\beta), q_j = 1/J = \Theta(T^{-\beta})$.

For $e_t^\top  W_{QK}^{(1)} f_{k_1}$ and $e_t^\top  W_{QK}^{(1)} f_{K+J+k_1}$, with the same technique and results in Lemma~\ref{lem:grad-wqk} and assumption in Theorem~\ref{thm:formal}, we can derive
\begin{align*}
e_t^\top  W_{QK}^{(1)} f_{k_1}
&= \eta_{QK}\sum_{k=1}^K p_{k}q_{j_1}\;
(\tau - 1)\,\gamma(k,j_1, t, K+j_1)\bone\{t\in\cF\}+O(T^{2+\beta}e^{-\Theta(T^{1-\beta})}),\\
e_t^\top  W_{QK}^{(1)} f_{K+J+k_1}
&= \eta_{QK}\, p_{k_1}\sum_{j=1}^J q_{j}\;
\gamma(k_1, j, t, K+J+k_1)\bone\{t\in\cF\}+O(T^{2+\beta}e^{-\Theta(T^{1-\beta})}).
\end{align*}
We can notice that the small error provided by the residual in $g_t$ in Eqn.~\ref{eq:gt-case} is exponentially small, and at most $O\big(T^{2+\beta}e^{-\Theta(T^{1-\beta})}\big)$. Thus, we conclude the proof of proposition~\ref{prop:wqk}.
\end{proof}

\subsection{Step~8: CFG Analysis by Time Regime}
\label{sec:cfg-analysis}

From Eqn.~\eqref{eq:equivalence-cfg}, we know applying CFG is equivalent to applying CFG to the attention output. We now analyze the CFG on the attention output $o_t^{\lambda}=o_t+\lambda(o_t-o_t^{\uc})$ at the trained parameters $(W_V^{(1)},W_{QK}^{(1)})$ for each time regime.  Recall the settings above, for simplicity, we denote $\logit_t$ for $\logit_t^\cond$, and $\logit_t^\uc$ for $\logit_t^\uncond$. For clarity, but we denote the $i$-th element of $\logit_t\in\RR^{K+J}$ as $\logit_t[i]$, which also apply for $\logit_t^\uc$ and $\logit_t^\lambda$.

\begin{lemma}[Background positions: $1\le t\le\tau-2$]
\label{lem:bg-before}
When $1\leq t < \tau-1$, suppose the input tokens are $\big(K+J+k_1, (K+j_1)^{\times t}\big)$, when $\lambda < 1 + o(1)$, where $o(1)$ denotes a positive vanishing term, the CFG logits satisfies $\arg\max_i\logit_t^\lambda[i] = K+j_1$.
\end{lemma}

\begin{proof}
From Proposition~\ref{prop:wqk}, when $1 \leq t < \tau - 1$, suppose the input tokens are $[K+J+k_1, (K+j_1)^{\times t}]$, we have
\begin{align*}
o_t = \alpha_{t0} W_{V}^{(1)}f_{K+J+k_1}+ \sum_{l=1}^t \alpha_{tl} W_{V}^{(1)}f_{K+j_1},
\quad
o_t^{\uc}=\sum_{l=1}^t \alpha_{tl}^{\uc} W_{V}^{(1)}f_{K+j_1}=W_{V}^{(1)}f_{K+j_1},
\end{align*}

The output of CFG with $\lambda$ is
\begin{align*}
o_t^{\lambda} &= o_t + \lambda(o_t - o_t^{\uc})\\
&= (1+\lambda)\big(\alpha_{t0} W_V^{(1)}f_{K+J+k_1}+ \sum_{l=1}^t \alpha_{tl} W_V^{(1)}f_{K+j_1}\big) - \lambda W_V^{(1)}f_{K+j_1}\\
&= \big(1 - (1+\lambda)\alpha_{t0}\big) W_V^{(1)} f_{K+j_1} + (1 + \lambda)\alpha_{t0}W_V^{(1)} f_{K+J+k_1},
\end{align*}
where the second equation comes from expansion of $o_t, o_t^\uc$, and the last equation comes from direct computation.
For attention score $\alpha_{t0}$, $1\leq t \leq \tau - 1$, we have
\begin{align*}
\alpha_{t0} &= \frac{\exp(e_t^\top  W_{QK}^{(1)} f_{K+J+k_1})}{\exp(e_t^\top  W_{QK}^{(1)} f_{K+J+k_1}) + \sum_{s=1}^t\exp(e_s^\top  W_{QK}^{(1)} f_{K+J+k_1})} = \frac{1}{t+1}(1 + o(1)),
\end{align*}
where the last equation using the results in Proposition~\ref{prop:wqk}, that when not inside the feature block, i.e. $t\notin \cF$, $e_t^\top W_{QK}^{(1)}f_{K+J+k_1} = o(1)$, where the error is exponentially small. Similarly, we know that $\alpha_{tl} = \big(1 + o(1)\big) / (t+1)$.

Therefore, we can know that the attention score is asymptotically uniform. With the attention score, we can further expand $W_{V}^{(1)}f_{K+j_1}, W_{V}^{(1)}f_{K+J+k_1}$ using results in Lemma~\ref{lem:wv-action}, which can change $o_t^\lambda$ into a linear combination of embeddings. Because we use weight tying language model head $W_{\lm}$, the coefficient before each embedding in $o_t^\lambda$ is the corresponing value in $\logit_t^\lambda$. Observe that in Lemma~\ref{lem:wv-action}: $f_{K+j_1}$ has an additional positive term than other $f_{K+j}$ in $W_{V}^{(1)}f_{K+j_1}$; $f_{k_1}$ has an additional positive term than other $f_{k}$ in $W_{V}^{(1)}f_{K+J+k_1}$. We can know that $\logit_t^\lambda[K+j_1] > \logit_t^\lambda[K+j], j\neq j_1$, $\logit_t^\lambda[k_1] > \logit_t^\lambda[k], k\neq k_1$, so we only need to compare those of $f_{K+j_1}$ and $f_{k_1}$. By expansion using Lemma~\ref{lem:wv-action} and direct substraction of the coefficient, we have
\begin{align*}
\eta_V^{-1}\big(\logit_t^\lambda[K+j_1] - \logit_t^\lambda[k_1]\big)
&= \big(1 - (1 + \lambda) \alpha_{t0}\big)(q_{j_1} B_{K+j_1} - p_{k_1}q_{j_1}C_{K+j_1})\\
&\quad + (1 + \lambda)\alpha_{t0}(p_{k_1}q_{j_1}B_{K+J+k_1} - p_{k_1}C_{K+J+k_1}),
\end{align*}
where the shared factor in coefficient of $\bar f$ is eliminated. We know that $\logit_t^\lambda[K+j_1] - \logit_t^\lambda[k_1] > 0$ holds when 
\begin{align*}
\lambda &< \frac{q_{j_1} B_{K+j_1} - p_{k_1}q_{j_1}C_{K+j_1}}{\alpha_{t0}(q_{j_1} B_{K+j_1} - p_{k_1}q_{j_1}C_{K+j_1}p_{k_1}q_{j_1} - p_{k_1}q_{j_1}B_{K+J+k_1} + p_{k_1}C_{K+J+k_1})} - 1\\
&= t\big(1 + o(1)\big)
\end{align*}
where the first equality comes from direct computation to make the gap positive, and the second equality follows from the asymptotic orders of the coefficients in Eqn.~\eqref{eq:coeff-orders}, noting that $B_{K+j_1}$ is the dominant term, as well as the previously established order of $\alpha_{t0}$. The residual $o(1)$ here is a positive vanishing term. Therefore, when $\lambda < 1+o(1)$, $K+j_1 = \arg\max_i \logit_t^{\lambda}[i]$, so in greedy decoding at position $1 \leq t < \tau - 1$, the model will predict $K+j_1$ in the next position. 

Thus, we conclude the proof of Lemma~\ref{lem:bg-before}.
\end{proof}

\begin{lemma}[Transition position: $t=\tau-1$]
\label{lem:transition}
When $t = \tau - 1$, suppose the input tokens are $\big(K+J+k_1, (K+j_1)^{\times (\tau - 1)}\big)$, when $\lambda > o(1)$, where $o(1)$ denotes a positive vanishing term, the CFG logits satisfies $\arg\max_i\logit_t^\lambda[i] = k_1$.
\end{lemma}

\begin{proof}
Similar to Lemme~\ref{lem:bg-before}, from Proposition~\ref{prop:wqk}, when $t = \tau - 1$, suppose the input tokens are $[K+J+k_1, (K+j_1)^{\times (\tau - 1)}]$, we have
\begin{align*}
o_t = \alpha_{t0} W_V^{(1)}f_{K+J+k_1}+ \sum_{l=1}^{\tau - 1}\alpha_{tl} W_V^{(1)}f_{K+j_1},
\qquad
o_t^{\uc}=\sum_{l=1}^{\tau - 1} \alpha_{tl}^{\uc} W_V^{(1)}f_{K+j_1}=W_V^{(1)}f_{K+j_1},
\end{align*}

The output of CFG with $\lambda$ is
\begin{align*}
o_t^{\lambda} &= o_t + \lambda(o_t - o_t^{\uc})
= \big(1 - (1 + \lambda)\alpha_{t0}\big) W_V^{(1)} f_{K+j_1} + \lambda\alpha_{t0}W_V^{(1)} f_{K+J+k_1},
\end{align*}
where we use the same computation as Lemma~\ref{lem:bg-before}. For attention score $\alpha_{t0}$, $t=\tau - 1$, we have
\begin{align*}
\alpha_{t0} &= \frac{\exp(e_t^\top  W_{QK}^{(1)} f_{K+J+k_1})}{\exp(e_t^\top  W_{QK}^{(1)} f_{K+J+k_1}) + \sum_{s=1}^{\tau - 1}\exp(e_t^\top  W_{QK}^{(1)} f_{K+j_1})}\\
&= \frac{1}{1 + (\tau - 1)\exp\big(e_t^\top  W_{QK}^{(1)} (f_{K+j_1} - f_{K+J+k_1})\big)}.
\end{align*}

where we divide $\exp(e_t^\top  W_{QK}^{(1)} f_{K+J+k_1})$ from numerator and denominator. So we need to focus on the order of $e_t^\top  W_{QK}^{(1)} (f_{K+j_1} - f_{K+J+k_1})$. Using the results in Lemma~\ref{lem:grad-wqk}, because $t = \tau - 1$, $n_F(t) = 0$, we can know that 
\begin{align*}
\gamma(k,j,t,K+j_1) &= \frac{1}{(t+1)^2}(p_{k_1}q_{j_1} B_{K+J+k_1}- p_{k_1}C_{K+J+k_1})\\
&\quad -\frac{1}{(t+1)^2} (q_{j_1} B_{K+j_1} - p_{k_1}q_{j_1} C_{K+j_1}),\\
\gamma(k,j,t,K+J+k_1) &= -\frac{t}{(t+1)^2}(p_{k_1}q_{j_1} B_{K+J+k_1}- p_{k_1}C_{K+J+k_1})\\
&+\frac{t - n_F(t)}{(t+1)^2}(q_{j_1} B_{K+j_1} - p_{k_1}q_{j_1} C_{K+j_1}).
\end{align*}
Using the results in Proposition~\ref{prop:wqk} to compute the position-feature correlation, and substract those with $f_{K+j_1}$ and $f_{K+J+k_1}$, we have
\begin{align*}
&e_t^\top W_{QK}^{(1)} (f_{K+j_1} - f_{K+J+k_1})\\
&\qquad = \eta_{QK}\sum_{k=1}^K p_{k}q_{j_1}
(\tau - 1)\gamma(k,j_1, t, K+j_1) - \eta_{QK}p_{k_1}\sum_{j=1}^J q_{j}
\gamma(k_1, j, t, K+J+k_1) + o(1)\\
&\qquad = -\Theta(T^{1-\beta}),
\end{align*}
where the second equation uses the order in Eqn.~\eqref{eq:coeff-orders} and the assumptions in Theorem~\ref{thm:formal} that $\tau = \Theta(T^\beta), q_j = \Theta(T^{-\beta})$. Therefore, we can get the attention score
\begin{align*}
\alpha_{t0} = \frac{1}{1 + \Theta(T^\beta)\exp\big(-\Theta(T^{1-\beta})\big)} = 1+o(1),
\end{align*}
where the first equation use the assumption that $\tau = \Theta(T^\beta)$. So we can observe that the attention score for conditional model is concentrates on the text token at $t=0$. With the attention score, we can further expand $W_{V}^{(1)}f_{K+j_1}, W_{V}^{(1)}f_{K+J+k_1}$ using results in Lemma~\ref{lem:wv-action}, which can change $o_t^\lambda$ into a linear combination of embeddings. Similar to Lemma~\ref{lem:bg-before}, we compare the logits by comparing the coefficient before embeddings. Observe that in Lemma~\ref{lem:wv-action}: $f_{K+j_1}$ has an additional positive term than other $f_{K+j}$ in $W_{V}^{(1)}f_{K+j_1}$; $f_{k_1}$ has an additional positive term than other $f_{k}$ in $W_{V}^{(1)}f_{K+J+k_1}$. At $t=\tau - 1$ we can observe that $\logit_t^\lambda[K+j_1] > \logit_t^\lambda[K+j], j\neq j_1$, $\logit_t^\lambda[k_1] > \logit_t^\lambda[k], k\neq k_1$, so we only need to compare those of $f_{K+j_1}$ and $f_{k_1}$. By expansion using Lemma~\ref{lem:wv-action} and direct substraction of the coefficient, we have
\begin{align*}
\eta_V^{-1}\big(\logit_t^\lambda[K+j_1] - \logit_t^\lambda[k_1]\big)
&= \big(1 - (1 + \lambda) \alpha_{t0}\big)(q_{j_1} B_{K+j_1} - p_{k_1}q_{j_1}C_{K+j_1})\\
&\quad + (1 + \lambda)\alpha_{t0}(p_{k_1}q_{j_1}B_{K+J+k_1} - p_{k_1}C_{K+J+k_1}),
\end{align*}
where the shared factor in coefficient of $\bar f$ is eliminated. We know that $\logit_t^\lambda[K+j_1] - \logit_t^\lambda[k_1] < 0$ holds when 
\begin{align*}
\lambda &> \frac{q_{j_1} B_{K+j_1} - p_{k_1}q_{j_1}C_{K+j_1}}{\alpha_{t0}(q_{j_1} B_{K+j_1} - p_{k_1}q_{j_1}C_{K+j_1}p_{k_1}q_{j_1} - B_{K+J+k_1} + p_{k_1}C_{K+J+k_1})} - 1 = o(1),
\end{align*}
where the first equality comes from direct computation to make the gap positive, and the second equality follows from the asymptotic orders of the coefficients in Eqn.~\eqref{eq:coeff-orders}, noting that $B_{K+j_1}$ is the dominant term, as well as the previously established order of $\alpha_{t0}$. The $o(1)$ here is a positive vanishing term. Therefore, when $\lambda > o(1)$, $k_1 = \arg\max_i \logit_{\tau - 1}^{\lambda}[i]$, so in greedy decoding, the model will predict $k_1$ in the next position, thus successully predict the feature tokens.

Thus, we conclude the proof of Lemma~\ref{lem:transition}.
\end{proof}

\begin{lemma}[Feature block: $\tau\le t\le\tau+L-2$]
\label{lem:feature-block}
When $\tau \leq t \leq \tau + L - 2$, suppose the input tokens are $\big(K+J+k_1, (K+j_1)^{\times (\tau - 1)}, (k_1)^{\times (t-\tau + 1)}\big)$, when $\lambda < 1 + o(1)$, where $o(1)$ denotes a positive vanishing term, the CFG logits satisfies $\arg\max_i\logit_t^\lambda[i] = k_1$.
\end{lemma}

\begin{proof}
When $\tau \leq t \leq \tau + L - 2$, there are some feature token after transition. Suppose the input tokens are $\big(K+J+k_1, (K+j_1)^{\times (\tau - 1)}, k_1^{\times (t - \tau + 1)}\big)$, we have
\begin{align*}
o_t &= \alpha_{t0} W_V^{(1)}f_{K+J+k_1}+ \sum_{l=1}^{\tau - 1} \alpha_{tl} W_V^{(1)}f_{K+j_1} + \sum_{l=\tau}^{t} \alpha_{tl} W_V^{(1)}f_{k_1},\\
o_t^{\uc}&=\sum_{l=1}^t \alpha_{tl}^{\uc} W_V^{(1)}f_{K+j_1} + \sum_{l=\tau}^{t} \alpha_{tl}^{\uc} W_V^{(1)}f_{k_1},
\end{align*}
which are direct expansion using the definition. Therefore, we know that
\begin{align*}
o_{t}^{\lambda} &= o_{t} + \lambda (o_{t} - o_{t}^{\uc})\\
&= \big(- \lambda\sum_{l=1}^{\tau - 1}\alpha_{tl}^{\uc} + (1 + \lambda)\sum_{t=1}^{\tau - 1}\alpha_{tl}\big)W_V^{(1)} f_{K+j_1} + \big(-\lambda\sum_{l=\tau}^{t}\alpha_{tl}^{\uc} + (1 + \lambda)\sum_{l=\tau}^{t}\alpha_{tl}\big)W_V^{(1)} f_{k_1}\\
&\quad+ (1 + \lambda) \alpha_{t0} W_V^{(1)} f_{K+J+k_1},
\end{align*}
where the second equation comes from direct expansion of $o_t, o_t^\uc$ and group according to embeddings. Now we need to compute the attention scores. Because of the structure, we group the attention score by the embeddings. Firstly, for conditional model, we have
\begin{align*}
\alpha_{t0} &\!=\! \frac{\exp(e_t^\top W_{QK}^{(1)} f_{K+J+k_1})}{\exp(e_t^\top W_{QK}^{(1)} f_{K+J+k_1}) + (\tau - 1)\exp(e_t^\top W_{QK}^{(1)} f_{K+j_1}) + (t - \tau + 1)\exp(e_t^\top W_{QK}^{(1)} f_{k_1})}\\
&\!=\! \frac{1}{1 + (\tau \!-\! 1)\!\exp\!\big(e_t^\top W_{QK}^{(1)} (f_{K+j_1} \!-\! f_{K+J+k_1})\big) \!+\! (t \!-\! \tau \!+\! 1)\!\exp\!\big(e_t^\top W_{QK}^{(1)} (f_{k_1} \!-\! f_{K+J+k_1})\big)},
\end{align*}
where we divide $\exp(e_t^\top W_{QK}^{(1)} f_{K+J+k_1})$ from the numerator and denominator. Similarly, using the same technique, we have
\begin{align*}
\sum_{l=\tau}^t \alpha_{tl}
&\!=\!\frac{t - \tau + 1}{\exp\!\big(e_t^\top W_{QK}^{(1)}(f_{K+J+k_1} - f_{k_1})\big) \!+\! (\tau - 1)\exp\!\big(e_t^\top W_{QK}^{(1)}(f_{K+j_1} - f_{k_1})\big)\! +\! (t - \tau + 1)},\\
\sum_{l=1}^{\tau -1} \alpha_{tl}
&\!=\! \frac{\tau -1}{\exp\!\big(e_t^\top W_{QK}^{(1)}(f_{K+J+k_1}\!-\!f_{K+j_1})\big) \!+\! (\tau\!-\!1) \!+\! (t-\tau + 1) \exp\!\big(e_t^\top W_{QK}^{(1)}(f_{k_1} \!-\! f_{K+j_1})\big)},
\end{align*}
where we also write the softmax function explicitly, but divide $\exp(e_t^\top W_{QK}^{(1)} f_{K+j_1})$ and $\exp(e_t^\top W_{QK}^{(1)} f_{k_1})$ respectively. 

Similar to conditional model, but expand with a different softmax structure in unconditional model, we have
\begin{align*}
\sum_{l=\tau}^t \alpha_{tl}^{\uc} &= \frac{\sum_{l=\tau}^t \exp(e_t^\top W_{QK}^{(1)}f_{k_1})}{ \sum_{l=1}^{\tau-1}\exp(e_t^\top W_{QK}^{(1)}f_{K+j_1}) + \sum_{l=\tau}^t \exp(e_t^\top W_{QK}^{(1)}f_{k_1})}\\
&= \frac{t - \tau + 1}{(\tau - 1)\exp\big(e_t^\top W_{QK}^{(1)}(f_{K+j_1} - f_{k_1})\big) + (t - \tau + 1)},
\end{align*}
where we divide $\exp(e_t^\top W_{QK}^{(1)} f_{k_1})$ from the numerator and denominator. Similarly, using the same technique, we have
\begin{align*}
\sum_{l=1}^{\tau -1} \alpha_{tl}^{\uc}
&= \frac{\tau-1}{(\tau -1) + (t-\tau + 1) \exp\big(e_t^\top W_{QK}^{(1)}(f_{k_1} - f_{K+j_1})\big)},
\end{align*}
where we also write the softmax function explicitly, but divide $\exp(e_t^\top W_{QK}^{(1)} f_{K+j_1})$.

From Eqn.~\eqref{eq:nf}, we know that within $\tau \leq t \leq \tau + L -1$, we have $n_F(t) =  t- \tau + 1$, so using results in Lemma~\ref{lem:grad-wqk}, we know that
\begin{align*}
&\gamma(k,j,t,K+j_1) \\
&\quad = \frac{1}{(t+1)^2}(p_{k_1}q_{j_1} B_{K+J+k_1} - p_{k_1}C_{K+J+k_1}) +\frac{t - \tau + 1}{(t+1)^2} (p_{k_1}q_{j_1}B_{k_1} - p_{k_1}C_{k_1})\\
&\quad\qquad -\frac{t - \tau + 2}{(t+1)^2} (q_{j_1} B_{K+j_1} - p_{k_1}q_{j_1} C_{K+j_1}),\\
&\gamma(k,j,t,k_1) \\
&\quad = \frac{1}{(t+1)^2}(p_{k_1}q_{j_1} B_{K+J+k_1} - p_{k_1}C_{K+J+k_1}) -\frac{\tau}{(t+1)^2}(p_{k_1}q_{j_1}B_{k_1} - p_{k_1}C_{k_1})\\
&\quad\qquad +\frac{\tau - 1}{(t+1)^2}(q_{j_1} B_{K+j_1} - p_{k_1}q_{j_1} C_{K+j_1}),\\
&\gamma(k,j,t,K+J+k_1) \\
&\quad = -\frac{t}{(t+1)^2}(p_{k_1}q_{j_1} B_{K+J+k_1} - p_{k_1}C_{K+J+k_1})+\frac{t - \tau + 1}{(t+1)^2}(p_{k_1}q_{j_1}B_{k_1} - p_{k_1}C_{k_1})\\
&\quad\qquad +\frac{\tau - 1}{(t+1)^2}(q_{j_1} B_{K+j_1} - p_{k_1}q_{j_1} C_{K+j_1}),
\end{align*}
where we substitute the value of $n_F(t)$ explicitly into $\gamma$. So from the structure of attention value, we need to compute the value of the different of position-feature correlation. So using the results in Proposition~\ref{prop:wqk} to compute the position-feature correlation, and substract those with $f_{K+j_1}$ and $f_{K+J+k_1}$, we have
\begin{align*}
&e_t^\top  W_{QK}^{(1)} (f_{K+j_1} - f_{K+J+k_1})\\
&\qquad =\eta_{QK}\sum_{k=1}^K p_{k}q_{j_1}
(\tau - 1)\gamma(k,j_1, t, K+j_1) - \eta_{QK}p_{k_1}\sum_{j=1}^J q_{j}
\gamma(k_1, j, t, K+J+k_1) + o(1)\\
&\qquad = -\Theta(T^{1-\beta}),
\end{align*}
where the second equation uses the asymptotic order in Eqn.~\eqref{eq:coeff-orders} and the assumptions in Theorem~\ref{thm:formal} that $\tau = \Theta(T^\beta), q_j=\Theta(T^{-\beta})$. With the same technique, we can similarly get
\begin{align*}
e_t^\top  W_{QK}^{(1)} (f_{K+j_1} - f_{k_1})&= -\Theta((t-\tau + 1)T^{1-\beta}).
\end{align*}
However, the computation of $e_t^\top  W_{QK}^{(1)} (f_{K+j_1} - f_{k_1})$ is more difficult, and we should firstly compute a intermediate results
\begin{align*}
    &(t - \tau + 1)\gamma(k_1, j, t, k_1) - \gamma(k_1, j, t, K+J+k_1)\\
    =& \frac{(t - \tau)(\tau - 1)}{(t+1)^2}\big(p_kq_{j_1} B_{K+j_1} \!-\! p_{k_1}q_{j_1} C_{K+j_1}\big) + \frac{(2t - \tau + 1)}{(t+1)^2}\big(p_{k_1}q_{j_1} B_{K+J+k_1} \!-\! p_{k_1}C_{K+J+k_1}\big)\\
    &\quad -\frac{(t - \tau + 1)(\tau+1)}{(t+1)^2}\Big(p_{k_1}q_{j_1}B_{k_1} - p_{k_1}C_{k_1}\Big)\\
    =& 
    \begin{cases}
    \Theta((t - \tau)T^{1-2\beta}), &\tau < t \leq \tau + L - 2,\\
    O\big(\frac{(\log T)^2}{T^{2\beta}}\big), &t = \tau,
    \end{cases}
\end{align*}
where the first equation comes from direct expansion using previous explicit form of $\gamma$, and the second equation uses the asymptotic order in Eqn.~\eqref{eq:coeff-orders} with the assumptions in Theorem~\ref{thm:formal} that $\tau = \Theta(T^\beta), q_j=\Theta(T^{-\beta})$. The case when $t=\tau$ is special because $t=\tau$ will make the dominant term $B_{K+j_1}$ disappear. Therefore, expand $e_t^\top  W_{QK}^{(1)} (f_{K+j_1} - f_{k_1})$ using Proposition~\ref{prop:wqk}, we have
\begin{align*}
& e_t^\top  W_{QK}^{(1)} (f_{k_1} - f_{K+J+k_1})\\
&\quad = \eta_{QK}p_{k_1}\sum_{j=1}^J q_j \big((t - \tau + 1)\gamma(k_1, j, t, k_1) - \gamma(k_1, j, t, K+J+k_1)\big) + o(1)\\
&\quad= 
\begin{cases}
    \Theta((t - \tau)T^{1-\beta}), &\tau < t \leq \tau + L - 2,\\
    O\big(\frac{(\log T)^2}{T^{\beta}}\big), &t = \tau,
    \end{cases}
\end{align*}
where the last equation use the assumption in Theorem~\ref{thm:formal} that $\eta_{QK} = \Theta(T^\beta), q_j = \Theta(T^{-\beta})$. Next, we discuss different cases.

\textbf{When $\tau < t \leq \tau + L -2$:} From the assumption in Theorem~\ref{thm:formal} that $L=\Theta(\log T)$, we know that $0 \leq t - \tau = O(\log T)$. So plug the results of order of the terms back to the attention score computed above, we have
\begin{align*}
    \sum_{l=\tau}^{t}\alpha_{tl}^{\uc} = 1 + o(1),\quad
    \sum_{l=\tau}^{t}\alpha_{tl} = 1 + o(1), \quad \alpha_{t0} = o(1),
\end{align*}
where we also use the assumption in Theorem~\ref{thm:formal} that $\tau=\Theta(T^\beta)$, and the error $o(1)$ is exponentially small. Therefore, we can find that the attention score of conditional and unconditional model concentrate on the feature block $\tau \leq l \leq t$, so from the construction of $o_t^\lambda$, we have $o_t^{\lambda}\approx W_{V}^{(1)} f_{k_1}$ with exponentially small error, for $\lambda < 1 + o(1)$, for sufficiently large $T$. From Lemma~\ref{lem:wv-action} and order in Eqn.~\eqref{eq:coeff-orders}, we only need to compare feature $k_1$ and any background $j\in[J]$, so substract the coefficient of theem and consider the error, we can get the logit
\begin{align*}
\eta_V^{-1}\big(\logit_t^\lambda[k_1] - \logit_t^\lambda[K+j]\big) = p_{k_1} q_{j}B_{k_1} - p_{k_1}C_{k_1}f_{k_1} + o(1) > 0,
\end{align*}
where we use the results of order in Eqn.~\eqref{eq:coeff-orders} and the assumption in Theorem~\ref{thm:formal} that $q_j = \Theta(T^{-\beta})$. The error $o(1)$ here is exponentially small. 

\textbf{When $t = \tau$:} Because $\sum_{l=\tau}^{t}\alpha_{tl}^{\uc}$ doew not contain $e_t^\top  W_{QK}^{(1)} (f_{k_1} - f_{K+J+k_1})$, we have the same results $\sum_{l=\tau}^{t}\alpha_{tl}^{\uc} = 1 + o(1)$. From the assumption in Theorem~\ref{thm:formal} that $L=\Theta(\log T)$, we know that $t = \tau = O(T^\beta)$. So plug the results of order of the terms back to the attention score computed above, we have
\begin{align*}
    \sum_{l=\tau}^{t}\alpha_{tl} &=\! \frac{t - \tau + 1}{\exp\Big(O\big(\frac{(\log T)^2}{T^{2\beta}}\big)\Big) + (\tau - 1)\exp\Big(-\Theta\big((t - \tau + 1) T^{1-\beta}\big)\Big) + t - \tau + 1} = \frac{1}{2} + o(1),\\
    \alpha_{t0}
    &=\! \frac{1}{1 + (\tau - 1)\exp\big(-\Theta(T^{1-\beta})\big) + (t - \tau + 1)\exp\Big(O\big(\frac{(\log T)^2}{T^{2\beta}}\big)\Big)} = \frac{1}{2} + o(1), 
\end{align*}

where the error $o(1)$ is exponentially small. Therefore, at $t=\tau$, the conditional model seperate half of the attention to the text token, and the other half still lies on the feature block. From the construction of $o_t^\lambda$, we have $o_t^{\lambda}\approx \big(1 - (\lambda + 1)/2\big)W_{V}^{(1)} f_{k_1} + \big((\lambda + 1)/2\big)W_{V}^{(1)} f_{K+J+k_1}$ with exponentially small error under finite $\lambda$. For sufficiently large $T$, considering results in Lemma~\ref{lem:wv-action} and order in Eqn.~\eqref{eq:coeff-orders}, we only need to compare feature $k_1$ and any background $j\in[J]$, so substract the coefficient of theem and consider the error, we can get the logit
\begin{align*}
\eta_V^{-1}\big(\logit_t^\lambda[k_1] - \logit_t^\lambda[K+j]\big) &= \frac{\lambda + 1}{2}(p_{k_1}C_{K+J+k_1}\! -\! p_{k_1}C_{k_1} \!+\! p_{k_1}q_{j}B_{k_1} \!-\! p_{k_1}q_jB_{K+J+k_1})\\
&\quad + p_{k_1}C_{k_1} - p_{k_1}q_j B_{k_1} + o(1),
\end{align*}
which comes from directly expansion using Lemma~\ref{lem:wv-action}, and the error term is exponentially small. From Proposition~\ref{prop:param-valid}, we know $q_{j}B_{k_1} < C_{k_1}, q_{j}B_{K+J+k_2} > C_{K+J+k_1}$. Because $\eta_V>0$, to get $\logit_t^\lambda[k_1] - \logit_t^\lambda[K+j] > 0$, it should satisfies
\begin{align*}
    \lambda < \frac{2(p_{k_1}C_{k_1} - p_{k_1} q_{j_1}B_{k_1})}{ p_{k_1}C_{k_1} - p_{k_1}q_{j}B_{k_1} + p_{k_1}q_jB_{K+J+k_1} - p_{k_1}C_{K+J+k_1}} - 1 = 1 + o(1),
\end{align*}
where the first equality comes from direct computation to make the gap position, and the second equality follows from the asymptotic orders of the coesfficients in Eqn.~\eqref{eq:coeff-orders}, noting that $B_{K+j_1}$ is the dominant term. The $o(1)$ here is a positive vanishing term.

In conclusion, at position $\tau \leq t < \tau + L - 1$, when $\lambda < 1 + o(1)$, with positive vanishing term $o(1)$, $k_1 = \arg\max_i \logit_t^{\lambda}[i]$, so in greedy decoding, the model will predict $k_1$ in the next position, thus successully predict the feature tokens.

Thus, we conclude the proof of Lemma~\ref{lem:feature-block}.
\end{proof}

\begin{lemma}[Post-feature positions: $t\ge\tau+L-1$]
\label{lem:bg-after}
When $t \geq \tau + L - 1$, suppose the input tokens are $\big(K+J+k_1, (K+j_1)^{\times (\tau - 1)}, k_1^{\times L}, (K+j_1)^{\times (t - \tau - L + 1)}\big)$, when $\lambda < 1 + o(1)$, the CFG logits satisfies $\arg\max_i\logit_t^\lambda[i] = K + j_1$.
\end{lemma}

\begin{proof}
When $t \geq \tau + L - 1$, the feature block completes, and the target tokens are the background tokens. Suppose the input tokens are $\big(K+J+k_1, (K+j_1)^{\times (\tau - 1)}, k_1^{\times L}, (K+j_1)^{\times (t - \tau - L + 1)}\big)$, from the definition of output, we have
\begin{align*}
o_t
&= \alpha_{t0} W_V^{(1)}f_{K+J+k_1}+ (\sum_{l=1}^{\tau - 1} \alpha_{tl}+\sum_{l=\tau + L}^{t}\alpha_{tl}) W_V^{(1)}f_{K+j_1} + \sum_{l=\tau}^{\tau + L - 1} \alpha_{tl} W_V^{(1)}f_{k_1},\\
o_t^{\uc}
&= (\sum_{l=1}^{\tau - 1} \alpha_{tl}^{\uc} +\sum_{l=\tau + L}^{t}\alpha_{tl}^{\uc}) W_V^{(1)}f_{K+j_1} + \sum_{l=\tau}^{\tau + L - 1} \alpha_{tl}^{\uc} W_V^{(1)}f_{k_1},
\end{align*}
where group the terms according to the embeddings for simplification. Therefore, we know that
\begin{align*}
    o_{t}^{\lambda} &= o_{t} + \lambda (o_{t} - o_{t}^{\uc})\\
    &= \big(- \lambda(\sum_{l=1}^{\tau - 1} \alpha_{tl}^{\uc} +\sum_{l=\tau + L}^{t}\alpha_{tl}^{\uc}) + (1+\lambda)(\sum_{l=1}^{\tau - 1} \alpha_{tl} +\sum_{l=\tau + L}^{t}\alpha_{tl})\big)W_V^{(1)} f_{K+j_1}\\
    &\quad + \big(- \lambda\sum_{l=\tau}^{\tau + L -1}\alpha_{tl}^{\uc} + (1 + \lambda)\sum_{l=\tau}^{\tau + L -1}\alpha_{tl}\big)W_V^{(1)} f_{k_1}+ (1 + \lambda) \alpha_{t0} W_V^{(1)} f_{K+J+k_1}.
\end{align*}

Apply softmax function on the position-feature correlation on the sequence to compute the attention score, we have

\begin{align*}
\alpha_{t0} &= \frac{\exp(e_t^\top W_{QK}^{(1)} f_{K+J+k_1})}{\exp(e_t^\top W_{QK}^{(1)} f_{K+J+k_1}) + (\tau - 1)\exp(e_t^\top W_{QK}^{(1)} f_{K+j_1}) + (t - \tau + 1)\exp(e_t^\top W_{QK}^{(1)} f_{k_1})}\\
&= \frac{1}{1\! \!+ (\tau\! -\! 1)\exp\big(e_t^\top W_{QK}^{(1)} (f_{K+j_1} \!-\! f_{K+J+k_1})\big) \!+\! (t\! - \tau\! +\! 1)\exp\big(e_t^\top W_{QK}^{(1)} (f_{k_1} \!-\! f_{K+J+k_1})\big)}\\
&= \frac{1}{1 + t}(1 + o(1)),
\end{align*}
where the second equation divides $\exp(e_t^\top W_{QK}^{(1)} f_{K+J+k_1})$ from the numerator and denomenator, and the thrid equation comes from the properties in Proposition~\ref{prop:wqk} that when $t \notin \cF$, we have $\exp(e_t^\top W_{QK}^{(1)} f_{K+J+k_1}) = o(1)$, $\exp(e_t^\top W_{QK}^{(1)} f_{K+j_1}) = o(1)$, $\exp(e_t^\top W_{QK}^{(1)} f_{k_1}) = o(1)$, and the $o(1)$ is exponentially small. Similarly, applying the same technique to expand softmax function, we have
\begin{align*}
\sum_{l=\tau}^{\tau + L -1} \alpha_{tl} = \frac{L}{1 + t}(1 + o(1)), \quad \sum_{l=1}^{\tau -1} \alpha_{tl} + \sum_{l=\tau + L}^{t}\alpha_{tl} = \frac{t - L}{1 + t}\big(1 + o(1)\big).
\end{align*}

For unconditional model, the only difference is the structure in softmax function, so we can also directly expand them and use the properties in Proposition~\ref{prop:wqk}, and get
\begin{align*}
\sum_{l=\tau}^{\tau + L -1} \alpha_{tl}^{\uc} = \frac{L}{(t - L)\exp\big(e_t^\top W_{QK}^{(1)}(f_{K+j_1} - f_{k_1})\big) + L} = \frac{L}{t}(1 + o(1)),
\end{align*}
where the first equation divides $\exp(e_t^\top W_{QK}^{(1)} f_{k_1})$ from the numerator and denomenator, and the second equation apply the fact that $\exp(e_t^\top W_{QK}^{(1)} f_{K+j_1}) = o(1)$ and $\exp(e_t^\top W_{QK}^{(1)} f_{k_1}) = o(1)$. Similarly, we apply the same technique to the background token block, we have
\begin{align*}
\sum_{l=1}^{\tau -1} \alpha_{tl}^{\uc} + \sum_{l=\tau + L}^{t}\alpha_{tl}^{\uc} = \frac{t - L}{t}(1 + o(1)),
\end{align*}
where the error is exponentially small. Therefore, we can know that the attention pattern is almost uniform. So we can plug in the attention score to the output applied CFG
\begin{align*}
    o_{t}^{\lambda}
    &= \big(- \lambda\frac{t-L}{t} + (1 + \lambda)\frac{t-L}{t+1} +o(1)\big)W_V^{(1)} f_{K+j_1}\\
    &\quad + \big(- \lambda\frac{L}{t} + (1 + \lambda)\frac{L}{t+1} + o(1)\big)W_V^{(1)} f_{k_1}+ (1 + \lambda) \frac{1}{t+1}(1 + o(1)) W_V^{(1)} f_{K+J+k_1}.
\end{align*}
We can further expand $W_{V}^{(1)}f_{K+j_1}, W_{V}^{(1)}f_{K+J+k_1}, W_{V}^{(1)}f_{k_1}$ using results in Lemma~\ref{lem:wv-action}, which can change $o_t^\lambda$ into a linear combination of embeddings. Because we use weight tying language model head $W_{\lm}$, the coefficient before each embedding in $o_t^\lambda$ is the corresponing value in $\logit_t^\lambda$. Observe that in Lemma~\ref{lem:wv-action}: $f_{K+j_1}$ has an additional positive term than other $f_{K+j}$ in $W_{V}^{(1)}f_{K+j_1}$; $f_{k_1}$ has an additional positive term than other $f_{k}$ in $W_{V}^{(1)}f_{K+J+k_1}$.  We can know that $\logit_t^\lambda[K+j_1] > \logit_t^\lambda[K+j], j\neq j_1$, $\logit_t^\lambda[k_1] > \logit_t^\lambda[k], k\neq k_1$, so we only need to compare those of $f_{K+j_1}$ and $f_{k_1}$. By expansion using Lemma~\ref{lem:wv-action} and direct substraction of the coefficient, we have
\begin{align*}
&\eta_V^{-1}\big(\logit_t^\lambda[K+j_1] - \logit_t^\lambda[k_1]\big)\\
&\qquad=\big((1 - \lambda)\frac{t-L}{t} + \lambda\frac{t-L}{t+1}\big)q_{j_1} B_{K+j_1} + \lambda \frac{1}{t+1}p_{k_1}q_{j_1}B_{K+J+k_1}\\
&\qquad\qquad+ \big((1 - \lambda)\frac{L}{t} + \lambda\frac{L}{t+1}\big)p_{k_1}q_{j_1}B_{k_1} - \big((1 - \lambda)\frac{t-L}{t} + \lambda\frac{t-L}{t+1}\big)p_{k_1}q_{j_1}C_{K+j_1}\\
&\qquad\qquad - \lambda\frac{1}{t+1}p_{k_1}C_{K+J+k_1} - \big((1 - \lambda)\frac{L}{t} + \lambda\frac{L}{t+1}\big)p_{k_1}C_{k_1} + o(1)\\
&\qquad= \big((1 - \lambda)\frac{t-L}{t} + \lambda\frac{t-L}{t+1}\big)q_{j_1} B_{K+j_1} + o(T^{1-\beta}),
\end{align*}
where the first equation comes from direct expansion using Lemma~\ref{lem:wv-action}, and the shared factor in coefficient of $\bar f$ is eliminated. The the second equality holds because of Eqn.~\eqref{eq:coeff-orders}. Because $\eta_V>0$, we know that $\logit_t^\lambda[K+j_1] - \logit_t^\lambda[k_1] < 0$ holds for sufficiently large $T$ when 
\begin{align*}
- \lambda\frac{t-L}{t} + (1 + \lambda)\frac{t-L}{t+1} < 0,
\end{align*}
which is a direct computation by setting the gap negative. This result is equivalent to $\lambda < t + 1$. Therefore, when $\lambda < 1+o(1) < t$, at $t > \tau + L - 1$, $k_1 = \arg\max_i \logit_{t}^{\lambda}[i]$, so in greedy decoding, the model will predict $k_1$ in the next position, thus successully predict the feature tokens.

Thus, we conclude the proof of Lemma~\ref{lem:bg-after}.
\end{proof}

\subsection{Step~9: Loss Bound and Proof of Theorem~\ref{thm:formal}}
\label{sec:loss}

Notice that the input prefix discuss in the lemmas of Section~\ref{sec:cfg-analysis} are the prefix of sequences from data distribution. So we can reuse the results in these lemmas to analyze the training loss.

\begin{proposition}[Minimal Training Loss]
\label{prop:loss}
Under assumption in Theorem~\ref{thm:formal}, running the conditional model with parameter $\theta^{(2)} = (W_V^{(1)}, W_{QK}^{(1)})$, the per-position losses of the conditional model are:
\begin{itemize}
    \item For $t\notin[\tau-1,\tau+L-2]$ (background):
    $-\log P_t[I_{t+1}]
    =\Theta\!\big(T^\beta e^{-\Theta(T^{1-\beta})}\big)$.
    \item For $t=\tau-1$ (transition): $-\log P_t[k_1]=\Theta(\log T)$.
    \item For $t\in[\tau,\tau+L-2]$ (feature block): $-\log P_t[k_1]=\Theta(\log T)$.
\end{itemize}
The population loss of the model is $\cL_{\vis}(\theta^{(2)}) = \Theta\big((\log T)^2/T\big)$, which is asymptotically small.
\end{proposition}

\begin{proof}
We firstly consider an input sequence with feature class $k_1$ and background class $j_1$. Note that in Lemma~\ref{lem:bg-before}, Lemma~\ref{lem:transition}, Lemma~\ref{lem:feature-block}, and Lemma~\ref{lem:bg-after}, the input are prefix of all standard sequence with feature class $k_1$ and background class $j_1$. So we can directly use these lemma to analyze the loss with sequence from the data distribution. Here we denote $P_t[k]$ as the probability of token $k$ in probability $P_t \in \RR^{K+J}$. 

\textbf{For $t\notin \cF$:} by Lemma~\ref{lem:bg-before} and Lemma~\ref{lem:bg-after}, we know that when $t \notin \mathcal{F}$, the attention score is asymptotically uniform, i.e. $\alpha_{tl}=1/(t+1)+o(1)$, with an exponential error. So the probability the same with those in Lemma~\ref{lem:pt-concentration} (before the second update) for sufficiently large $T$. So we know that
\begin{align*}
-\log P_t[x_{t+1}]=-\log P_t[K+j_1]
    =-\log\Big(1-\Theta\big(T^\beta e^{-\Theta(T^{1-\beta})}\big)\Big)
    =\Theta\big(T^\beta e^{-\Theta(T^{1-\beta})}\big),
\end{align*}
where the first equation uses $x_{t+1}=K+j_1$ when $t\in\cF$, the second equation use results in Lemma~\ref{lem:pt-concentration} because the attention score is asymptotic uniform with exponentially small, and the third equation directly computation use $-\log(1-x)= x + O(x^2)$ when $x$ is small.

\textbf{For $t\in \cF$:} by Lemma~\ref{lem:feature-block}, we know that when $\tau < t \leq \tau + L - 2$, the attention concentrates on the feature block, i.e. $\sum_{l=\tau}^{t}\alpha_{tl} = 1 + o(1)$, with exponentially small error, so we have
\begin{align*}
o_t &= W_{V}^{(1)}f_{k_1} + o(1)(W_V^{(1)}f_{K+J+k_1} + W_V^{(1)}f_{K+j_1})\\
&=\eta_V[-p_{k_1} A_{k_1}\bar{f} + p_{k_1}\sum_{j} q_j B_{k_1}f_{K+j} + p_{k_1} C_{k_1}f_{k_1}] + o(1)(W_V^{(1)}f_{K+J+k_1} + W_V^{(1)}f_{K+j_1}),
\end{align*}
where the second equation is a direct expansion. Use the results in Eqn.~\eqref{eq:coeff-orders}, because $f_k, k\neq k_1$ only show up in $\bar f$, so the logits is $\logit_t[k] = -\eta_V p_{k_1}A_{k_1}/(K+J) + o(1), k\neq k_1$, and this comes from the exponentially small $o(1)$. Since the coefficient of $\bar f$ contribute the same to all the embeddings, we have the results of other logits
\begin{align*}
    \logit_t[k_1] &= \frac{\eta_V p_{k_1}A_{k_1}}{K+J} + \Theta\big(\frac{L(L-1)}{2\tau}\big),\\
    \quad \logit_t[K+j] &= \frac{\eta_Vp_{k_1}A_{k_1}}{K+J} + \Theta\big(\frac{cL^2(1-\beta)}{\tau}\big), j\in [J]
\end{align*}
which are results plug in the asymptotic order in Eqn.~\eqref{eq:coeff-orders} and the assumption in Theorem~\ref{thm:formal}. So we can apply softmax function to compute the probability
\begin{align*}
P_t[k_1] &= \frac{\exp\big(\logit_t[k_1]\big)}{\sum_{j=1}^J \exp\big(\logit_t[K+j]) + \sum_{k=1}^K \exp\big(\logit_t[k])\big)}\\
&= \frac{1}{\Theta(T^\beta)\exp\Big(\Theta\big(\frac{cL^2(1-\beta)}{\tau}\big) -\Theta\big(\frac{L(L-1)}{2\tau}\big)\Big) + \Theta\Big(\exp\big(-T^{-\beta}(\log T)^2\big)\Big) + \Theta(1)},
\end{align*}
where the second equation divide $\exp\big(\logit_t[k_1]\big)$ from the numerator and denominator, and then plug in the asymptotic results. Therefore, we can directly compute the log probability, which is $-\log P_t[k_1] = \Theta(\log T)$.

For $t = \tau$, also from Lemma~\ref{lem:transition}, we know that the conditional model distribute half of attention to the text token, i.e. $\alpha_{t0}=\frac{1}{2}(1+o(1))$ and $\alpha_{t\tau}=\frac{1}{2}(1+o(1))$, with exponentially small error $o(1)$, so we have
\begin{align*}
o_t &=\frac{1}{2}\eta_V[-p_{k_1} A_{k_1}\bar{f} \!+\! p_{k_1}\sum_{j} q_j B_{k_1}f_{K+j} + p_{k_1} C_{k_1}f_{k_1}]\\
&\quad + \frac{1}{2}\eta_V\!\big[\!-p_{k_1} A_{K+J+k_1} \bar{f} \!+\! p_{k_1}\sum_{j} q_j B_{K+J+k_1}f_{K+j} + p_{k_1}C_{K+J+k_1}f_{k_1}\big]\!+\! o(1)W_V^{(1)}f_{K+j_1},
\end{align*}
which is a direct results from Lemma~\ref{lem:wv-action}.
Similarly, use the results in Eqn.~\eqref{eq:coeff-orders} and notice that the coefficient of $\bar f$ contribute the same to all the embeddings, we have
\begin{align*}
    \logit_t[k_1] &= -\frac{\eta_V(p_{k_1}A_{k_1} + p_{k_1}A_{K+J+k_1})}{2(K+J)} + \Theta\big(\frac{L(L-1)}{4\tau}\big),\\
    \logit_t[K+j] &= -\frac{\eta_V(p_{k_1}A_{k_1} + p_{k_1}A_{K+J+k_1})}{2(K+J)} + \Theta\big(\frac{cL^2(1-\beta)}{2\tau}\big), j\in [J],\\
    \logit_t[k] &= -\frac{\eta_V(p_{k_1}A_{k_1} + p_{k_1}A_{K+J+k_1})}{2(K+J)} +o(1), k\in [K]
\end{align*}
which are results plug in the asymptotic order in Eqn.~\eqref{eq:coeff-orders} and the assumption in Theorem~\ref{thm:formal}. So we can apply softmax function to compute the probability
\begin{align*}
P_t[k_1] &= \frac{1}{\Theta(T^\beta)\exp\Big(\Theta\big(\frac{cL^2(1-\beta)}{2\tau}\big) - \Theta\big(\frac{L(L-1)}{4\tau}\big)\Big) + \Theta\Big(\exp\big(-T^{-\beta}(\log T)^2\big)\Big) + \Theta(1)},
\end{align*}
where we also apply the same technique to divide $\exp\big(\logit_t[k_1]\big)$ from the numerator and denominator, and then plug in the asymptotic results. Therefore, we can directly compute the log probability, which is $-\log P_t[k_1] = \Theta(\log T)$.

So in conclusion, we know that for $t\in\cF$, i.e. $\tau \leq t\leq \tau + L - 2$, every position contributes to loss with $-\log P_t[I_{t+1}] = -\log P_t[k_1] = \Theta(\log T)$.

\textbf{For $t=\tau - 1$:} From Lemma~\ref{lem:transition}, we know that when $t = \tau - 1$, the attention concentrates on the text token, i.e. $\alpha_{t0} = 1 + o(1)$, with exponentially small error, so we have
\begin{align*}
    o_t = W_V^{(1)}f_{K+J+k_1} + o(1)W_V^{(1)}(f_{K+J+k_1} + f_{K+j_1} - f_{K+J+k_0})
\end{align*}
where the equation comes from direct expansion and the error $o(1)$ is exponentially samll. Similarly, applying results from Lemma~\ref{lem:wv-action} and using the results in Eqn.~\eqref{eq:coeff-orders} and notice that the coefficient of $\bar f$ contribute the same to all the embeddings, we have
so the logits is
\begin{align*}
    \logit_t[k_1] &= -\frac{\eta_V p_{k_1}A_{K+J+k_1}}{K+J} + \Theta\big(\frac{L}{\tau}\big),\\
    \logit_t[K+j] &= -\frac{\eta_V p_{k_1}A_{K+J+k_1}}{K+J} + \Theta\big(\frac{cL}{\tau}\big), j\in [J],\\
    \logit_t[k] &= -\frac{\eta_V p_{k_1}A_{K+J+k_1}}{K+J} + o(1), k\in [K]
\end{align*}
which are results plug in the asymptotic order in Eqn.~\eqref{eq:coeff-orders} and the assumption in Theorem~\ref{thm:formal}. So we can apply softmax function to compute the probability
\begin{align*}
P_t[k_1] &= \frac{1}{\Theta(T^\beta)\exp\Big(\Theta\big((c-1)T^{-\beta} \log T\big)\Big) + \Theta\big(\exp(-T^{-\beta}\log T)\big) + \Theta(1)}
\end{align*}
where we also apply the same technique to divide $\exp\big(\logit_t[k_1]\big)$ from the numerator and denominator, and then plug in the asymptotic results. Therefore, we can directly compute the log probability, which is $-\log P_t[k_1] = \Theta(\log T)$.

In conclusion, we can compute the loss of models on a sequence with feature $k_1$ and background $j_1$
\begin{align*}
\cL(k_1, j_1; \theta^{(2)}) &= -\frac{1}{T-1}\sum_{t=1}^{T-1}\log P_t[x_{t+1}]\\
&= \frac{1}{T-1}\Big((T - L - 1)\Theta\big(T^{\beta}e^{-\Theta(T^{1-\beta})}\big) + L\Theta(\log T)\Big)\\
&= \Theta\bigg(\frac{(\log T)^2}{T}\bigg),
\end{align*}
where these first equations use the definition in Eqn.~\eqref{eq:loss-kj}, the second equation plug in the previous asymptotic results of the error at every position, and the last one use the assumption in Theorem~\ref{thm:formal} to expand. Therefore, compute the population loss use definition in Eqn.~\eqref{eq:pop-loss}, we have
\begin{align*}
\cL_{\vis}(\theta^{(2)}) = \sum_{k=1}^K p_k \sum_{j=1}^J q_j \cL(k_1, j_1;\theta^{(2)}) = \Theta\bigg(\frac{(\log T)^2}{T}\bigg).
\end{align*}
Notably, the loss is primarily driven by the errors on the feature block, which scale as $\Theta(\log T)$ at each position. In contrast, the error on the background block is exponentially small, as the model consistently predicts the background tokens with high accuracy.

Thus, we conclude the proof of Proposition~\ref{prop:loss}.
\end{proof}

\subsection{Step~10: Correctly Generation with CFG}
\label{sec:correct-gen}

With the analyze of CFG in Section~\ref{sec:cfg-analysis}, we can know analyze the correctness of generation using suitable guidance scale. Here, we use $\hat{I}^\lambda_{k, j}[t_1:t_2]$ to denote the token between position $t_1, t_2$ in the generated image $\hat{I}^\lambda_{k, j}$ with feature class $k$ and background $j$.

\begin{lemma}[Correctness of Generation with CFG]
\label{lem:correct-gen}
With guidance scale $o(1) < \lambda < 1 + o(1)$, where $o(1)$ denotes a positive vanishing term, the trained Transformer generates sequence $\hat{I}^{\lambda}_{k, j}(\theta^{(2)}) =I_{k, j}$, for all $k\in[K], j\in[J]$.
\end{lemma}

\begin{proof}
Consider feature class $k_1$ and background class $j_1$, follow the inference procedure defined in Sec.~\ref{sec:theory}, starting with prefix $(K+J+k_1, K+j_1)$.

\textbf{Before feature:} From Lemma~\ref{lem:bg-before}, when $1 \leq t < \tau - 1$, we know that with $o(1) < \lambda < 1 + o(1)$, given input $\big(K+J+k_1, (K+j_1)^{\times t}\big)$, the greedy decoding satisfies $\arg\max_{i}\logit_t[i] = K+j_1$. So the background feature at $t=2, \cdots, \tau-1$ can be correctly generated, i.e. the generated image satisfies $\hat{I}^{\lambda}_{k_1,j_1}[1:\tau - 1] = \big((K+j_1)^{\times (\tau - 1)}\big)$.

\textbf{Transition:} From Lemma~\ref{lem:transition}, when $t = \tau - 1$, we know that with $o(1) < \lambda < 1 + o(1)$, given input $\big(K+J+k_1, (K+j_1)^{\times (\tau - 1)}\big)$, the greedy decoding satisfies $\arg\max_{i}\logit_t[i] = k_1$. So the required feature at $t=\tau$ can be correctly generated, i.e. we have $\hat{I}^{\lambda}_{k_1,j_1}[1:\tau] = \big((K+j_1)^{\times (\tau - 1)},k_1\big)$.

\textbf{Feature:} From Lemma~\ref{lem:feature-block}, when $\tau \leq t \leq \tau + L - 2$, we know that with $o(1) < \lambda < 1 + o(1)$, given input $\big(K+J+k_1, (K+j_1)^{\times (\tau - 1)},(k_1)^{\times (t-\tau + 1)}\big)$, the greedy decoding will output $k_1$, i.e. $\arg\max_{i}\logit_t[i] = k_1$. So the required feature at $t=\tau+1,\cdots,\tau+L-1$ can be correctly generated, i.e. the generated image satisfies $\hat{I}_{k_1,j_1}^{\lambda}[1:\tau+L-1] = \big((K+j_1)^{\times (\tau - 1)},(k_1)^{\times L}\big)$.

\textbf{After feature:} From Lemma~\ref{lem:bg-after}, when $\tau \leq t \leq \tau + L - 2$, we know that with $o(1) < \lambda < 1 + o(1)$, given input $\big(K+J+k_1, (K+j_1)^{\times (\tau - 1)},(k_1)^{\times (t-\tau + 1)},(K+j_1)^{\times (t-\tau-L+1)}\big)$, the greedy decoding satisfies $\arg\max_{i}\logit_t[i] = K+j_1$. So the required feature at $t=\tau+L,\cdots,T$ can be correctly generated, i.e. we have $\hat{I}_{k_1,j_1}^{\lambda}[1:T] = \big((K+j_1)^{\times (\tau - 1)},(k_1)^{\times L}, (K+j_1)^{\times (T-\tau - L +1)}\big)$.

This pattern holds for all $k\in[K], j\in [J]$ because $k_1, j_1$ in the proof are arbitrary. In conclusion, the final image $\hat{I}^{\lambda}_{k_1,j_1}= I_{k_1}$, when $o(1) <\lambda < 1+o(1)$. Thus, we conclude the proof of Lemma~\ref{lem:correct-gen}.
\end{proof}

\subsection{Step~11: Incorrectly Generation with Conditional Model}
\label{sec:incorrect-gen}

Similarly, we now analyze the incorrectness of generation with conditional model. The intuition is that the conditional model fails to predict correctly at the transition position, triggering a cascade of errors in subsequent feature blocks and causing the generation to collapse into repetitive sequences.

\begin{lemma}[Incorrectness of Generation with Conditional Model]
\label{lem:incorrect-gen}
Only with conditional model, i.e. with guidance scale $\lambda = 0$, the trained Transformer generates invalid sequence that repeats background $\hat{I}^{0}_{k, j}(\theta^{(2)}) = \big((K+j)^{\times T}\big) \neq I_{k, j}$, for all $k\in[K], j\in[J]$.
\end{lemma}

\begin{proof}
Consider feature class $k_1$ and background class $j_1$, follow the inference procedure defined in Section~\ref{sec:theory}, starting with prefix $[K+J+k_1, K+j_1]$.

\textbf{Before feature:} From Lemma~\ref{lem:bg-before}, when $1 \leq t < \tau - 1$,given input $\big(K+J+k_1, (K+j_1)^{\times t}\big)$, we know that the attention pattern is asymptotic uniform, i.e. $\alpha_{tl} = 1/(t+1)$, for $l \leq t$, so the output logits is the same with those in Lemma~\ref{lem:pt-concentration} for sufficiently large $T$. Therefore, we know that $\arg\max_i \logit_t[i] = K+j_1$. So the background feature at $t=2, \cdots, \tau-1$ can be correctly generated, i.e. the generated image satisfies $\hat{I}^0_{k_1,j_1}[1:\tau - 1] = \big((K+j_1)^{\times (\tau - 1)}\big)$.

\textbf{Transition:} From Lemma~\ref{lem:transition}, when $t = \tau - 1$, given input $\big(K+J+k_1, (K+j_1)^{\times (\tau - 1)}\big)$, the attention concentrate to $f_{K+J+k_1}$, i.e. $\alpha_{t0} = 1 + o(1), \sum_{l=1}^{\tau - 1}\alpha_{tl} = o(1)$, where the error is exponentially small. With direct expansion, we have $o_t = W_V^{(1)} f_{K+J+k_1} + o(1)W_V^{(1)} (f_{K+j_1} - f_{K+J+k_1})$. Use results in Lemma~\ref{lem:wv-action}, and use similar technique in Lemma~\ref{lem:bg-before}--Lemma~\ref{lem:bg-after} to compute the logits using the coefficient before each embeddings, we have
\begin{align*}
\logit_t[k_1] &=  -\frac{\eta_Vp_{k_1} A_{K+J+k_1}}{K+J} + \eta_V p_{k_1}C_{K+J+k_1} + o(1),\\
\logit_t[K+j] &= -\frac{\eta_Vp_{k_1} A_{K+J+k_1}}{K+J} + p_{k_1}q_jB_{K+J+k_1} + o(1), j\in [J],\\
\logit_t[k] &= -\frac{\eta_Vp_{k_1} A_{K+J+k_1}}{K+J} + o(1), k\in [K],
\end{align*}
where the shared term comes from the coefficient of $\bar f$. So using Eqn.~\eqref{eq:coeff-orders} and Proposition \ref{prop:param-valid}, we know that $C_{K+J+k_1} < q_j B_{K+J+k_1}$ with sufficiently large $T$, so with greedy decoding, we have $\arg\max_i \logit_t[i] = K+j_1$. Therefore, the required feature token $k_1$ at $t=\tau$ \textbf{cannot} be correctly generated. So the generated images is already not valid, i.e. $\hat{I}_{k_1, j_1}^0(\theta^{(2)})\neq I_{k_1, j_1}$.

\textbf{Feature:} Because at $t=\tau-1$, the model predict $K+j_1$, the given input at $t=\tau$ still only contain text token and background token, i.e. $\big(K+J+k_1, (K+j_1)^{\times \tau}\big)$. Know we use induction to show that the model keep predict $K+j_1$ using greedy decoding in $t=\tau, \cdots, \tau + L - 2$.

Formally, we should prove: for $\tau \leq t \leq \tau + L - 2$, $\arg\max_i \logit_t[i] = K+j_1$ and the next input is the input is $\big(K+J+k_1, (K+j_1)^{\times (t+1)}\big)$.

\textit{Base Case:} At $t=\tau$, directly expand the softmax function with feature-position correlation, we can compute the attention pattern
\begin{align*}
\alpha_{t0} &= \frac{\exp(e_t^\top W_{QK}^{(1)} f_{K+J+k_1})}{\exp(e_t^\top W_{QK}^{(1)} f_{K+J+k_1}) + \tau\exp(e_t^\top W_{QK}^{(1)} f_{K+j_1})}\\
&= \frac{1}{1 + \tau\exp\big(e_t^\top W_{QK}^{(1)} (f_{K+j_1} - f_{K+J+k_1})\big)}\\
&= 1 + o(1),
\end{align*}
where the second divide $\exp(e_t^\top W_{QK}^{(1)} f_{K+J+k_1})$ from the numerator and denominator, and the last equality using the assumption in Theorem~\ref{thm:formal} that $\tau=\Theta(T^\beta)$ and results from Lemma~\ref{lem:feature-block} that at $t=\tau$, $e_t^\top W_{QK}^{(1)} (f_{K+j_1} - f_{K+J+k_1}) = -\Theta(T^{1-\beta})$. So we know that $\sum_{l=1}^\tau \alpha_{tl} = o(1)$, where $o(1)$ is exponentially small. Now this case degenerate to the same case as the transition case above. Similarly, with direct expansion, we have $o_t = W_V^{(1)} f_{K+J+k_1} + o(1)W_V^{(1)} (f_{K+j_1} - f_{K+J+k_1})$. Use results in Lemma~\ref{lem:wv-action}, and use similar technique in Lemma~\ref{lem:bg-before}--Lemma~\ref{lem:bg-after} to compute the logits using the coefficient before each embeddings, we have
\begin{align*}
\logit_t[k_1] &=  -\frac{\eta_Vp_{k_1} A_{K+J+k_1}}{K+J} + \eta_V p_{k_1}C_{K+J+k_1} + o(1),\\
\logit_t[K+j] &= -\frac{\eta_Vp_{k_1} A_{K+J+k_1}}{K+J} + p_{k_1}q_jB_{K+J+k_1} + o(1), j\in [J],\\
\logit_t[k] &= -\frac{\eta_Vp_{k_1} A_{K+J+k_1}}{K+J} + o(1), k\in [K],
\end{align*}
where the shared term comes from the coefficient of $\bar f$. So using Eqn.~\eqref{eq:coeff-orders} and Proposition \ref{prop:param-valid}, we know that $C_{K+J+k_1} < q_j B_{K+J+k_1}$ with sufficiently large $T$, so with greedy decoding, we have $\arg\max_i \logit_t[i] = K+j_1$. Therefore, the required feature token $k_1$ at $t=\tau+1$ \textbf{cannot} be correctly generated. So the generated images is already not valid, i.e. $\hat{I}_{k_1, j_1}^0(\theta^{(2)})\neq I_{k_1, j_1}$. And the input for next time step is $\big(K+J+k_1, (K+j_1)^{\times (\tau + 1)}\big)$.

\textit{Induction:} Suppose the induction assumptions holds for $\tau < s \leq \tau + L - 3$, we know that at $t=s+1$, the input is $\big(K+J+k_1, (K+j_1)^{\times (s + 1)}\big)$, so by directly expand the softmax function with position-feature correlation, we can compute attention score
\begin{align*}
\alpha_{t0} &= \frac{\exp(e_t^\top W_{QK}^{(1)} f_{K+J+k_1})}{\exp(e_t^\top W_{QK}^{(1)} f_{K+J+k_1}) + (s+1)\exp(e_t^\top W_{QK}^{(1)} f_{K+j_1})} = 1 + o(1),
\end{align*}
where we use the same technique as the base case, because they are almost the same except there are more terms related to background feature $f_{K+j_1}$, but $\tau < s \leq \tau + L - 3$ is still small compared to the exponential term $\exp\big(-\Theta(T^{1-\beta})\big)$ comes from $\exp\big(e_t^\top W_{QK}^{(1)} (f_{K+j_1}-f_{K+J+k_1})\big)$. So the attention pattern is the same as the base case, which concentrates on the text token at $t=0$. With the same procedure, using Eqn.~\eqref{eq:coeff-orders} and Proposition \ref{prop:param-valid}, when $T$ is sufficiently large, we have $\arg\max_i\logit_t[i] = K+j_1$. So the required feature at $t=s+1$ \textbf{cannot} be correctly generated. And the input for next time step is $\big(K+J+k_1, (K+j_1)^{\times (s + 2)}\big)$.

Therefore, we prove that in the generated image, the origional position for feature $k_1$ is now all $K+j_1$, i.e. $\hat{I}_{k_1, j_1}^0[1:\tau+L-1] = \big((K+j_1)^{\times(\tau + L -1)}\big)$.

\textbf{After feature:} 
Because from previous results, at $t=\tau+L - 1$, the input are all background token, i.e. $\hat{I}_{k_1, j_1}^0[1:\tau+L-1] = \big((K+j_1)^{\times(\tau + L -1)}\big)$. Know we use induction to show that the model keep predict $K+j_1$ using greedy decoding in $\tau + L - 1 \leq t\leq T-1$.

Formally, we should prove: for $\tau + L - 1 \leq t\leq T-1$, $\arg\max_i \logit_t[i] = K+j_1$ and the next input is the input is $\big(K+J+k_1, (K+j_1)^{\times (t+1)}\big)$.

\textit{Base Case:} At $t=\tau + L - 1$, directly expand the softmax function with feature-position correlation, we can compute the attention pattern
\begin{align*}
\alpha_{t0} &= \frac{\exp(e_t^\top W_{QK}^{(1)} f_{K+J+k_1})}{\exp(e_t^\top W_{QK}^{(1)} f_{K+J+k_1}) + (\tau + L - 1)\exp(e_t^\top W_{QK}^{(1)} f_{K+j_1})} = \frac{1}{\tau + L} + o(1),
\end{align*}
where the second divide $\exp(e_t^\top W_{QK}^{(1)} f_{K+J+k_1})$ from the numerator and denominator, and the last equality using the assumption in Theorem~\ref{thm:formal} that $\tau=\Theta(T^\beta)$ and results from Proposition~\ref{prop:wqk} that at $t\notin \cF$, $e_t^\top W_{QK}^{(1)} (f_{K+j_1} - f_{K+J+k_1}) = o(1)$, which is exponentially small. With similar technique, we know that $\sum_{l=1}^t \alpha_{tl} = (\tau + L - 1)/(\tau + L) + o(1)$. Therefore, the attention is asymptotically uniform, and using results from Lemma~\ref{lem:wv-action}, and similar technique above to compute the logits by the coefficient before each embeddings, we have
\begin{align*}
\logit_t[k] &=\tilde{A}(t) + \frac{\eta_{V}(\tau + L - 1)}{\tau + L}p_k q_{j_1} C_{K+j_1} + o(1), \quad k\in[K], k\neq k_1,\\
\logit_t[k_1]&=\tilde{A}(t) +\frac{\eta_{V}}{\tau + L}p_{k_1}\Big((\tau + L - 1)q_{j_1} C_{K+j_1}+C_{K+J+k_1}\Big)+ o(1),\\
\logit_t[K+j] &=\tilde{A}(t) +\frac{\eta_{V}}{t+1}q_j p_{k_1} B_{K+J+k_1}+o(1), \quad j \in [J], j \neq j_1,\\
\logit_t[K+j_1] &=\tilde{A}(t) +\frac{\eta_{V}}{t+1}q_{j_1}p_{k_1} B_{K+J+k_1} + \frac{\eta_{V}}{t+1}(\tau +L-1) q_{j_1}B_{K+j_1}+o(1),
\end{align*}
where the $\tilde{A}(t)$ is the shared factor in Lemma~\ref{lem:logit-first-step}, because this case is asymptotic the same as the uniform case. So using the asymptotic order in Eqn.~\ref{eq:coeff-orders}, with sufficiently large $T$, $B_{K+j_1}$ dominates, we have greedy decoding $\arg\max_i \logit_t[i] = K+j_1$. And the input for next time step is $\big(K+J+k_1, (K+j_1)^{\times (\tau + 1)}\big)$.

\textit{Induction:} Suppose the induction assumptions holds for $\tau + L -1 < s \leq T - 2$, we know that
at $t=s+1$, the input is $\big(K+J+k_1, (K+j_1)^{\times (s + 1)}\big)$, so attention pattern is
\begin{align*}
\alpha_{t0} &= \frac{\exp(e_t^\top W_{QK}^{(1)} f_{K+J+k_1})}{\exp(e_t^\top W_{QK}^{(1)} f_{K+J+k_1}) + (s+1)\exp(e_t^\top W_{QK}^{(1)} f_{K+j_1})}= \frac{1}{s + 2} + o(1),
\end{align*}
where we use the same technique as the base case, because they are almost the same except there are more terms related to background feature $f_{K+j_1}$, and in this area, $t\notin \cF$, so we still have $e_t^\top W_{QK}^{(1)} f_{K+j_1}=o(1), e_t^\top W_{QK}^{(1)} f_{k_1}=o(1)$. And we have $\sum_{l=1}^t \alpha_{tl}=(s+1)/(s+2) + o(1)$. So the attention pattern is the same as the base case, which is asymptotically uniform. With the same procedure, using Eqn.~\eqref{eq:coeff-orders} and Proposition \ref{prop:param-valid}, when $T$ is sufficiently large, we have $\arg\max_i\logit_t[i] = K+j_1$. And the input for next time step is $\big(K+J+k_1, (K+j_1)^{\times (s + 1)}\big)$.

Therefore, we finish the induction. And we know that, with greedy decoding, the predictions will repeat $K+j_1$ at every position. This pattern holds for all $k\in[K], j\in [J]$ because $k_1, j_1$ in the proof are arbitrary. In conclusion, $\hat{I}_{k, j}^0(\theta^{(2)}) = \big((K+j)^{\times T}\big) \neq I_{k,j}$. Thus, we conclude the proof of Lemma~\ref{lem:incorrect-gen}.
\end{proof}

\subsection{Step~12: Alignment of Conditional Model and Unconditional Model}

Now we prove that during sucessful generation, the conditional model and unconditional model have the same prediction with greedy decoding.

\begin{lemma}[Alignment of Conditional Model and Unconditional Model]
\label{lem:ali_cond}
With guidance scale $o(1) < \lambda < 1 + o(1)$, where $o(1)$ denotes a positive vanishing term, during the generation, the conditional model and unconditional model has the same prediction, which means
\begin{align*}
\arg\max_i \logit_t[i] = \arg\max_i \logit^\uc_t[i].
\end{align*}
\end{lemma}

\begin{proof}
Because we are now considering the unconditional model and conditional model seperately, without CFG operation. From Lemma~\ref{lem:bg-before}--Lemma~\ref{lem:bg-after}, we know that the attention pattern are concentrate on some tokens, with exponentially small error. So we firstly analyze the dominant embedding in $W_V^{(1)}f_{K+J+k_1}$, $W_V^{(1)}f_{K+j_1}$ or $W_V^{(1)}f_{k_1}$, and then discuss for different time regime.

\textbf{Dominant embedding of $W_V^{(1)}f_{K+J+k_1}, W_V^{(1)}f_{K+j_1}, W_V^{(1)}f_{k_1}$:} From Lemma~\ref{lem:wv-action}, we know that after projection through $W_V^{(1)}$, the value vector is a linear combination of embeddings
\begin{align*}
W_{V}^{(1)}f_{K+j_1} &= \eta_V\!\big[-q_{j_1}A_{K+j_1}\bar{f} + q_{j_1} B_{K+j_1}f_{K+j_1} + q_{j_1} C_{K+j_1}\sum_k p_k f_{k}\big],\\
W_{V}^{(1)}f_{k_1} &= \eta_V\!\big[-p_{k_1} A_{k_1}\bar{f} + p_{k_1}\sum_{j} q_j B_{k_1}f_{K+j} + p_{k_1} C_{k_1}f_{k_1}\big],\\
W_{V}^{(1)}f_{K+J+k_1} &= \eta_V\!\big[-p_{k_1} A_{K+J+k_1} \bar{f} + p_{k_1}\sum_{j} q_j B_{K+J+k_1}f_{K+j} + p_{k_1}C_{K+J+k_1}f_{k_1}\big].
\end{align*}
Further applying results in Eqn.~\eqref{eq:coeff-orders}, we know that: For $W_V^{(1)}f_{K+j_1}$, because $q_{j_1}B_{K+j_1}$ dominates, the dominant embedding is $f_{K+j_1}$; for $W_V^{(1)}f_{k_1}$, from Proposition~\ref{prop:param-valid}, we know that $q_jB_{k_1} <C_{k_1}$, so the dominant embedding is $f_{k_1}$; for $W_V^{(1)}f_{K+J+k_1}$, also from Proposition~\ref{prop:param-valid}, we know that $q_jB_{B+J+k_1} <C_{K+J+k_1}$, so the dominant embedding is $f_{K+J+k_1}$. In the above analysis, we ignore the coefficient for $\bar f$, because they are the same for every embeddings.

We follow the same structure of proof of Lemma~\ref{lem:correct-gen}, and we already knows that the generation is correct. Consider feature class $k_1$ and background class $j_1$, follow the inference procedure stated in Section~\ref{sec:theory}, starting with prefix $\big(K+J+k_1, K+j_1\big)$.

\textbf{Before feature:} From Lemma~\ref{lem:bg-before}, when $1 \leq t < \tau - 1$, we know that, we know that with suitable guidance scale, $o(1) < \lambda < 1 + o(1)$, given input $\big(K+J+k_1, (K+j_1)^{\times t}\big)$, the attention of conditional model is asymptotically uniform, i.e. $\alpha_{tl}=\big(1 + o(1)\big) / (t+1)$ for $0\leq l\leq t$, with exponentially small $o(1)$. Because the dominant embedding of $W_V^{(1)}f_{K+J+k_1}, W_V^{(1)}f_{K+j_1}$ are all $f_{K+j_1}$, the dominant embedding of $o_t$ is also $f_{K+j_1}$. Because of weight typing, we can know that $\arg\max_i \logit_t[i] = K+j_1$.

For unconditional model, all the prefix are $f_{K+j_1}$, so the output is $o_t^\uc = W_V^{(1)} f_{K+j_1}$. Because the dominant embedding in $W_V^{(1)} f_{K+j_1}$ is $f_{K+j_1}$ and weight tying, we know that $\arg\max_i \logit_t^\uc[i] = K+j_1$.

So when $1 \leq t < \tau - 1$, we know that $\arg\max_i \logit_t[i] = \arg\max_i \logit_t^\uc[i] = K+j_1$, the conditional model aligns with the unconditional model.

\textbf{Transition:} From Lemma~\ref{lem:transition}, when $t = \tau - 1$, we know that with $o(1) < \lambda < 1 + o(1)$, given input $\big(K+J+k_1, (K+j_1)^{\times (\tau - 1)}\big)$, the attention of conditional model concentrate on text token at $t=0$, i.e. $\alpha_{t0}=1 + o(1)$, with exponentially small error. Because the dominant embedding of $W_V^{(1)}f_{K+J+k_1}, W_V^{(1)}f_{K+j_1}$ are all $f_{K+j_1}$, the dominant embedding of $o_t$ is also $f_{K+j_1}$. Because of weight typing, we can know that $\arg\max_i \logit_t[i] = K+j_1$.

For unconditional model, all the prefix are $f_{K+j_1}$, so the output is $o_t^\uc = W_V^{(1)} f_{K+j_1}$. Because the dominant embedding in $W_V^{(1)} f_{K+j_1}$ is $f_{K+j_1}$ and weight tying, we know that the maximum is still $K+j_1$, i.e. $\arg\max_i \logit_t^\uc[i] = K+j_1$.

So when $t = \tau - 1$, we know that $\arg\max_i \logit_t[i] = \arg\max_i \logit_t^\uc[i] = K+j_1$, the conditional model aligns with the unconditional model.

\textbf{Feature:} From Lemma~\ref{lem:feature-block}, we know that with suitable guidance scale $o(1) < \lambda < 1 + o(1)$, given input $\big(K+J+k_1, (K+j_1)^{\times (\tau - 1)},(k_1)^{\times (t-\tau + 1)}\big)$, we need to discuss two cases: $t=\tau$ and $\tau < t \leq \tau + L -2$.

When $t=\tau$, from the proof of Lemma~\ref{lem:feature-block}, the conditional model half the attention score on text token at $t=0$ and the seen feature block, i.e. $\alpha_{t0} = \big(1+o(1)\big)/2$, $\alpha_{t\tau} = \big(1+o(1)\big)/2$, with exponential small error. For sufficiently large $T$, we only need to find the dominant embedding for $(W_V^{(1)}f_{K+J+k_1} + W_V^{(1)}f_{k_1}) / 2$. Use Lemma~\ref{lem:wv-action} to expand, we know that we need to compare the coefficient for $f_{K+j}, f_{k_1}$, which are $q_j(B_{K+J+k_1} + B_{k_1})/2$ and $(C_{K+J+k_1} + C_{k_1})/2$. From the results in Proposition~\ref{prop:param-valid}, we have
\begin{align*}
\frac{C_{K+J+k_1} + C_{k_1}}{B_{K+J+k_1} + B_{k_1}} = \frac{L\big(1 + o(1) \big) / \tau + L(L-1)/(2\tau) + O(L^3/\tau^2)}{\log T + O(1) + L(1-\beta)\log T + O(1)} = \frac{L-1}{2(1-\beta)\tau\log T} + o(1),
\end{align*}
where the first equation is direct expansion using exact expression in Proposition~\ref{prop:coefficients}, the second asymptotic equivalence and the last inequality use the assumption in Theorem~\ref{thm:formal}. So use the Eqn.~\eqref{eq:qj-region} and proof in Proposition~\ref{prop:coefficients}, we know that for sufficiently large $T$, $(C_{K+J+k_1} + C_{k_1})/(B_{K+J+k_1} + B_{k_1}) < q_j$. With direct computation, we know that $q_j(B_{K+J+k_1} + B_{k_1})/2 < (C_{K+J+k_1} + C_{k_1})/2$, so the dominant embedding in $o_t$ is $f_{k_1}$. Because of weight tying, we know that $\arg\max_i \logit_t^\uc[i] = k_1$.

For unconditional model, from Lemma~\ref{lem:feature-block}, the attention concentrate on the feature block, so the output is $o_t^\uc = W_V^{(1)} f_{K+j_1} + o(1)W_V^{(1)} (f_{k_1} - f_{K+j_1})$, with exponentially small $o(1)$. For sufficiently large $T$, because the dominant embedding in $W_V^{(1)} f_{k_1}$ and weight tying, we know that the maximum logit is at $k_1$, i.e. $\arg\max_i \logit_t^\uc[i] = k_1$, which is the same as unconditional model.

When $\tau < t < \tau + L -2$, from the proof of Lemma~\ref{lem:feature-block}, the conditional model and unconditional model concentrate the attention score on the feature block, i.e. $\sum_{l=\tau}^t\alpha_{tl}=1+o(1)$, $\sum_{l=\tau}^t\alpha^\uc_{tl}=1+o(1)$, with exponentially small error term $o(1)$. For sufficiently large $T$, because the dominant embedding in $W_V^{(1)} f_{k_1}$ and weight tying, we know that $\arg\max_i \logit_t[i]=\arg\max_i \logit_t^\uc[i] = k_1$, which is the same as unconditional model.

In conclusion, for feature block $\tau \leq t \leq \tau + L -2$, the conditional model and unconditional model aligns in the prediction, i.e. $\arg\max_i \logit_t[i]=\arg\max_i \logit_t^\uc[i]$.

\textbf{After feature:} From Lemma~\ref{lem:bg-after}, when $\tau \leq t \leq \tau + L - 2$, we know that with $o(1) < \lambda < 1 + o(1)$, given input $\big(K+J+k_1, (K+j_1)^{\times (\tau - 1)},(k_1)^{\times (t-\tau + 1)},(K+j_1)^{\times (t-\tau-L+1)}\big)$, the attention of conditional model and unconditional model are asymptotically uniform, i.e. $\alpha_{tl}=\big(1 + o(1)\big) / (t+1)$ for $0\leq l\leq t$, with exponentially small $o(1)$. From assumption in Theorem~\ref{thm:formal}, we know that $\tau = \Theta(T^\beta), L=\Theta(\log T)$, so the dominant part is $W_V^{(1)}f_{K+j_1}$, for sufficiently large $T$, because the number of token $K+j_1$ is the most. Because the dominant embedding of $W_Vf_{K+j_1}$ are all $f_{K+j_1}$, the dominant embedding of $o_t$ is also $f_{K+j_1}$. Because of weight typing, we can know that $\arg\max_i \logit_t[i]= \arg\max_i \logit_t^\uc[i] = K+j_1$.

In conclusion, we know that the conditional model and unconditional model has the same prediction, i.e. $\arg\max_i \logit_t[i]= \arg\max_i \logit_t^\uc[i]$. Thus, we conclude the proof of Lemma~\ref{lem:ali_cond}.
\end{proof}

\subsection{Step~13: Proof of Theorem~\ref{thm:formal}}
\label{sec:final-proof}

\begin{proof}[Proof of Theorem~\ref{thm:formal}]
The three parts of Theorem~\ref{thm:formal} can be proved with previous results.

\textbf{Part I (Loss):} From Proposition~\ref{prop:loss}, the population loss of the trained model is $\cL_{\vis}(\theta^{(2)}) = \Theta\big((\log T)^2/T\big)$.

\textbf{Part II (Correctness of Generation with CFG):} From Lemma~\ref{lem:correct-gen}, we know that with suitable guidance scale $\lambda^*\in \big(o(1), 1+o(1)\big)$, where $o(1)$ is positive vanishing term, the generated images is valid, i.e. $\hat{I}_{k, j}^{\lambda^*} = I_{k,j}$ for $k\in[K], j\in[J]$.

\textbf{Part III (Incorrectness of Generation with CFG):} From Lemma~\ref{lem:incorrect-gen}, we know that only with conditional models, i.e. $\lambda=0$, it generates invalid images that repeats backgrounds, i.e. $\hat{I}^{0}_{k, j}(\theta^{(2)}) = \big((K+j)^{\times T}\big) \neq I_{k,j}$, for all $k\in[K], j\in[J]$.

\textbf{Part IV (Alignment of Conditional and Unconditional Model):} From Lemma~\ref{lem:ali_cond}, we know that with $\lambda^*\in \big(1 + o(1)\big)$, the conditional model and unconditional model has the same prediction (with greedy decoding), which means, $\arg\max_i \logit_t[i]= \arg\max_i \logit_t^\uc[i]$.

Thus, we conclude the proof of Theorem~\ref{thm:formal}.
\end{proof}

\end{document}